\PassOptionsToPackage{numbers,sort&compress}{natbib}
\documentclass[final,12pt,numbers]{elsarticle}
\usepackage{graphicx}
\usepackage{amssymb}
\usepackage{amsmath}
\usepackage{hyperref}
\usepackage[labelfont=bf]{caption} 
\usepackage{float}
\usepackage{bm}
\usepackage{xcolor}
\usepackage{booktabs}
\usepackage{tabularx}
\usepackage{makecell}
\usepackage[title]{appendix}

\begin{document}

\begin{frontmatter}

\title{Spectral bias in physics-informed and operator learning: Analysis and mitigation guidelines}

\author[1]{Siavash Khodakarami}
\author[2]{Vivek Oommen}
\author[2]{Nazanin Ahmadi Daryakenari}
\author{Maxim Beekenkamp}
\author[1]{George Em Karniadakis\corref{cor1}}

\address[1]{Division of Applied Mathematics, Brown University, Providence, RI 02912, USA}
\address[2]{School of Engineering, Brown University, Providence, RI 02912, USA}

\cortext[cor1]{Corresponding author. Email: george\_karniadakis@brown.edu}

\begin{abstract}
Solving partial differential equations (PDEs) by neural networks as well as Kolmogorov-Arnold Networks (KANs), including physics-informed neural networks (PINNs), physics-informed KANs (PIKANs), and neural operators, are known to exhibit spectral bias, whereby low-frequency components of the solution are learned significantly faster than high-frequency modes. While spectral bias is often treated as an intrinsic representational limitation of neural architectures, its interaction with optimization dynamics and physics-based loss formulations remains poorly understood. In this work, we provide a systematic investigation of spectral bias in physics-informed and operator learning frameworks, with emphasis on the coupled roles of network architecture, activation functions, loss design, and optimization strategy.
We quantify spectral bias through frequency-resolved error metrics, Barron-norm diagnostics, and higher-order statistical moments, enabling a unified analysis across elliptic, hyperbolic, and dispersive PDEs. Through diverse benchmark problems, including the Korteweg-de Vries, wave and steady-state diffusion-reaction equations, turbulent flow reconstruction, and earthquake dynamics, we demonstrate that spectral bias is not simply representational but fundamentally dynamical. In particular, second-order optimization methods substantially alter the spectral learning order, enabling earlier and more accurate recovery of high-frequency modes for all PDE types. 
For neural operators, we further show that spectral bias is dependent on the neural operator architecture and can also be effectively mitigated through spectral-aware loss formulations without increasing the inference cost.    
\end{abstract}

\begin{keyword}
    Spectral bias \sep PINN \sep PIKAN \sep Neural operator \sep second-order optimization
\end{keyword}

\end{frontmatter}

\section{Introduction} 

\noindent The simulation of partial differential equations (PDEs) remains the cornerstone of modern engineering and scientific discovery, driving progress in fields such as fluid dynamics, thermal transport, and structural mechanics. While traditional discretization schemes including Finite Element (FEM) and Finite Difference (FDM) methods offer high reliability, they are often constrained by high computational costs and the necessity for time-intensive mesh generation. These bottlenecks become particularly restrictive in iterative contexts, such as parametric optimization, uncertainty quantification, and inverse problems, where repeated high-fidelity simulations are required.\\

Recent progress in scientific machine learning has given rise to neural PDE solvers, which employ neural networks to approximate solutions to PDEs while directly incorporating physical constraints into the training process. Among these methods, physics-informed neural networks (PINNs) \cite{RAISSI2019686} have attracted considerable interest for their capacity to enforce governing equations, boundary conditions, and initial conditions via automatic differentiation. By minimizing the residuals of the PDEs across a continuous domain, PINNs offer a mesh-free approach that seamlessly handles complex or irregular geometries, limited data, and inverse problems. Consequently, PINNs have been effectively applied to both forward and inverse problems in a variety of physical areas such as fluid mechanics, heat transfer, wave propagation, and coupled multiphysics systems \cite{review1, toscano2410pinns, cai2021physics, zhao2024comprehensive, HE2020103610, shukla2020physics}. In parallel, a distinct class of learners known as neural operators (e.g., Fourier neural operator (FNO) \cite{li2021fourier} and DeepONet \cite{lu2021learning}) has emerged. Unlike PINNs, which optimize for a single solution instance, neural operators aim to learn the underlying mapping between infinite-dimensional function spaces and learning the solution operator of the PDE. These methods have demonstrated strong performance in surrogate modeling, uncertainty quantification, and real-time prediction scenarios. \\

Despite their expressive power, neural PDE solvers are fundamentally influenced by the spectral bias inherent to neural network architectures and training dynamics. Spectral bias refers to the tendency of neural networks to learn low-frequency (smooth) components of a target function more rapidly than high-frequency or multiscale features \cite{rahaman2019spectral}. In the context of PINNs, this bias manifests itself as an imbalance in the learning of solution modes, where smooth global structures are captured early during training while sharp gradients, boundary layers, and localized phenomena remain under-resolved. As a result, PINNs may satisfy PDE residuals in an averaged sense while failing to accurately represent fine-scale physics.\\

Spectral bias also plays a central role in the performance of neural operators \cite{KHODAKARAMI2026108027, oommen2025integrating}. Architectures such as the Fourier neural operator explicitly operate in spectral space and typically truncate or attenuate high-frequency modes to improve numerical stability and computational efficiency. Although this design enables strong generalization and rapid inference, it can limit the operator’s ability to resolve small-scale structures, sharp interfaces, and localized discontinuities. Even in operator learning frameworks that do not explicitly employ Fourier transforms, spectral bias arises from the smoothness-inducing nature of common neural network parameterizations and optimization schemes. The impact of spectral bias is further amplified in multiscale and multiphysics problems, where accurate solution representations require simultaneous resolution of widely separated spatial and temporal frequencies. In such settings, both PINNs and neural operators may converge to solutions that satisfy coarse-scale physics while systematically under-representing fine-scale dynamics.\\

Recognizing the adverse impact of spectral bias on learning multiscale functions, a growing body of work has proposed spectral bias mitigation strategies across neural function approximation, physics-informed learning, and operator learning frameworks. For function approximation, early studies demonstrated that standard multilayer perceptron (MLP) trained with first-order gradient-based optimization exhibits a strong preference for low-frequency components. To counteract this effect, techniques such as Fourier feature embeddings \cite{tancik2020fourier}, multi-scale neural networks \cite{WANG2025107179}, and enhanced activation functions \cite{hong2022activation} have been introduced to explicitly enrich the representational capacity of neural networks at higher frequencies. In parallel, operator learning networks have adopted complementary approaches to mitigate spectral bias at the operator level. FNOs explicitly control spectral resolution through mode truncation and filtering, striking a balance between numerical stability and expressive power. Multi-scale DeepONet \cite{wang2025multi}, high frequency scaling \cite{KHODAKARAMI2026108027}, and integration of neural operators with generative AI \cite{oommen2025learning} are among the recent methods for neural operator spectral bias mitigation.\\

Within physics-informed neural networks, spectral bias presents additional challenges due to the coupling between PDE residual minimization and other loss terms (e.g., boundary conditions). Several studies have shown that standard PINN formulations tend to prioritize low-frequency residual reduction, leading to physically consistent yet spectrally incomplete solutions for problems such as high-Reynolds-number fluid flows or wave-like problems \cite{chai_10630853, sallam2023use}. To address this issue, various strategies have been proposed, including adaptive collocation point sampling based on residual magnitude \cite{WU2023115671}, spatially-adaptive Fourier feature encoding \cite{liu2025diminishing}, and separated-variable spectral neural networks \cite{xiong2025separated}. Architectural enhancements such as Fourier-featured PINNs \cite{sallam2023use}, exponential and sinusoidal input feature layers \cite{exp, sine}, and adaptive activation functions \cite{JAGTAP2020109136} have also been explored to improve high-frequency representation while preserving physical constraints.\\

In this work, we systematically investigate the role of neural architecture and optimization scheme in learning high-frequencies. Our aim is to move spectral bias characterization from an empirical phenomenon to a more analyzable one through proper quantification across different physics-informed and data-driven learning tasks. We show that spectral bias is not only representational but also dynamical, meaning that is strongly impacted by the training strategies and optimization procedure. While enhanced neural architectures can increase the representational power of the neural network, proper optimization is necessary to change the dynamics of spectral learning during the training. This is particularly important for high-dimensional neural operator learning and physics-informed learning with multiple coupled loss components. Specifically, we compare the performance of a first-order optimizer (e.g., Adam), with a quasi second-order optimizer (SOAP) and a second-order optimizer (self-scaled Broyden (SS-Broyden)). The following are the specific questions we pose and answer in this paper:

\begin{itemize}
    \item What is the role of optimizer in spectral bias mitigation? 
    \item What is the role of network architecture (e.g., MLP vs KAN) in spectral bias mitigation in physics-informed networks?
    \item What is the role of activation functions in PINNs and univariate functions in PIKANs in spectral bias mitigation?
    \item Can spectral bias be mitigated in the same manner for dispersive, hyperbolic, and elliptic equations?
    \item How is spectral bias manifested by first, second, third, and fourth moments?
    \item For neural operators, how different neural architectures (e.g., DeepONet, DeepOKAN, FNO, and CNO) and different loss functions (e.g., MSE and binned spectral loss \cite{chakraborty2025binned}), as well as optimizers are compared in solving problems with high-frequencies.
\end{itemize}

The paper is organized as follows: In section \ref{spectral_bias_metric}, spectral bias of neural networks, theoretical analysis on the role of activation function and optimization scheme, and the proposed metrics to investigate spectral bias quantitatively are described. In section \ref{methods}, the representational methods including PINN, PIKAN, and neural operators are described. In section \ref{results}, we show the results of effect of optimization, neural architecture, and activation functions in spectral bias of PINNs and PIKANs solving multiple PDEs, followed by the results of data-driven neural operator learning settings for turbulent jet reconstruction and earthquake dynamic prediction problems. In section \ref{summary}, we summarize our results and findings.

\section{Spectral bias: Theoretical analysis and metrics}
\label{spectral_bias_metric}

\noindent Let $\Omega$ be a bounded domain and $u: \Omega \to \mathbb{R}$, $u(x)$ can be represented as follows:
\begin{equation}
    u(x) = \sum_{k=1}^{\infty} u_k \, \phi_k(x),
\end{equation}

\noindent where $\phi_k(x)$ are the orthonormal basis of $L^2(\Omega)$ ordered by increasing spatial frequency. The error between the ground truth solution and the neural network solution ($u_{\theta}$) can be written as follows:
\begin{equation}
    e(x) = u_{\theta} (x) - u(x) = \sum_{k=1}^{\infty} (e_k) \, \phi_k(x),
\end{equation}

\noindent where $e_k = u_{\theta,k} - u_k$. Assuming that basis are obtained through Fourier series expansion, $e_k$ will be the Fourier coefficient of the error at frequency $k$. A neural PDE solver exhibits spectral bias if the training error associated with low-frequency modes converges significantly faster than the error of high-frequency modes. Note that by using Parseval's theorem, the common $L^2$ neural training loss can be written in form of Fourier coefficients:
\begin{equation}
    \|u - u_\theta\|_{L^2(\Omega)}^2 = \sum_{k \in \mathbb{Z}^d} \big| \hat{u}(k) - \hat{u}_\theta(k) \big|^2
= \sum_{k \in \mathbb{Z}^d} |\hat{e}_k|^2,
\end{equation}

\noindent where $\hat{u}(k)$ and $\hat{u}_{\theta}(k)$ are the Fourier representation of the ground truth and neural network solutions. Since for most physical systems $|\hat{e}_k| > |\hat{e}_{k^*}|$ if $k < k^*$ at the start of the training, then the low-frequency modes have larger energies and contribute more to the total $L^2$ loss. Therefore, the optimizer of the neural network tends to learn low-frequency modes first, and higher modes at later stages of training if learned at all. In fact, using neural tangent kernel (NTK) theory \cite{jacot2018neural}, it can be shown that the learning dynamics for each mode ($k$) is proportional to $|\hat{e}_k|$ \cite{chakraborty2025binned}.

\subsection{Effect of activation function on spectral bias}

\noindent In this work, we study the impact of activation function, neural representation, and optimization scheme separately as well as their cross-correlated combined effect on the neural network spectral bias. Activation functions in neural networks have significant impact on the training dynamics and thereby on the spectral bias of the neural network. Consider a one hidden layer neural network in one dimension where each neuron contributes a ridge function. The Fourier transform of a single ridge function is shown below:
\begin{equation}
    \widehat{\sigma(w.x+b)}(k) = \frac{1}{|w|}e^{\frac{ibk}{|w|}}\hat{\sigma}(\frac{k}{|w|}),
\end{equation}

\noindent where $w$, and $b$ are the learnable weights and biases, $\hat{\sigma}$ is the Fourier transformation of the activation function ($\sigma$). The hyperbolic tangent (Tanh) activation function has an exponentially decaying Fourier transform.
\begin{equation}
    |\widehat{Tanh(k)}| \sim \frac{\pi}{sinh(\frac{\pi.k}{2})} = \frac{2}{e^{\frac{\pi.k}{2}} - e^{\frac{-\pi.k}{2}}},
\end{equation}

\noindent Thus, for higher frequencies as $k \to \infty$, $\widehat{Tanh(\frac{k}{|w|})} \sim 2\pi e^{\frac{-\pi.k}{2|w|}}$. This demonstrates that high frequency components are exponentially suppressed and large weights $|w|$ are required to represent oscillatory features. This explains why Tanh-based PINNs struggle with sharp gradients, high-frequency solutions, and multiscale PDEs. On the other hand, sine activations are used in the SIREN networks which have shown superior performance in recovering the high-frequencies and mitigate blurriness in computer vision tasks \cite{sitzmann2020implicit}. SIREN networks use the activation $\sigma(x) = sin(w_0 x)$, where $w_0 > 0$ is a fixed frequency scaling that is pre-determined separately for each layer of the network. The Fourier transform of a sine activation can be written as follows:
\begin{equation}
    \widehat{sin(k)} = \frac{\pi}{i}[\delta(k-w_0) - \delta(k+w_0)],
\end{equation}

\noindent where $\delta$ is the Dirac delta function. Note that the $|\widehat{sin(k)}|$ has no decay. Therefore, each neuron can contribute a sharply localized frequency, and varying the $w$ and $b$ (learnable parameters) shifts and modulates these frequencies without attenuation.

\subsection{Effect of optimization on spectral bias}
\noindent Additionally, we investigate the role of optimization scheme in spectral bias mitigation in physics-informed networks and neural operators for solving multiple PDEs and also data-driven learning in multi-scale problems. Here, we conduct a linearized analysis for the impact of first-order and second-order optimizations in the training dynamics for different frequencies. Let $f(x;\theta)$ be a neural network parameterized by $\theta \in \mathbb{R}^p$, trained to approximate a target function $f^{\star}(x)$ on a domain $\Omega$. Consider the commonly used $L^2$ loss as the objective function:

\begin{equation}
    \mathcal{L}(\theta)
    = \frac{1}{2} \int_{\Omega}
    \big( f(x;\theta) - f^{\ast}(x) \big)^2 \, dx.
    \label{eq:L2_loss}
\end{equation}

Let $\theta_0$ be a reference parameter vector (e.g., initialization of the network or a set of parameter along training). Linearizing the network around $\theta_0$ gives the following:

\begin{equation}
    f(x;\theta)
    \;\approx\;
    f(x;\theta_0)
    \;+\;
    J(x;\theta_0)\,(\theta - \theta_0),
\end{equation}

\noindent where $J(x;\theta_0) = \nabla_{\theta}f(x;\theta_0)$ is the Jacobian showing the sensitivity of the network outputs to the parameters. Substituting this linearized approximation into the Eq. \ref{eq:L2_loss}, and calculating the gradient of the loss function at $\theta = \theta_0$, we get:
\begin{equation}
    \nabla_{\theta}\mathcal{L} = \int_{\Omega}J^{T}(x;\theta_0) \ e(x) \, dx,
    \label{eq:grad_L}
\end{equation}

\noindent where $e(x) = f(x;\theta_0) - f^{\star}(x)$ is the neural network estimation error with parameters $\theta_0$. Under the first-order optimization scheme (e.g., gradient descent), the training dynamics becomes:
\begin{equation}
    \dot{\theta} = -\eta\nabla_{\theta}\mathcal{L} = -\eta\int_{\Omega}J^{T}(z;\theta_0) \ e(z) \,dz,
    \label{eq:theta_dot}
\end{equation}

\noindent where $\eta$ is the learning rate and $z$ is the dummy variable for integration. The training dynamics in parameter space can be transferred to function space using the chain rule $\dot{f}(x) = J(x) \ \dot{\theta}$. Therefore, using Eq. \ref{eq:theta_dot} the induced dynamics in function space can be written as:
\begin{equation}
    \dot{f}(x) = -\eta \int_{\Omega}J(x) \cdot J^{T}(z) \ e(z)  \, dz = -\eta \int_{\Omega}K(x,z) \ e(z) \, dz,
    \label{eq:f_dot}
\end{equation}

\noindent where $K(x,z)$ is the neural tangent kernel (NTK)~\cite{jacot2018neural}. The neural network estimation error at training time $t$ can be expanded in the basis $\phi_k$ with coefficients $e_k(t)$:
\begin{equation}
    e(z,t) = \sum_k e_k(t) \ \phi_k(z).
    \label{eq:e_z_t}
\end{equation}

\noindent Substituting Eq. \ref{eq:e_z_t} into Eq. \ref{eq:f_dot}, we get:
\begin{equation}
    \dot{f}(x) = -\eta \sum_k e_k(t) \int_{\Omega}K(x,z) \ \phi_k(z) \,dz.
    \label{eq: f_dot2}
\end{equation}

\noindent By using the definition of eigenfunction, Eq. \ref{eq: f_dot2} can be written as a function of eigenvalues ($\lambda_k$).

\begin{equation}
    \int_{\Omega} K(x,z) \ \phi_k(z) \,dz = \lambda_k \ \phi_k(x)
\end{equation}

\begin{equation}
    \dot{f}(x) = -\eta \sum_k \lambda_k \ \phi_k(x) \ e_k(t).
    \label{eq:f_dot_final}
\end{equation}

\noindent Based on the definition of $e(x) = f(x) - f^{\star}(x)$, the learning dynamics at training time $t$ in function space can be also written as:
\begin{equation}
    \dot{f}(x) = \sum_k \dot{e}_k(t) \ \phi_k(x).
    \label{eq:f_dot_def}
\end{equation}

\noindent Using Eqs. \ref{eq:f_dot_final} and \ref{eq:f_dot_def}, we obtain the decoupled dynamics for the estimation error:
\begin{equation}
    \dot{e}_k(t) = -\eta \ \lambda_k \ e_k(t).
\end{equation}

\noindent Thus, convergence rates (error decays) are proportional to $\lambda_k$. For typical activations (e.g., Tanh, ReLU), $\lambda_k$ decays rapidly with frequency, causing spectral bias.\\

Next, we show that such a decay with frequency does not exist for second-order optimization (e.g. SS-Broyden). In a second-order optimization scheme, the training dynamics for network parameters becomes:
\begin{equation}
    \dot{\theta} = -H^{-1} \ \nabla_{\theta}(\mathcal{L}),
\end{equation}

\noindent where $H = \int_{\Omega} J^{T}(z) \ J(z) \, dz$ is the Hessian matrix. Therefore, the induced dynamics in function space is modified from Eq. \ref{eq:f_dot} by incorporating the Hessian: 
\begin{equation}
    \dot{f}(x) = -J(x) [\int_{\Omega} J^{T}(z) \ J(z) \, dz]^{-1} \ \int_{\Omega} J^{T}(z) \ e(z) \, dz
    \label{eq:second_order_f_dot}
\end{equation}

\noindent For simplicity, we consider the vector form of Eq. \ref{eq:second_order_f_dot}.
\begin{equation}
    \dot{f} = -P \cdot e,
    \label{eq:Pr}
\end{equation}

\noindent where $P = -J(J^T J)^{-1}J^{T}$ is the projection matrix onto the range of the Jacobian ($J$). This structure is analogous to the normal equations in linear regression, where $(J^TJ)^{-1}J^T$ represents the Moore-Penrose pseudoinverse \cite{seber2003linear}. Given an over-parameterized neural network which is almost always the case in deep learning, we can assume that the target function ($\dot{f}$) is realizable by the network. This implies that the residual vector ($e$) exists within the column space of $J$, meaning that the orthogonal component of the projection is zero. Consequently,
\begin{equation}
    P \cdot e \approx e \rightarrow \dot{f} \approx -e.
    \label{eq:f_dot_e}
\end{equation}

\noindent Projecting Eq. \ref{eq:f_dot_e} into eigenbasis ${\phi_k}$ yields:
\begin{equation}
    \dot{e}_k (t) = -e_k(t).
\end{equation}

\noindent This demonstrates that the convergence rate for each frequency is independent of $\lambda_k$ and all modes converge at comparable rates. 

It is important to investigate the coupled effect of activation function and optimization scheme on spectral bias. The activation function influences the smoothness and decay of NTK spectrum ($\lambda_k$) and gradient magnitudes associated with high-frequency modes. It can also affect the conditioning of the Hessian in second-order optimization methods. Under first-order optimization methods, these effects directly translate into frequency-dependent convergence rates and play a direct role in spectral bias. Under second-order optimization methods, the approximated curvature information can mitigate this dependence, explaining why second-order methods are less sensitive to the choice of activation function. This analysis explains why:
\begin{itemize}
    \item Spectral bias is prominent with first-order optimizers.
    \item SIREN neural networks outperform networks with Tanh activation under first-order optimization for high-frequency problems.
    \item The effect of activation function is less significant with quasi-second order optimizer (e.g., SOAP), and is very small with second-order optimizer (e.g., SS-Broyden).
\end{itemize}

Our analysis and experiments suggest that spectral bias is primarily an optimization-induced phenomenon rather than a representational problem, although the inductive biases such as proper choice of activation function for learning high-frequencies in the representational model can help with the mitigation of spectral bias. Spectral bias is mainly caused from ill-conditioned training dynamics. Second-order optimization effectively preconditions these dynamics, equalizing convergence rates across all frequency models and substantially helps with spectral bias mitigation. This theoretical analysis is supported by empirical results in section \ref{results}.

\subsection{Spectral bias metrics}
\noindent When studying neural networks spectral bias, it is important to consider appropriate metrics for visualizing and quantifying it. Intuitively, comparing the results in frequency domain reveal more information compared to physical space. Thus, the following unified metric can be used to quantify the error at each wavenumber of the predicted solution, its gradient, and Laplacian:

\begin{equation}
    e_{\mathcal{F},p} = \frac{1}{N_x^2} \sum_{k} |k|^p |\hat{e}_k|^2
\end{equation}
where $N_x$ is the grid size, $k$ is the frequency in physical space, and the error $L^2$, gradient, and Laplacian errors in frequency domain correspond to $p = 0$, $p = 2$, and $p = 4$, respectively. Note that the errors due to spectral bias generally happen around regions with high-frequency features and sharp gradients. Hence, the prediction errors will be magnified in the gradient and Laplacian of the solutions rather than the solution itself.\\

Another potential metric for analysis of spectral bias is to measure the amount of oscillations within the solution and the predictions. The Barron norm measures how much a function oscillates and is calculated as the average of the norm of the frequency vector weighted by the Fourier magnitude \cite{barron2002universal}. Assume $\hat{f}(\omega)$ is the Fourier representation of function $f:\mathbb{R}^d \rightarrow \mathbb{R}$. The Barron norm of $f$ is defined as follows, provided that the integral is finite: 
\begin{equation}
\| f \|_{\mathcal{B}}
\;:=\;
\int_{\mathbb{R}^d} \|\omega\|_2 \, |\hat{f}(\omega)| \, d\omega,
\label{eq:Barron_norm}
\end{equation}

\noindent where $\omega$ is the wavenumber. We employ the Barron norm to quantify the oscillatory content of the ground-truth solution and the corresponding physics-informed network and neural operator predictions, thereby providing a measure of spectral bias in the learned solutions. Since the Barron norm weights the Fourier spectrum by frequency, functions with smaller Barron norms are generally easier for neural networks to approximate. Consequently, an underestimated Barron norm in the model predictions indicates a loss of high-frequency content and increased spectral bias.\\

Additionally, looking into the first four statistical moments can provide insight into the failure modes of the predictions and connections to the spectral bias. The first two moments are the mean and variance, respectively, and are dominated by low-frequency modes in most physical systems. The third moment shows the skewness in the distribution, making it sensitive to higher modes. The fourth moment (Kurtosis) measures the peakedness in the distribution and is associated with localized and sharp features. These measures for a time-dependent problem are defined in appendix \ref{App:Moments}. \\

\section{Methods}
\label{methods}
\subsection{PINN and PIKAN} 


\noindent In the general form, consider a physical system governed by a set of PDEs defined over a spatiotemporal domain $\Omega \times \mathcal{T}$, where $\mathbf{x} \in \Omega \subset \mathbb{R}^d$ denotes the spatial coordinates and $t \in \mathcal{T}$ denotes time. The governing equations are written in the general form as follows:

\begin{equation}
\mathcal{N}\big(u(\mathbf{x},t)\big) = f(\mathbf{x},t),
\quad (\mathbf{x},t) \in \Omega \times \mathcal{T},
\end{equation}
\begin{equation}
\mathcal{B}\big(u(\mathbf{x},t)\big) = g(\mathbf{x},t),
\quad (\mathbf{x},t) \in \partial \Omega \times \mathcal{T},
\end{equation}
\begin{equation}
u(\mathbf{x},0) = u_0(\mathbf{x}),
\quad \mathbf{x} \in \Omega,
\end{equation}

\noindent where $\mathcal{N}(\cdot)$ denotes a differential operator defined in $\mathbf{x} \in \Omega$, and $\mathcal{B}(\cdot)$ represents the boundary operators defined in $\mathbf{x} \in \partial\Omega$.\\

Physics-informed learning incorporates the governing equations, boundary conditions, and initial conditions (if any) directly into the training process. Let $\mathbf{u}_\theta (x,t)$ denote a neural network parameterized by $\theta$. The total physics-informed loss function is constructed as a weighted sum of multiple components as shown in Eq. \ref{eq:loss}: 
\begin{equation}
\mathcal{L}(\theta)
=
\lambda_{pde} \mathcal{L}_{\mathrm{PDE}}
+
\lambda_b \mathcal{L}_{\mathrm{BC}}
+
\lambda_{ic} \mathcal{L}_{\mathrm{IC}}
+
\lambda_{\mathrm{data}} \mathcal{L}_{\mathrm{data}},
\label{eq:loss}
\end{equation}
where $\lambda_i$ is the weights for $i^{th}$ loss component. Note that in this study no sensor data is used for training the networks and therefore $L_{data}$ is discarded. $L_{pde}$ is the governing PDE residual, $L_{BC}$ is the error at boundary locations, and $L_{IC}$ is the error for the initial condition, each calculated on the corresponding collocation points as shown in the following equations. 

\begin{equation}
    \mathcal{L}_{\mathrm{PDE}} = \frac{1}{N_{pde}} \sum_{i=1}^{N_{pde}} \left\| \mathcal{N}[\mathbf{u}_\theta](x_f^i, t_f^i) - f(x_f^i, t_f^i) \right\|^2
\end{equation}

\begin{equation}
    \mathcal{L}_{\mathrm{BC}} = \frac{1}{N_b} \sum_{i=1}^{N_b} \left\| \mathcal{B}[\mathbf{u}_\theta](x_b^i, t_b^i) - g(x_b^i, t_b^i) \right\|^2
\end{equation}

\begin{equation}
    \mathcal{L}_{\mathrm{IC}} = \frac{1}{N_i} \sum_{i=1}^{N_i} \left\| \mathbf{u}_\theta(x_i^i, 0) - u_0(x_i^i) \right\|^2
\end{equation}

\noindent In the PINN framework, the solution is represented by a feedforward neural network composed of fully connected layers with non-linear activation functions. The parameters include weights ($W_i$) and biases ($b_i$) as shown in Eq. \ref{eq: MLP} for a neural network with $L$ layers. While usually fixed, it is possible to make the activation functions ($\sigma_i$) adaptive with learnable parameters varying across each layer or even across neurons to accelerate the convergence in PINNs \cite{JAGTAP2020109136}.
\begin{equation}
    u_{\theta}(x)_{NN} = \sigma_L \left( W_L \left( \sigma_{L-1} \left( \dots \sigma_1(W_1 x + b_1) \dots \right) \right) + b_L \right)
    \label{eq: MLP}
\end{equation}

\noindent In the physics-informed Kolmogorov-Arnold network (PIKAN) framework, a KAN structure is used as the function approximator instead of a MLP. However, the mathematical integration of physical losses remains similar to the PINN. In KAN, univariate functions with learnable coefficients are used in the edges while only summation operation is done on nodes. Therefore, similar to MLP, a KAN can be formulated as shown in Eq. \ref{eq: KAN} \cite{SHUKLA2024117290}: 

\begin{equation}
    \resizebox{\linewidth}{!}{$
    u_{\theta}(\mathbf{x})_{\mathrm{KAN}} =
    \sum_{i_{L-1}=1}^{n_{L-1}}
    \phi_{L-1, i_L, i_{L-1}} \Bigg(
    \sum_{i_{L-2}=1}^{n_{L-2}} \cdots
    \Big(
    \sum_{i_2=1}^{n_2}
    \phi_{2, i_3, i_2} \Big(
    \sum_{i_1=1}^{n_1}
    \phi_{1, i_2, i_1} \Big(
    \sum_{i_0=1}^{n_0}
    \phi_{0, i_1, i_0}(x_{i_0})
    \Big)\Big)\Big)\Bigg)
    $}
    \label{eq: KAN}
\end{equation}

\noindent where $L$ is the number of layers, $n_j$ is the number of neurons in the \textit{j}th layer, and $\phi_{k,j,i}$ are the univariate activation functions. In this work, we explore B-Splines, radial basis functions (RBFs), and Chebyshev polynomials as the univariate functions.\\

Building on the PIKAN formulation above, we also consider a stabilized Chebyshev-based variant, referred to as Tanh-cPIKAN, originally introduced in \cite{AHMADIDARYAKENARI}. In this architecture, Chebyshev polynomial expansions are augmented with additional nonlinear contractions using Tanh activation, which maps intermediate representations to a bounded interval. The Tanh nonlinearity is applied both prior to and after each Chebyshev expansion, resulting in a nested nonlinear--polynomial composition. For an $L$-layer network, the resulting approximation can be written as
\begin{equation}
\resizebox{\linewidth}{!}{$
u_{\theta}(\boldsymbol{x})_{\text{Tanh-cKAN}} = 
W \cdot \sigma \Bigg( 
\sum_{i_{L-1}=1}^{n_{L-1}} \sum_{d_L=0}^{D} c_{i_L, i_{L-1}, d_L}^{(L)} 
T_{d_L} \Big( 
\sigma \big( 
\cdots 
\sum_{i_0=1}^{n_0} \sum_{d_1=0}^{D} c_{i_1, i_0, d_1}^{(1)} 
T_{d_1}(\sigma(x_{i_0})) 
\cdots 
\big) 
\Big) 
\Bigg)
$}
\end{equation}
where $T_d(\cdot)$ denotes the Chebyshev polynomial of degree $d$, $D$ is the maximum polynomial degree, $\{n_\ell\}_{\ell=0}^{L}$ are the layer widths (with $n_0$ being the input dimension), $\sigma(\cdot)$ refers to the Tanh activation function, $x_{i_0}$ represents the $i_0$-th feature of the input vector $\boldsymbol{x}$, and the trainable parameters $\theta$ consist of the final linear weight matrix $W$ and the Chebyshev coefficients $\{c_{i_\ell, i_{\ell-1}, d_\ell}^{(\ell)}\}$, where each coefficient corresponds to the $d$-th degree polynomial on the edge connecting node $i_{\ell-1}$ of the previous layer to node $i_\ell$ of the current layer. The repeated application of Tanh enforces a contraction of intermediate activations, thereby limiting the growth of high-order polynomial contributions and reducing the tendency of polynomial expansions to amplify high-frequency components. In the final layer, a linear readout is applied to the Tanh-activated hidden representation, allowing the output scale to be adjusted via $W$ while preserving bounded intermediate features throughout the network. Beyond bounding intermediate activations, the Tanh nonlinearity directly
modifies the optimization geometry. Let $\mathbf{z}(\theta)$ denote the output
of a Chebyshev expansion, and define the contracted representation
\[
\tilde{\mathbf{z}} = Tanh(\mathbf{z}).
\]
By the chain rule, the Jacobian with respect to the parameters $\theta$ becomes
\[
\frac{\partial \tilde{\mathbf{z}}}{\partial \theta}
=
\operatorname{diag}\!\big(1 - Tanh^2(\mathbf{z})\big)
\frac{\partial \mathbf{z}}{\partial \theta}.
\]
Thus, Tanh inserts a diagonal contraction matrix with entries in $(0,1]$
into the Jacobian. For the squared residual loss $\mathcal{L}(\theta)=\|r(\theta)\|^2$,
the dominant curvature term is the Gauss--Newton matrix
$H_{GN}=J^\top J$, where $J=\partial r/\partial\theta$.
With Tanh activations, this becomes
\[
H_{GN}
=
J_0^\top
\operatorname{diag}\!\big(1 - Tanh^2(\mathbf{z})\big)^2
J_0,
\]
where $J_0$ denotes the Jacobian of the polynomial expansion without contraction. Therefore, Tanh rescales curvature directions and attenuates those associated with large-magnitude activations. In Chebyshev-based networks, high-degree polynomial interactions may amplify derivatives and induce sharp curvature in the loss landscape. The repeated contraction induced by Tanh mitigates this effect by limiting activation growth and reducing curvature anisotropy, thereby improving numerical stability during physics-informed training. For first-order optimizers such as Adam, this contraction leads to more balanced gradient magnitudes across parameters, reducing oscillations caused by high-frequency components and promoting more stable convergence.

\subsection{Neural operators}

\noindent Neural operators aim to directly learn mappings between spaces of functions, and is based on the universal operator approximation theorem by Chen and Chen \cite{chen1995universal}. 
Let \(\mathcal{U}\) and \(\mathcal{V}\) be Banach spaces of functions defined on \(\Omega\), and let the ground-truth operator be
\begin{equation}
    \mathcal{N}:\mathcal{U}\mapsto\mathcal{V}, 
    \qquad v = \mathcal{N}(u),
    \label{eq:true_operator}
\end{equation}
where \(u \in \mathcal{U}\) denotes the input function and \(v \in \mathcal{V}\) denotes the output function. 
Given a training set of paired samples \(\{(u_i,v_i)\}_{i=1}^{N}\), a neural operator \(\mathcal{G}_{\theta}\) is trained to approximate \(\mathcal{N}\) by minimizing a data misfit in function space, most commonly an \(L^2\) objective:
\begin{equation}
    \theta^{\star} =
    \arg\min_{\theta}
    \frac{1}{N}\sum_{i=1}^{N}
    \big\|
    \mathcal{G}_{\theta}(u_i) - v_i
    \big\|_{L^2(\Omega)} .
    \label{eq:no_l2_loss}
\end{equation}

We consider four representative neural operator families that differ in the way they represent the operator and propagate information across function space.
Deep Operator Network (DeepONet) \cite{lu2021learning} represents \(\mathcal{G}_{\theta}\) through a separable branch and trunk construction, in which the input function is encoded into coefficients by the branch network and combined with a coordinate-dependent basis learned by the trunk network. 
A standard form is
\begin{equation}
    \big(\mathcal{G}_{\theta}u\big)(\bm{x})
    =
    \sum_{j=1}^{p}
    b_{\theta_b,j}(u)\, t_{\theta_t,j}(\bm{x}),
    \label{eq:deeponet_form}
\end{equation}
where \(b_{\theta_b,j}\) are the branch outputs and \(t_{\theta_t,j}\) are the trunk outputs evaluated at \(\bm{x}\). 
DeepOKAN follows the same operator-learning principle but replaces the MLP components with KAN style parameterizations to increase expressivity in function approximation \cite{SHUKLA2024117290}. 
The Fourier Neural Operator (FNO) learns an operator by alternating pointwise mixing with global convolutions implemented in Fourier space \cite{li2021fourier}. 
In a typical layer,
\begin{equation}
    v^{\ell+1}(\bm{x})
    =
    \sigma\Big(
    W v^{\ell}(\bm{x})
    +
    \mathcal{F}^{-1}\big(
    R^{\ell}(k)\, \mathcal{F}(v^{\ell})(k)
    \big)(\bm{x})
    \Big),
    \label{eq:fno_layer}
\end{equation}
where \(\mathcal{F}\) denotes the Fourier transform, \(R^{\ell}\) is a learned spectral multiplier restricted to a finite set of modes, and \(\sigma\) is a nonlinearity. 
The Convolutional Neural Operator (CNO) learns the operator using multi-resolution convolutional blocks that combine local receptive fields with coarse-scale context, but unlike a standard UNet it is formulated as an operator on function spaces with resolution-invariant layers, so the same learned mapping can be applied across discretizations without relying on explicit Fourier truncation \cite{raonic2023convolutional}.
These distinct inductive biases suggest potentially different failure modes under spectral bias, especially for flows with slowly decaying spectra and sharp density-gradient features.\\

A central question in this work is how much spectral bias persists in neural operator surrogates across architectures, and whether it can be mitigated by modifying the standard \(L^2\) training objective in Eq. \ref{eq:no_l2_loss}. 
While operator architectures such as FNO explicitly use Fourier representations, the optimization is still typically driven by an \(L^2\) error that is dominated by low-frequency content, which can lead to underestimation of high-wavenumber power in the predictions. 
To investigate this, we evaluate each model using complementary diagnostics in physical and spectral domains, including field error, energy-spectrum error, and Barron-norm error. 
We also investigate an alternative to the plain \(L^2\) objective by augmenting training with a binned spectral power (BSP)  loss \cite{chakraborty2025binned}. 
We train operator surrogates both with the baseline objective in Eq. \ref{eq:no_l2_loss} and with BSP augmentation. Then, we compare how different architectures trade field accuracy for spectral fidelity in a setting where high-wavenumber content is essential.

\section{Results}
\label{results}
\noindent 
We begin by analyzing the spectral learning behavior in purely data-driven function approximation to isolate architectural effects (Section 4.1). We then examine the role of optimization in mitigating spectral bias in physics-informed networks (Section 4.2), followed by a detailed study of activation functions and representation models (Section 4.3). Finally, we assess the coupled effects of optimization and representation across hyperbolic and elliptic PDEs and extend the analysis to neural operator learning (Section 4.4).

\subsection{Effect of network architecture on data-driven high- and multi-frequency learning}
\label{app:data_driven_spectral}

\noindent In this section, we investigate how different neural network architectures influence spectral learning behavior in purely data-driven settings. The primary objective is to assess the ability of each architecture to capture high-frequency components, multi-scale spectral content, and sharp transitions when trained solely on observational data, without physics-informed constraints.\\

We consider several neural network architectures that span a range of representational and spectral biases. Specifically, we evaluate standard MLPs with Tanh activation functions (MLP--Tanh), MLPs equipped with sinusoidal activation functions (MLP--SIREN), Chebyshev Kolmogorov--Arnold Networks (cKAN), and Tanh- Chebyshev Kolmogorov--Arnold Networks (Tanh-cKAN), introduced in \cite{AHMADIDARYAKENARI}. All architectures are configured to have comparable depth and width to isolate the effect of representation rather than parameter count. The MLP-based models rely on fixed nonlinear activation functions, while the cKAN and Tanh-cKAN architectures employ adaptive Chebyshev polynomial expansions that enable localized spectral representations. The Tanh-cKAN variant further combines polynomial bases with nonlinear activation to enhance expressivity across frequency scales.\\



Two data-driven benchmark problems are considered, each designed to probe a different manifestation of spectral bias in neural network training.

\paragraph{Case 1: Discontinuous function with high-frequency content}
The first benchmark consists of a piecewise-defined target function with a jump discontinuity, defined as
\[
f(x) =
\begin{cases}
5 + \displaystyle\sum_{k=1}^{4} \sin(kx), & x < 0, \\
\cos(10x), & x \geq 0.
\end{cases}
\]
The presence of a discontinuity introduces broad-band high-frequency components in the Fourier domain, making this problem particularly challenging for neural networks that exhibit a low-frequency learning bias. For this case, $80$ uniformly spaced training samples are drawn from the interval $[-\pi, \pi]$.

\paragraph{Case 2: Smooth multi-frequency, multi-scale function}
The second benchmark considers a smooth target function composed of multiple sinusoidal components with well-separated frequencies and amplitudes:
\[
f(t) =
\sin(2\pi \cdot 0.01\, t)
+ 0.5\, \sin(2\pi \cdot 0.05\, t)
+ 0.2\, \sin(2\pi \cdot 0.1\, t).
\]

\noindent Although the function is smooth, the coexistence of multiple frequency scales makes it a canonical benchmark for evaluating a model’s ability to simultaneously capture low- and high-frequency modes without non-smoothness-induced artifacts. This function is sampled using $300$ uniformly spaced points over the temporal interval $[0, 300]$.

\begin{table}[H]
\centering
\caption{\textbf{Data-driven spectral benchmark errors.}
Comparison of neural network architectures for Case~1 (discontinuous function) and Case~2 (smooth multi-frequency function). Performance is evaluated using physical- and frequency-domain error metrics.}
\label{tab:data_driven_spectral_benchmark}
\setlength{\tabcolsep}{3pt}
\resizebox{\textwidth}{!}{%
\begin{tabular}{lccccc}
\hline
\textbf{Architecture} &
{\makecell{Rel. $L^2$\\ Error}} &
{\makecell{Barron Norm\\ Rel. Error}} &
\textbf{$\log(e_{\mathcal{F},p=0})$} &
\textbf{$\log(e_{\mathcal{F},p=2})$} &
\textbf{$\log(e_{\mathcal{F},p=4})$} \\
\hline
\multicolumn{1}{c}{\textbf{Case 1: Discontinuous Function}} \\
\hline
MLP--Tanh              &  $4.96 \times 10^{-4}$ & $2.35 \times 10^{-4}$ & $-3.72$ & $1.31$ &$6.38$ \\
\textbf{MLP--SIREN} & $\mathbf{4.68 \times 10^{-4}}$ & $\mathbf{6.48 \times 10^{-4}}$ & $\mathbf{-3.78}$ & $\mathbf{0.01}$ & $\mathbf{4.78}$ \\
cKAN                   & $3.67 \times 10^{-2}$ & $1.62 \times 10^{-2}$ & $0.01$ & $5.07$ & $10.17$ \\
Tanh-cKAN              & $5.19 \times 10^{-4}$ & $3.28 \times 10^{-4}$ & $-3.68$ & $-0.04$ & $4.86$  \\

\hline
\multicolumn{2}{c}{\textbf{Case 2: Smooth Multi-Frequency Function}} \\
\hline
MLP--Tanh              & $5.26 \times 10^{-2}$ & $6.94 \times 10^{-2}$ & $-0.27$ & $3.70$ & $8.06$\\

\textbf{MLP--SIREN} & $\mathbf{7.93 \times 10^{-4}}$ & $\mathbf{1.13 \times 10^{-3}}$ & $\mathbf{-3.91}$  & $\mathbf{-0.56}$  & $\mathbf{3.82}$ \\

cKAN                   & $2.61 \times 10^{-1}$ & $1.52 \times 10^{-1}$ & $1.123$ & $4.42$ & $8.45$\\
Tanh-cKAN              & $2.94 \times 10^{-2}$ & $1.97 \times 10^{-1}$ & $-0.78$ & $3.44$ & $7.98$\\

\hline
\end{tabular}
}
\end{table}

\begin{figure}[H]
    \centering
    \includegraphics[width=0.9\linewidth]{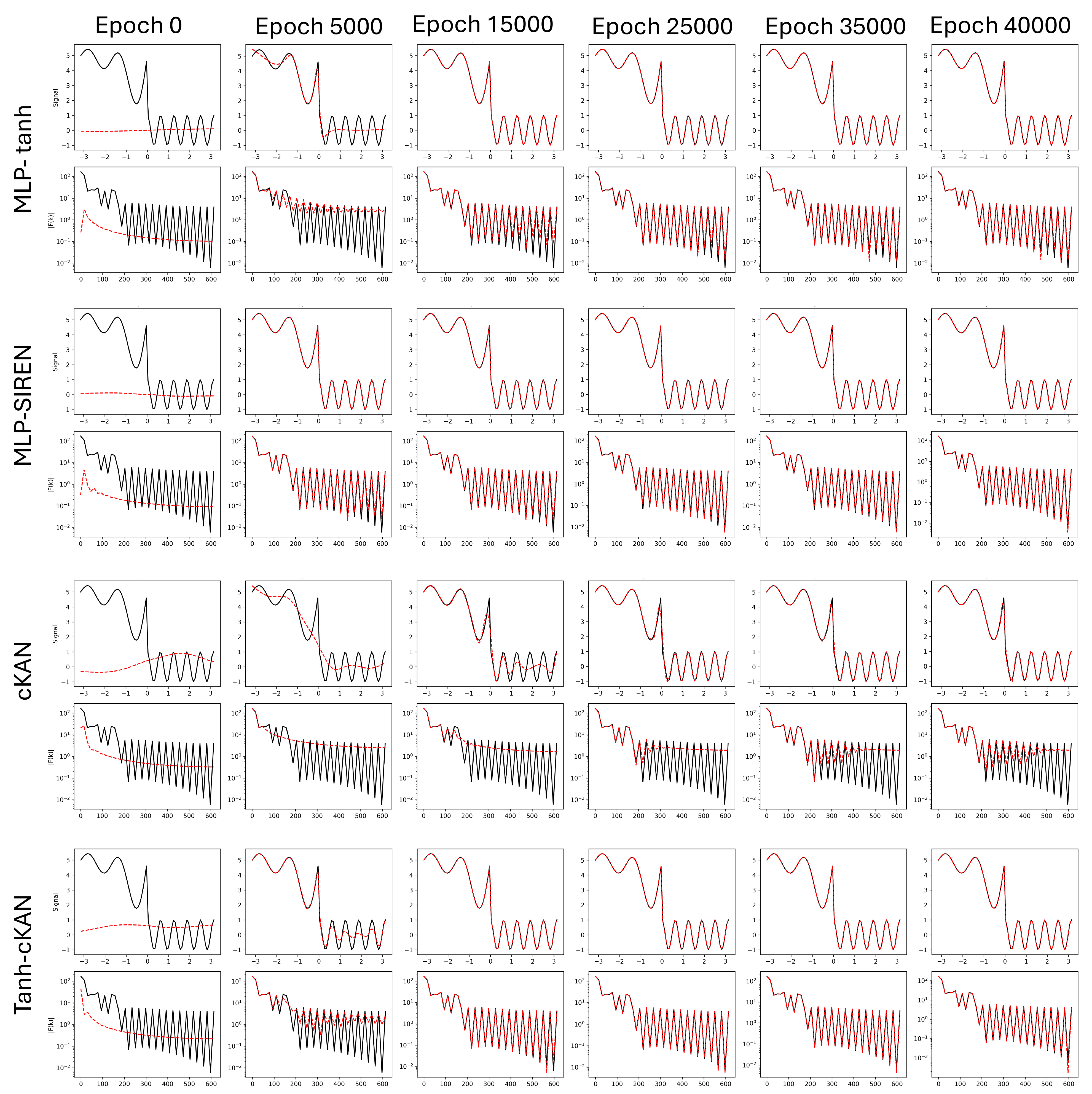}
    \caption{\textbf{Case 1: Training-time spectral evolution for the discontinuous benchmark.} Comparison of MLP--Tanh, MLP--SIREN, cKAN, and Tanh-cKAN on the piecewise-discontinuous target function. Columns correspond to different training epochs, illustrating the evolution of the predicted signal in physical space (top row of each block) and the corresponding Fourier amplitude spectrum (bottom row of each block). The ground-truth solution is shown in black, while network predictions are shown in red. The discontinuity induces broad-band high-frequency content in the frequency domain, highlighting differences in how each architecture and optimization strategy capture sharp transitions and recover high-frequency modes over training.}
    \label{fig:case1}
\end{figure}

\begin{figure}[H]
    \centering
    \includegraphics[width=0.9\linewidth]{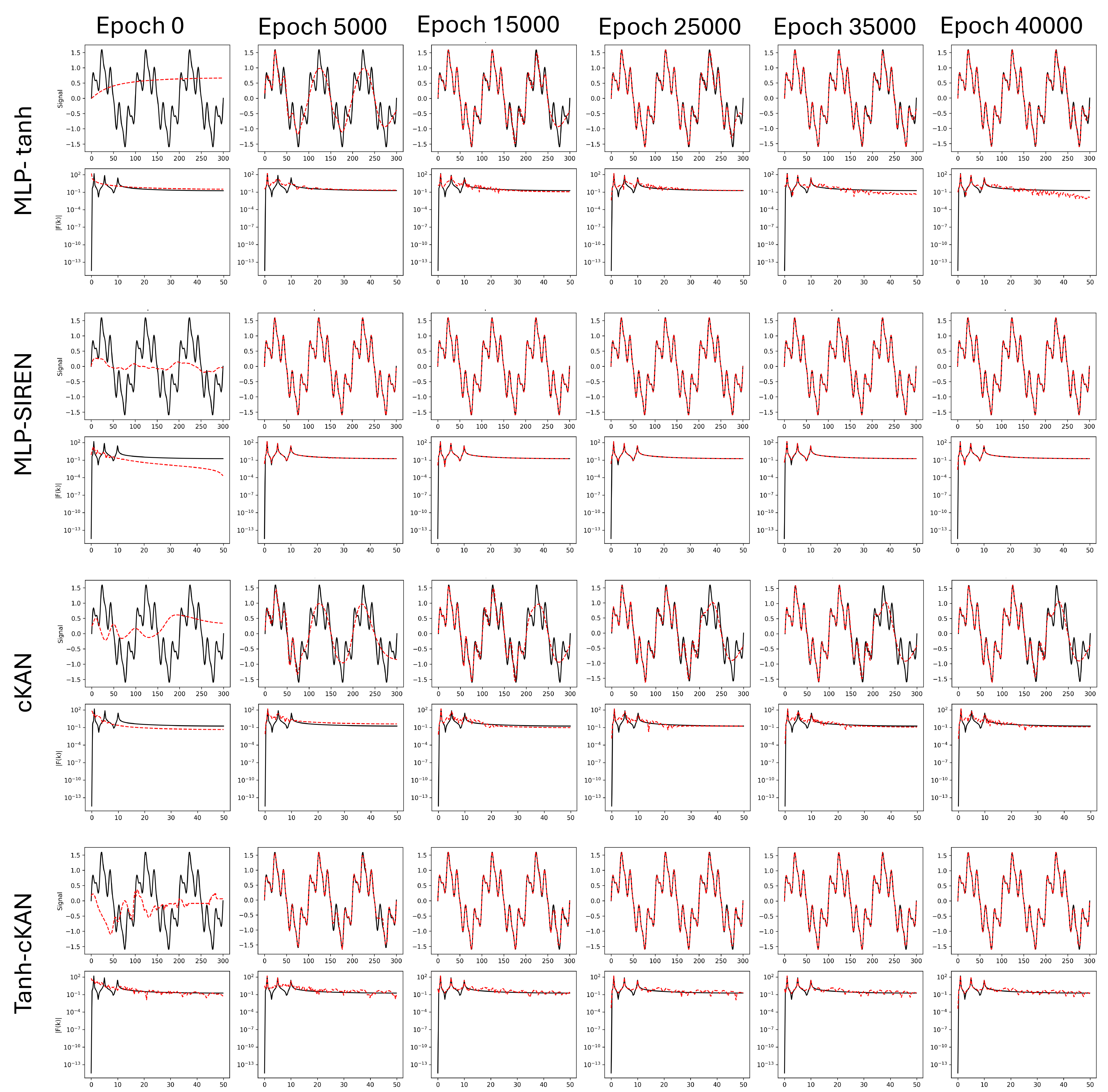}
    \caption{\textbf{Case 2: Training-time spectral evolution for the smooth multi-frequency benchmark.}
    Comparison of MLP--Tanh, MLP--SIREN, cKAN, and Tanh-cKAN on the multi-scale multi-frequency target function. Columns correspond to different training epochs to illustrate the evolution of the predicted signal in physical space (top row of each block) and the corresponding Fourier amplitude spectrum (bottom row of each block). The ground-truth solution is shown in black, while network predictions are shown in red. Although the target function is smooth, the presence of multiple well-separated frequency components makes this benchmark sensitive to spectral bias to reveal differences in how each architecture captures and balances low- and high-frequency modes during training.}
    \label{fig:case2}
\end{figure}

\noindent The quantitative results for both benchmarks are summarized in Table~\ref{tab:data_driven_spectral_benchmark}, while the training-time evolution in physical and frequency domains is illustrated in Figures~\ref{fig:case1} and~\ref{fig:case2}. For Case~1, which involves a discontinuous target function, MLP-based architectures with Tanh and SIREN activations achieve low relative $L^2$ errors; however, notable differences emerge in the frequency-domain metrics. In particular, the MLP--Tanh model exhibits large gradient- and Laplacian-weighted spectral errors. This indicates difficulty in accurately recovering high-frequency components induced by the discontinuity. By contrast, MLP--SIREN and Tanh-cKAN substantially reduce higher-order spectral errors, consistent with the improved recovery of high-frequency modes observed in the Fourier spectra in Fig.~\ref{fig:case1}. The cKAN model shows significantly larger errors across all spectral metrics. This suggests that polynomial bases alone, as configured here, are insufficient to capture sharp spectral transitions in the absence of additional nonlinear expressivity. For Case~2, which corresponds to a smooth but multi-frequency signal, the differences between architectures become more pronounced. The MLP--SIREN model achieves the lowest errors across both physical and spectral metrics. While the MLP--Tanh model captures the dominant low-frequency component, it suffers from substantial errors in higher-order spectral norms. This reflects incomplete learning of higher-frequency modes. The Tanh-cKAN architecture improves upon standard cKAN by reducing the physical-space error; however, its frequency-domain errors remain larger than those of MLP--SIREN, indicating residual spectral imbalance. These trends are clearly reflected in the frequency-domain plots in Fig.~\ref{fig:case2}, where SIREN-based models exhibit faster and more uniform recovery of all frequency components during training. The combined evidence from Table~\ref{tab:data_driven_spectral_benchmark} and Figs.~\ref{fig:case1}-\ref{fig:case2} highlights that architectural choices play an important role in mitigating spectral bias in data-driven settings.

\subsection{Effect of optimization on spectral bias in PINNs and PIKANs}

\noindent Recent studies have shown promising results in solving forward problems with PINNs optimized with second-order optimizers \cite{kiyani2025optimizing,AHMADIDARYAKENARI, wang2025gradient}. During the optimization of the neural network, it is important to determine the optimal direction and step size for each parameter update. Commonly used gradient descent method only uses the first-order derivative information for the update direction (negative gradient direction). However, this may not be sufficient for multi-objective optimizations such as those occurring in physics-informed network training. In these cases, the network gets stuck in a local minima and an early flattened training loss history is observed. A common practice in training PINNs is to start with Adam optimizer and switch to low-memory BFGS (L-BFGS) optimizer in later iterations for faster convergence. Both of these optimizers are from a family of line search optimization methods \cite{nocedal2006numerical}, where the parameters are iteratively updated as follows:

\begin{equation}
    \theta_{k+1} = \theta_k + \alpha_k p_k,
\end{equation}

\noindent where $\theta$ is the network parameters, $\alpha$ is the step size and $p_k$ is the direction at step $k$. In gradient descent method, $p_k = -\nabla L(\theta)$. Another example is Newton's method with quadratic rate of convergence near the solution. In Newton's method, both first order and second order derivatives are used in determining the optimal update direction, $p_k = -\mathcal{H}_k^{-1} \nabla L(\theta_k)$, where $\mathcal{H}$ is the Hessian matrix. Calculation of inverse Hessian can become computationally intractable for high-dimensional problems in deep learning and PINNs. As a result, quasi-Newton methods have emerged as alternatives to approximate the inverse Hessian. Example of such optimizers are BFGS and self-scaled Broyden (SS-Broyden) \cite{URBAN2025113656, AHMADIDARYAKENARI, kiyani2025optimizing, toscano2025variational}.\\

Consider the one-dimensional Korteweg--de Vries (KdV) equation that models the propagation of waves in nonlinear dispersive media:
\begin{equation}
u_t + \eta\,u\,u_x + \frac{\mu}{2}u_{xxx} = 0, \quad t \in (0,1), \; x \in (-1,1),
\end{equation}
subject to the initial condition
\begin{equation}
u(x,0) = \cos(\pi x),
\end{equation}
and periodic boundary conditions
\begin{equation}
u(t,-1) = u(t,1).
\end{equation}
Here, \(u\) denotes the wave amplitude (or free-surface elevation), \(\eta\) characterizes the strength of the nonlinear interaction, and \(\mu\) determines the level of dispersion. In this study, we used the classical values of \(\eta\) = 1, and \(\mu\) = 0.022 \cite{zabusky1965interaction}. Our results demonstrate that the choice of optimizer plays a dominant role in the performance of PINNs, particularly in mitigating spectral bias and enabling the learning of high-frequency solution components. We compare four optimization strategies of Adam, L-BFGS, SOAP, and SS-Broyden for training the PINN in solving the forward KdV equation.\\

The first-order Adam optimizer consistently exhibits early stagnation during training, characterized by flattened loss histories and entrapment in local minima. While Adam is effective at rapidly reducing the loss during the initial training stages, it struggles to further minimize the residual once higher-frequency components become dominant. This behavior results in limited accuracy and poor convergence, especially for problems with stiff or highly oscillatory dynamics. Adam weak performance can be enhanced by switching to L-BFGS after adequate warm-up iterations with Adam. In contrast, the quasi-second-order SOAP optimizer substantially improves convergence behavior and final accuracy without requiring any warm-up steps. SOAP effectively handles the coupled loss components from the very beginning of training, leading to stable convergence and approximately three and two orders of magnitude improvement in error compared to Adam- and L-BFGS- based training, respectively. A second-order optimizer such as SS-Broyden can significantly improves the convergence and accuracy. This demonstrates that curvature-aware optimization can significantly alleviate spectral bias without relying on carefully staged optimization schedules. Note that all the SS-Broyden results in this work are based on Wolfe line search \cite{kiyani2025optimizing}.\\

\begin{figure}[H]
    \centering
    \includegraphics[width=0.9\linewidth]{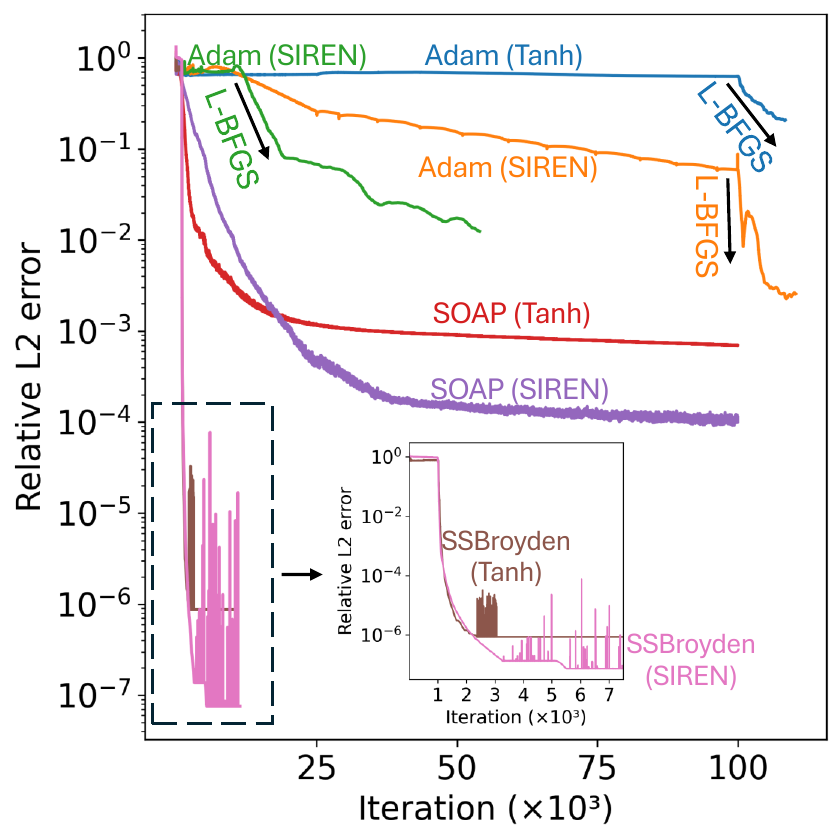}
    \caption{\textbf{KdV equation:} PINN training history with different optimizers (Adam, L-BFGS, SOAP, and SS-Broyden) and activation functions (Tanh, and SIREN) for KdV equation. The sub-figure shows the zoomed-in training history for the SS-Broyden optimizer.}
    \label{fig:Loss_history_KdV}
        
\end{figure}

Among all tested optimizers, SS-Broyden consistently achieves the best performance. By more accurately approximating curvature dynamics, SS-Broyden attains approximately three and five of magnitude lower error compared to SOAP and Adam, respectively, while requiring similar or lower computational time. Although each SS-Broyden iteration incurs a higher computational cost due to the estimation of second-order information, the dramatic reduction loss at each iteration leads to superior overall efficiency and accuracy. These results highlight the critical importance of second-order optimization strategies for training PINNs in problems with high-frequencies. Figure \ref{fig:Loss_history_KdV} shows the training history of the PINN with different optimizers, demonstrating how the second-order and quasi-second order optimizer losses start decaying right after the beginning of the training while Adam is stuck. 

\begin{figure}[H]
    \centering
    \includegraphics[width=1.0\linewidth]{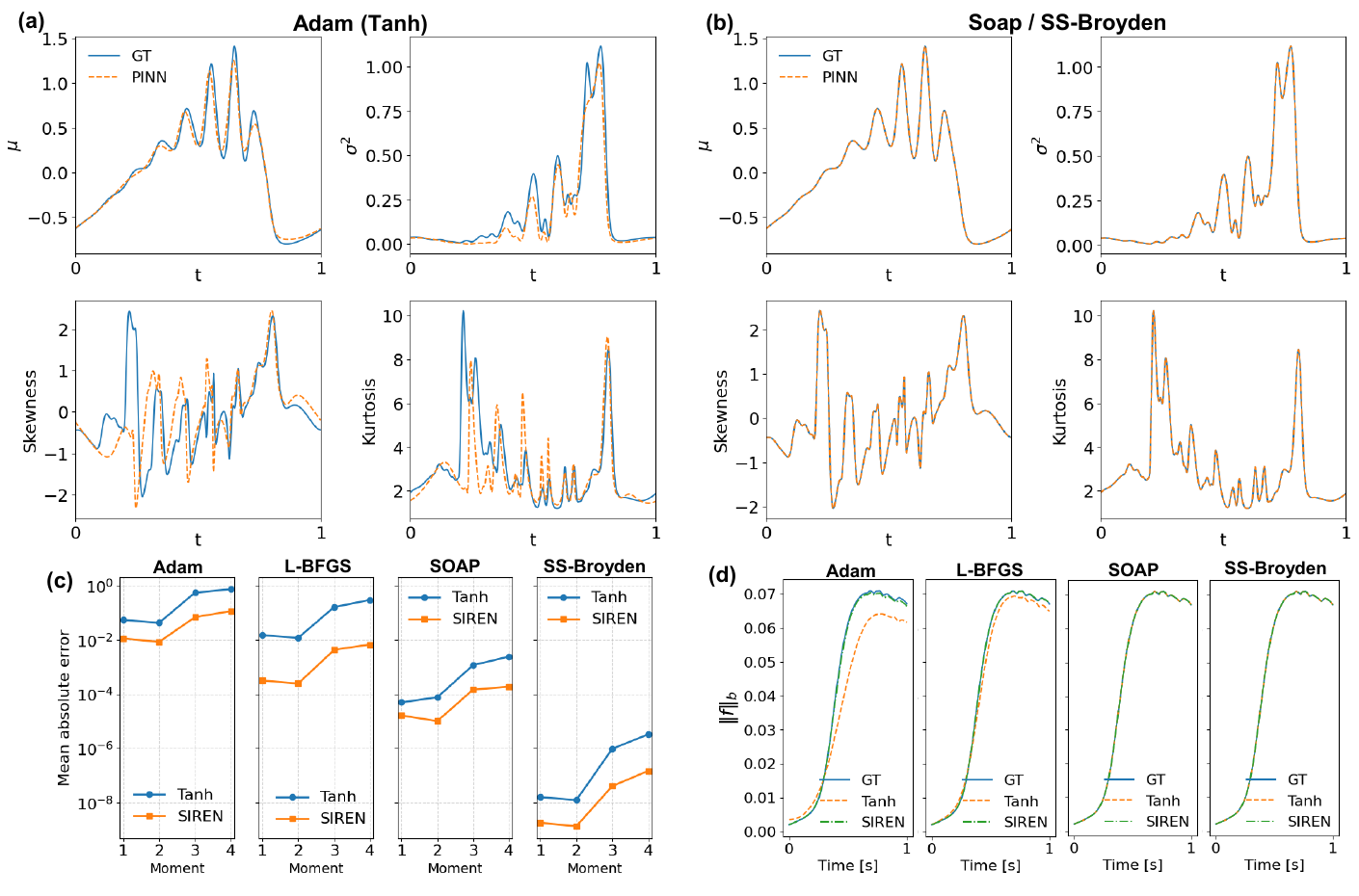}
    \caption{\textbf{KdV equation: Analysis of the predictions.} (a, b) First four moments (mean, variance, skewness, and kurtosis) of the PINN predictions with (a) Adam optimizer and Tanh activation, and (b) Soap or SS-Broyden optimizer with Tanh activations. (c) Time-averaged absolute error at each moment for PINN predictions with different optimizers and activations. (d) Barron norm at each time-step of PINN predictions with different optimizers.}
    \label{fig:KdV_analysis}
\end{figure}

\noindent When looking into the first four statistical moments of the PINN predictions, it is obvious that while Adam can approximately recover the first two moments, it completely fails to recover the third and fourth moments. On the other hand, SOAP and SS-Broyden recover all the four moments accurately with SS-Broyden being more accurate (Fig. \ref{fig:KdV_analysis}). The Barron norm which measures the amount of oscillations in the solutions also confirm the same findings (Fig. \ref{fig:KdV_analysis}d). The SOAP and SS-Broyden results follow the same norm as the ground truth solutions while Adam results demonstrate smaller Barron norm as the wave propagates in time. Note that L-BFGS improves the results of Adam but still shows smaller Barron norm, suggesting that the conventional optimizers systematically fail to capture high-frequencies in PINNs.

\subsection{Effect of activation functions and representation models on spectral bias}

\noindent We investigate how the activation functions in PINNs and univariate functions in PIKANs performs differently in solving PDEs with high-frequencies. It is important to investigate the effect of the neural architecture in conjunction with the optimizer. In PINN, we investigate how changing the activation function from Tanh to oscillatory and periodic Sine function can help with the capture of high-frequencies in the solution. Particularly, we used $sin(\omega_i \cdot W\mathbf{X} + b)$ as the activation function, where $W$ and $b$ are weight and bias in the network, and $\omega_i$ is the scaling factor for the $i^{th}$ layer of the network. Similar to the original SIREN neural network \cite{sitzmann2020implicit}, we found out that using $\omega_0$ = 30 and $\omega_i$ = 1 for the rest of the layers, provide the best performance while keeping the training stable. 

\begin{figure}[H]
    \centering
    \includegraphics[width=1.0\linewidth]{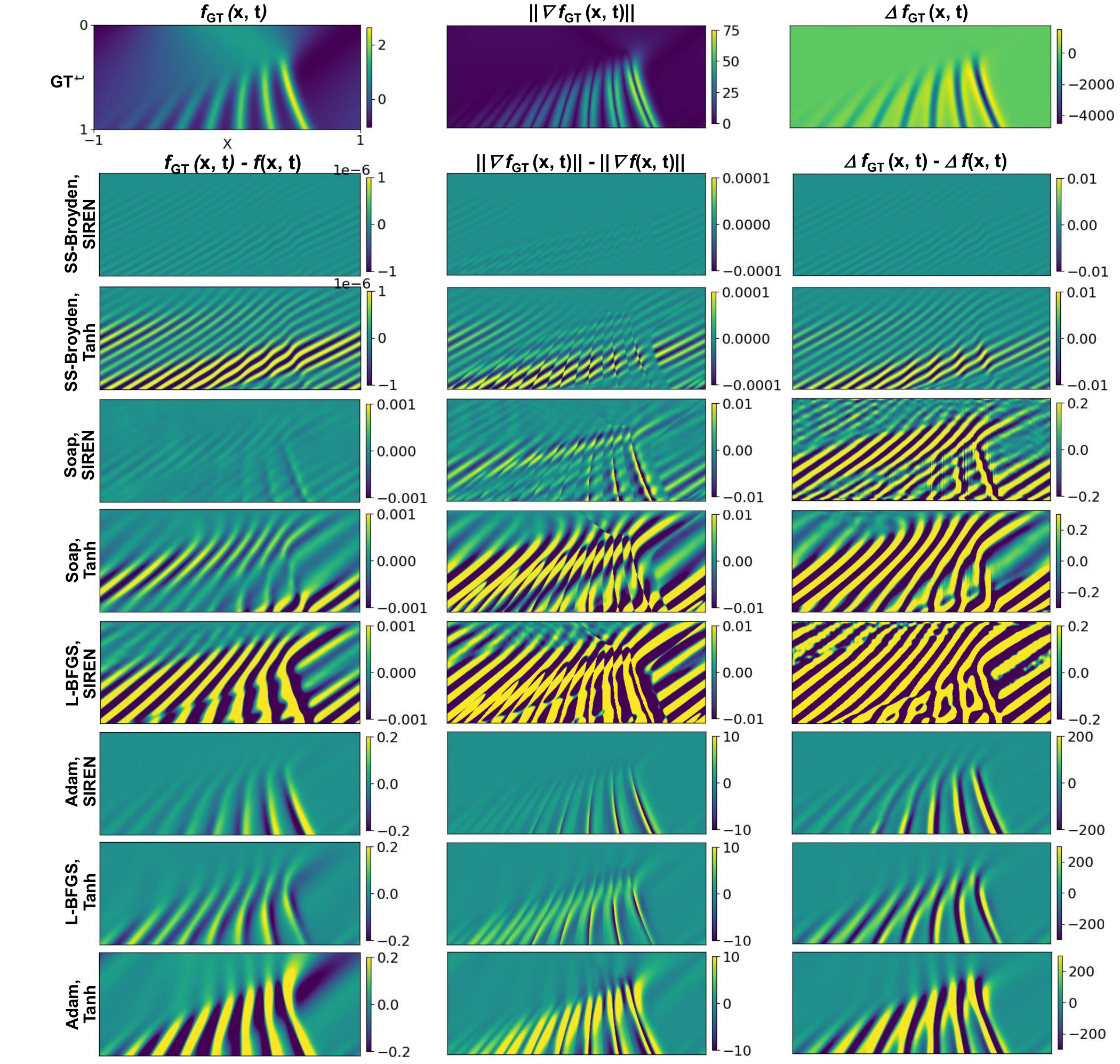}
    \caption{\textbf{KdV equation: Effect of optimizer and activation function.} First row: Ground truth (GT) solution, gradient of the GT solution, and Laplacian of the GT solution. The other rows in first, second and third columns show the errors in PINN solution, errors in gradient magnitude of the solution, and errors in Laplacian of the solution, respectively. The rows are organized from the most to the least accurate results (top to bottom). The axis ranges ($x \in [-1,1]$, $t \in [0,1]$) shown in top left subplot is applicable to all.}
    \label{fig:KdV_results}
\end{figure}

\noindent Fig. \ref{fig:KdV_results} demonstrates the KdV equation predictions of PINNs with different activation functions trained with different optimizers. It can be seen that both activation function and the optimizer are effective in spectral bias mitigation, with optimizer playing a more dominant role. Each optimizer and each activation function demonstrates a different dynamics during the training. For example, when using Adam optimizer, SIREN layers can help getting out of the local minima, breaking the flattened loss history curve (Fig. \ref{fig:Loss_history_KdV}). However, for quasi-second-order and second-order optimizers, SIREN provides more representational power with negligible additional computational cost that can be useful for capturing the high-frequencies. Also, note that the impact of activation function reduces with the use of second-order optimizers. Therefore, the impact is rather small for SS-Broyden compared to SOAP and Adam (Fig. \ref{fig:Loss_history_KdV}).\\

\begin{figure}[H]
    \centering
    \includegraphics[width=1.0\linewidth]{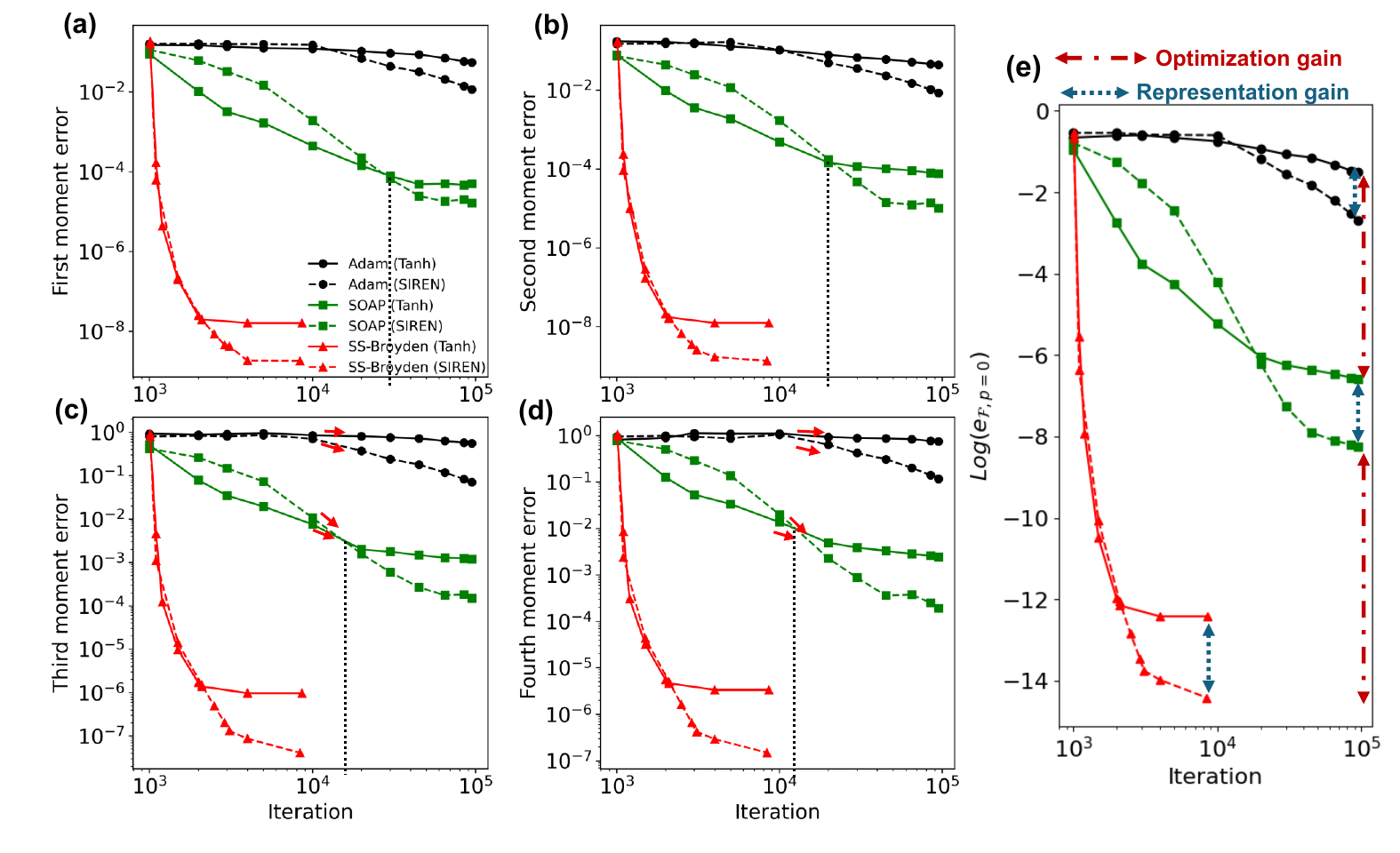}
    \caption{\textbf{KdV equation: Spectral bias dynamics during the training.} (a-d) Errors of first four statistical moments during the training with different optimizers and activation functions. (e) Error in the frequency domain during the training. The small red arrows show the approximate slope of learning the high-frequencies (shown in third and fourth moments). The dotted black lines in the four moments shows the iteration at which the Tanh and SIREN networks trained with SOAP intersect. Note how the intersection happens earlier at higher moments. The legends in (a) are applicable to all.}
    \label{fig:KdV_training_dynamics}
\end{figure}

For a more comprehensive analysis, we examined the learning dynamics and related them to spectral bias by tracking the prediction of the statistical moments during training. The results in Figure \ref{fig:KdV_training_dynamics} demonstrate that the slope of learning of the first two moments using Tanh with Adam is very small and only begins after approximately 10,000 iterations, and there is almost no learning for the third and fourth moments. When using SIREN with Adam, the slope of learning increases from a negligible value to a small but noticeable one. Using the SOAP optimizer, learning begins from the first iteration. Interestingly, SOAP with Tanh initially has faster error drop, however, SOAP with SIREN surpasses it in the middle of training showing larger learning slope in later iterations. This is potentially due to the higher high-frequency representational power of SIREN compared to Tanh accompanied with harder optimization with SIREN in the early stages which explains the faster drop with Tanh. Notably, the intersection point between the Tanh and SIREN with SOAP curves shifts to earlier iterations from the first to fourth moments. For example, the curves intersect around iteration 30,000 for the first moment and around 11,000 for the fourth moment. This results indicate that SIREN is more effective in capturing high-frequency features and is directly effective for spectral bias mitigation. When using SS-Broyden, the effect of the activation function on learning dynamics is reduced, although SIREN still provides greater representational capacity for high frequencies. As a result, the first four moments are learned at similar rates with Tanh or SIREN networks trained with SS-Broyden. However, at the final steps of the training, the Tanh network plateaus, while the SIREN one keeps decreasing for roughly another order of magnitude. Therefore, while activation function can significantly impact the high-frequency learning dynamics when optimized with Adam and SOAP, its impact is reduced when optimized with SS-Broyden. In this case, both networks follow similar learning trajectory, with SIREN providing additional improvement in the final training stage. Interestingly, a similar pattern of a slightly faster initial error decrease (early in the training) with Tanh is also observed with SS-Broyden, as seen with SOAP. However, this difference is much smaller with SS-Broyden, indicating weaker sensitivity to the choice of activation function.

 \noindent The problem with Tanh PINN trained with Adam can be mitigated by adopting adaptive Tanh activations. While adaptive Tanh helps prevent premature convergence to poor local minima early in the training, the proper initialization of the adaptive parameters remains crucial. Additionally, we observed that employing slope recovery with layer-wise adaptive activations can substantially accelerate convergence. Introduced in a previous work by Jagtap et al. \cite{jagtap2020locally}, slope recovery is an additional loss term and can be defined by Eq. \ref{eq:slope_recovery} for layerwise adaptive activations:

\begin{equation}
    \mathcal{L}(\alpha) = \frac{L - 1}{\sum_{i=1}^{L} \exp(\alpha^i)},
    \label{eq:slope_recovery}
\end{equation}

\noindent where $L$ is the number of layers and $\alpha$ is the adaptive parameter of the activation function. Initializing the adaptive parameters to unity, effectively mirroring a standard Tanh 
activation, while employing a negligible weight for the slope recovery loss term  ($w_{\text{sr}}$) often leads to early entrapment in the same local minima as fixed Tanh architectures. However, strategically increasing $w_{\text{sr}}$ significantly accelerates convergence by encouraging the activation functions to adapt their slopes earlier in the training (Fig. \ref{fig:KdV_loss_history}).  It is critical to note that an excessively high weight (typically $w_{\text{sr}} \gtrsim 1$)  causes this term to dominate the objective function, resulting in numerical instability  during training. While adaptive Tanh activations facilitate gradient flow and assist  the Adam optimizer in bypassing local minima, the quasi-second order or second order optimizers demonstrate inherent robustness to these challenges. Consequently, we observed no significant performance gains when integrating adaptive Tanh or Sine activations into PINNs trained via the SOAP optimizer or SS-Broyden.\\

Previous studies have shown that KANs may suffer less from spectral bias and handle high-frequencies better compared to MLPs \cite{wang2024expressiveness}. Here we explore how KANs constrained by physical laws (PIKANs) compare to PINNs in terms of spectral bias. Also, how changing the univariate function in the KAN architecture can improve the results. We studied three different variants of KANs by changing the univariate functions, including B-Splines, radial basis functions (RBFs), and Chebyshev polynomials with degrees of three, five, or seven. Except for the B-spline PIKAN where the model stuck in a local minima and did not show proper convergence behavior. The rest of the PIKANs showed comparable results to the PINN with SIREN. Among the PIKANs, Chebyshev PIKAN (C-PIKAN) achieved the best result, while having the highest computational time. Note that in general, PIKANs took two to four times longer to converge compared to PINNs for solving the KdV equation. By increasing the polynomial degree in C-PIKAN, the results are only slightly improved while the computational time increases significantly for the same number of parameters. Therefore, C-PIKAN with degree three or five seems to be the most viable option within the PIKANs. Nevertheless, for the same optimizer, PINN with SIREN provides more accurate results with lower computational cost compared to the best PIKAN for KdV equation. Summary of the quantitative results including relative errors, Barron norm errors, errors in frequency domain, and convergence time are shown in Table \ref{table:KdV_results}.


\begin{table}[H]
\centering
\caption{\textbf{KdV equation prediction errors.} Comparison of the performance of PINNs with different activation functions and PIKANs with different univariate functions trained with different optimizers. All models except those trained with SS-Broyden have $\sim$ 40,000 parameters. The models trained with SS-Broyden have $\sim$ 12,900 parameters.}
\label{tab:pinn_pikan_comparison}
\setlength{\tabcolsep}{3pt}
\resizebox{\textwidth}{!}{%
\begin{tabular}{lccccccc}
\toprule
\textbf{} & {\makecell{Rel. $L^2$\\{Error}}} & {\makecell{Barron Norm\\{Rel. $L^2$ Error}}} & 
\textbf{$ Log(e_{\mathcal{F},p=0})$} &
\textbf{$ Log(e_{\mathcal{F},p=2})$} &
\textbf{$ Log(e_{\mathcal{F},p=4})$} &
{\makecell{Convergence \\{Time [hr.]}}} & 
{\makecell{Time to \\{1\% Error}}} \\
\midrule
\multicolumn{4}{l}{\textbf{PINNs}} \\
\midrule
Adam (Tanh)   & $2.47 \times 10^{-1}$ & $1.58 \times 10^{-1}$ & -1.52 & 0.77 & 5.34 & 0.056 & NA \\
Adam+LBFGS (Tanh)  & $8.12 \times 10^{-2}$ & $4.14 \times 10^{-2}$ &-2.48 &-0.15 &4.35 &0.056+0.21& NA \\
Adam (SIREN)  & $6.35 \times 10^{-2}$ & $1.06 \times 10^{-2}$ & -2.69 & -0.19 & 4.37 & 0.60& NA \\
Adam+LBFGS (SIREN) & $1.76 \times 10^{-3}$ & $1.28 \times 10^{-3}$ &-5.48 &-2.94 &1.37 &0.60+0.045 & 0.29 \\
SOAP (Tanh)   & $7.05 \times 10^{-4}$ & $5.62 \times 10^{-4}$ & -6.59 & -3.91& 0.39& 0.40& 0.033 \\
{SOAP (SIREN)}  & ${1.06 \times 10^{-4}}$ & {${5.92 \times 10^{-5}}$} & {-8.25}& {-5.58}& {-1.32} & 0.43& 0.11 \\
{SS-Broyden (Tanh)} & $8.69 \times 10^{-7}$ & $6.47 \times 10^{-7}$ & -12.42 & -9.37 & -5.53 & 0.70& 0.022 \\
{SS-Broyden (SIREN)} & $\boldsymbol{7.54 \times 10^{-8}}$ & $\boldsymbol{5.51 \times 10^{-8}}$ & \textbf{-14.41} & \textbf{-11.37} & \textbf{-7.66} & 1.57& \textbf{0.021} \\
\midrule
\multicolumn{4}{l}{\textbf{PIKANs}} \\
\midrule
SOAP (B-Spline)    & $9.20 \times 10^{-1}$ & $2.46 \times 10^{-1}$ &-0.37 &1.85 &6.38 & NA& NA \\
SOAP (RBF) & $5.99 \times 10^{-4}$ & $4.19 \times 10^{-4}$ &-6.74 &-4.17 &0.066 &0.94& 0.14 \\
SOAP (C7)     & $2.62 \times 10^{-4}$ & $2.45 \times 10^{-4}$ &-7.46 &-4.78 &-0.44 &1.67& 0.58 \\
SOAP (C5)     & $2.44 \times 10^{-4}$ & $2.46 \times 10^{-4}$ &-7.52 &-4.84 &-0.63 &1.71& 0.4 \\
SOAP (C3)     & $2.80 \times 10^{-4}$ & $2.55 \times 10^{-4}$ &-7.40 &-4.73 &-0.47 &1.35& 0.21\\
SS-Broyden (C7) & $2.20 \times 10^{-7}$ & $2.05 \times 10^{-7}$ &  -13.59 & -10.48 & -6.67 & 1.53& 0.026 \\
{SS-Broyden (C5)} & $\boldsymbol{1.48 \times 10^{-7}}$ & $\boldsymbol{1.38 \times 10^{-7}}$ &  \textbf{-13.92} & \textbf{-10.81} & \textbf{-7.00} & 1.07& 0.025 \\
SS-Broyden (C3) & $2.280 \times 10^{-7}$ & $2.44 \times 10^{-7}$ &  -13.43 & -10.33 & -6.53 & 0.71& \textbf{0.024} \\
\bottomrule
\end{tabular}
}
\label{table:KdV_results}
\end{table}

\subsection{Other experiments: Coupled effect of optimizer and activation function in hyperbolic and elliptic PDEs}

\noindent To further investigate the role of optimizers and activation functions, we examine how spectral bias manifests itself in hyperbolic equations such as wave equation and elliptic equations, such as steady-state diffusion-reaction equation.\\

We designed a systematic study with the hyperbolic wave equation to analyze the effect of different activation functions and optimizers in spectral bias of PINNs by gradually adding more frequencies to the solution and making the problem harder. Unlike the KdV equation, which exhibits either dispersive behavior, the wave equation introduces explicit propagation dynamics. The problem of interest involves a resting system with a time-dependent boundary condition where we systematically inject sinusoidal functions with multiple frequencies. The 1D wave equation is expressed as:
\begin{equation}
\begin{cases} 
\frac{\partial^2 u}{\partial t^2} - c^2 \frac{\partial^2 u}{\partial x^2} = 0 & x \in (0, 1), \, t \in (0, T] \\
u(x, 0) = 0, \quad \frac{\partial u}{\partial t}(x, 0) = 0 \\
u(0, t) = f(t), \quad u(1, t) = 0,
\end{cases}
\end{equation}
where $c = 1$, $T = 1$, and $f(t)$ is the time-dependent boundary condition. To systematically control the spectral complexity of the solution, we define
$f(t)$ as a superposition of sinusoidal modes:
\begin{equation}
f(t)
=
r(t;\tau)\,
\exp\!\left(
-\chi\,\frac{(t-\mu)^2}{2\sigma^2}
\right)
\sum_{i=1}^{N}
A_i \sin\!\left(2\pi f_i t\right),
\end{equation}

\noindent where $r(t;\tau)$ is the smooth ramp function ensuring compatibility with the initial conditions, $\chi$ activates the Gaussian envelope and removes it when $\chi = 0$, $A_i$ are the amplitudes, and $f_i$ are the frequencies. We consider the following four equations:

\begin{itemize}
\item[(i)] 
\(A_1=1,\;
f_1=1,\;
\chi=0.\)

\item[(ii)] 
\(
A_1=1,\;
f_1=10,\;
\chi=0.\)

\item[(iii)] 
\(
\{A_i\}_{i=1}^4=\{0.25,\,0.25,\,0.25,\,0.25\},\;
\{f_i\}_{i=1}^4=\{1,\,5,\,10,\,20\},\;
\chi=0.\)

\item[(iv)] 
\(
\{A_i\}_{i=1}^5=\{0.25,\,0.1,\,0.25,\,0.5,\,0.4\},\;
\{f_i\}_{i=1}^5=\{1,\,5,\,10,\,20,\,40\},\;
\chi=1.\)
\end{itemize}

\noindent In all cases, $r(t;\tau=0.05)$ is defined as follows:
\begin{equation}
r(t;\tau)=
\begin{cases}
0, & t \le 0, \\[6pt]
s^3\left(10 - 15s + 6s^2\right), 
& 0 < t < \tau, \quad s=\dfrac{t}{\tau}, \\[8pt]
1, & t \ge \tau.
\end{cases}
\end{equation}

\begin{figure}[H]
    \centering
    \includegraphics[width=1.0\linewidth]{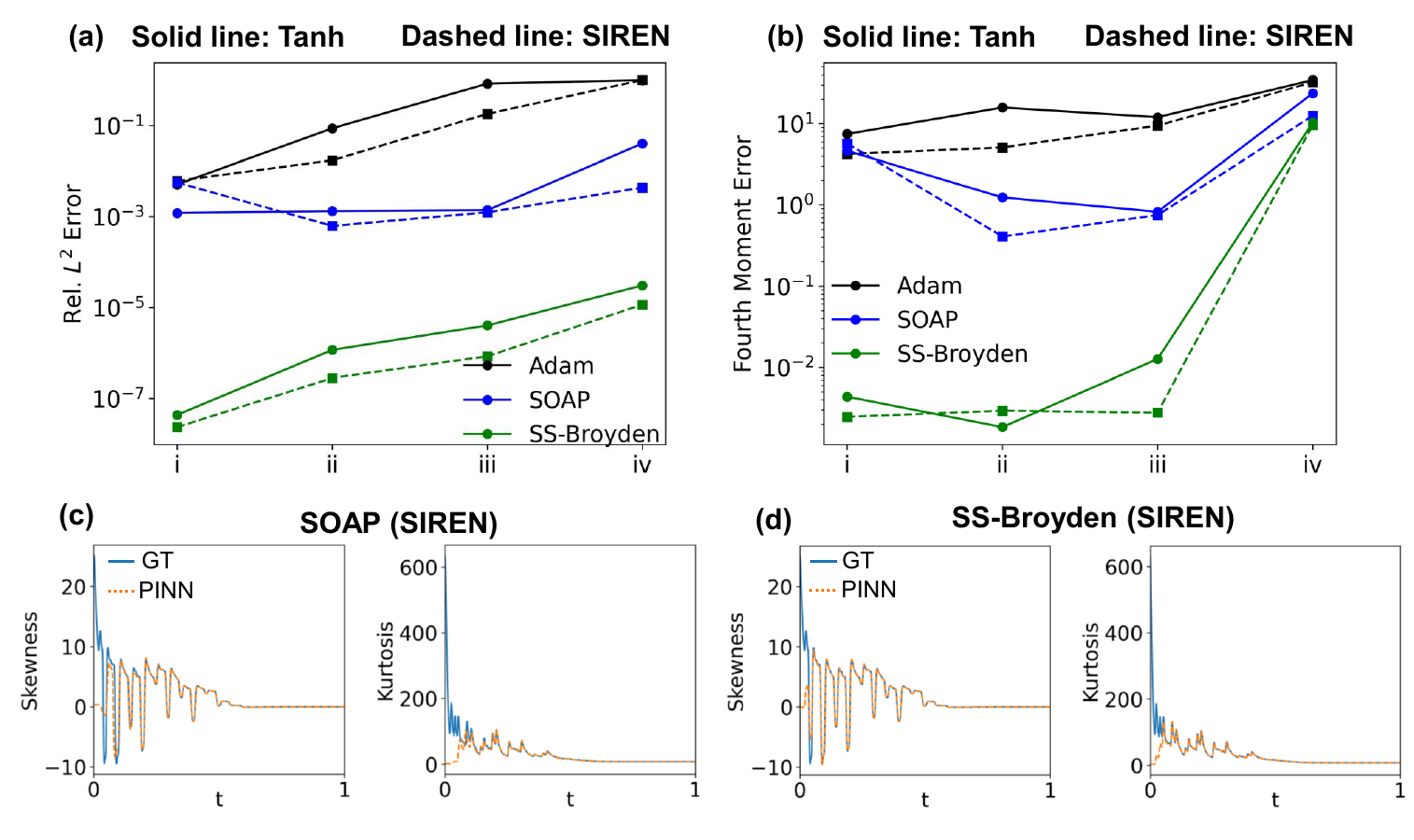}
    \caption{\textbf{Wave equation results with different optimizers and activation functions} (a-b). Relative $L^2$ error and fourth moment error of each of the wave problems trained with different optimizers and activation functions. (c) Skewness and Kurtosis of PINN prediction with SOAP optimizer and SIREN activation. (d) Skewness and Kurtosis of PINN prediction with SS-Broyden optimizer and SIREN activation.}
    \label{fig:Wave_Eq}
\end{figure}

\noindent Note that from case (i) to (iv) the frequencies injected at the boundary are increased, leading to monotonic increase in problem difficulty. Figure \ref{fig:Wave_Eq} shows how prediction errors increase with spectral complexity. The results demonstrate that the networks struggle more in the high-frequency regime (e.g., case (iv)). For low-frequency excitations (case~(i)), first-order
optimization with Adam is sufficient to achieve accurate solutions regardless of the activation function. As higher frequencies are introduced, SIREN provides some improvements when used with Adam. However, this benefit diminishes for more complex cases, where Adam ultimately fails to converge. In contrast, both SOAP and SS-Broyden exhibit significantly improved robustness as spectral complexity increases. Notably, SS-Broyden achieves two to four orders of magnitude lower relative $L^2$ errors compared to SOAP. Additionally, higher-order statistical diagnostics reveal clear differences. SS-Broyden achieves much lower errors in the third and fourth moments as well as in frequency errors, indicating superior recovery of high-frequency components (Fig. \ref{fig:Wave_Eq}). The quantitative results for the case with largest frequencies (the most complex problem) is shown in Table \ref{table:Wave_Eq}. The results of other cases are shown in the appendix \ref{tab:wave_cases_123}. 

\noindent Note that the choice of activation function has a strong impact when using the Adam optimizer; however, this dependence diminishes when using SOAP and becomes nearly negligible for SS-Broyden, which exhibits minimal sensitivity to the activation function. Additionally, the appropriate choice of activation function depends on the frequency content of the problem. For example, SIREN is better suited for case (iv) of the wave equation, where high-frequency components are prominent. In contrast, Tanh may be more suitable for problems dominated by lower-frequency content, unless SIREN is adjusted with a smaller frequency scaling parameter (see appendix \ref{tab:wave_cases_123}). This can be interpreted as an inductive bias of the network: for problems with strong high-frequency content, an activation function with inherently oscillatory, high-frequency behavior is advantageous, whereas for smoother, low-frequency problems, a monotonic activation such as Tanh may be more appropriate. Similarly, this choice becomes less significant with SS-Broyden.

\begin{table}[H]
\centering
\caption{\textbf{Wave equation case (iv) prediction errors.} Comparison of PINN and PIKAN performance for the high-frequency wave problem. Results for simpler cases (i)–(iii) are reported in the appendix. All models except those trained with SS-Broyden have $\sim$ 40,000 parameters. The models trained with SS-Broyden have $\sim$ 12,900 parameters.}
\label{table:Wave_Eq}
\setlength{\tabcolsep}{3pt}
\resizebox{\textwidth}{!}{%
\begin{tabular}{lccccc}
\hline
Method &
{\makecell{Rel. $L^2$\\Error}} &
{\makecell{Barron Norm\\Rel. $L^2$ Error}} &
$\log(e_{\mathcal{F},p=0})$ &
$\log(e_{\mathcal{F},p=2})$ &
$\log(e_{\mathcal{F},p=4})$  \\
\hline

\multicolumn{6}{l}{\textbf{PINN}} \\
\hline
Adam (Tanh)        & 1.00  & 0.99  & -1.65 & 2.23 & 6.80  \\
Adam (SIREN)       & 1.00  & 0.95  & -1.65 & 2.23 & 6.80  \\
SOAP (Tanh)        & $4.03 \times 10^{-2}$ & $2.12 \times 10^{-2}$ & -4.44 & -0.69 & 3.92  \\
SOAP (SIREN)       & $4.34 \times 10^{-3}$ &$2.54 \times 10^{-3}$ & -6.37 & -2.27 & 2.28  \\
{SS-Broyden (Tanh)}   & $3.05 \times 10^{-5}$ & $2.37 \times 10^{-4}$ & -10.67 & -5.22 & 0.56  \\
{SS-Broyden (SIREN)}  & $\boldsymbol{1.16 \times 10^{-5}}$ & $\boldsymbol{4.37 \times 10^{-5}}$ & \textbf{-11.52} & \textbf{-6.12} & \textbf{-0.37} \\

\hline
\multicolumn{6}{l}{\textbf{PIKAN}} \\
\hline
SOAP (C3)        & $3.36 \times 10^{-2}$ &  $2.07 \times 10^{-2}$ & -4.59 & -1.16 & 3.93  \\
SOAP (C5)       & $3.22 \times 10^{-2}$ & $2.23 \times 10^{-2}$ & -4.63 & -0.75 & 4.03  \\
SOAP (C7)       & $3.12 \times 10^{-2}$ & $1.83 \times 10^{-2}$ & -4.66 & -0.98 & 3.87  \\
{SS-Broyden (C3)}   & $3.80 \times 10^{-5}$ & $1.84 \times 10^{-4}$ & -10.46 & --5.05 & 0.73 \\
{SS-Broyden (C5)}  & $\boldsymbol{1.18 \times 10^{-5}}$ & $\boldsymbol{3.88 \times 10^{-5}}$ & \textbf{-11.50} & \textbf{-6.18} & \textbf{-0.41}  \\
{SS-Broyden (C7)}  & $1.75 \times 10^{-5}$ & $7.42 \times 10^{-5}$ & -11.16 & -5.71 & 0.065 \\

\hline
\end{tabular}
}
\end{table}

For an elliptic PDE, we consider a nonlinear steady reaction–diffusion equation posed on a two-dimensional periodic domain,
\begin{equation}
\Delta u(x,y) - k_r u(x,y)^2 = f(x,y),
\qquad (x,y) \in \Omega := [-1,1]^2,
\end{equation}
where $\Delta$ denotes the Laplacian operator and $k_r>0$ is a reaction coefficient controlling the strength of the nonlinear sink term. In all experiments, we fix $k_r = 0.1$. This problem serves as a representative nonlinear elliptic PDE with smooth but oscillatory solutions, commonly encountered in chemical kinetics and reaction–diffusion systems.\\

To enable quantitative error assessment, the forcing term $f(x,y)$ is constructed via the method of manufactured solutions. Specifically, we prescribe the exact solution
\begin{equation}
u^\ast(x,y) = \sin(3\pi x)\cos(3\pi y),
\end{equation}
which exhibits moderate spatial oscillations in both coordinate directions. Substituting $u^\ast$ into the governing equation yields the forcing
\begin{equation}
f(x,y) = \Delta u^\ast(x,y) - k_r \big(u^\ast(x,y)\big)^2
= -18\pi^2 u^\ast(x,y) - k_r \big(u^\ast(x,y)\big)^2.
\end{equation}

\noindent We impose periodic boundary conditions in both spatial dimensions,
\begin{equation}
u(-1,y)=u(1,y), \qquad u(x,-1)=u(x,1),
\end{equation}
along with periodicity of all derivatives. Rather than enforcing these constraints explicitly through boundary residuals, periodicity is embedded directly into the neural representation using a trigonometric feature map, ensuring the learned solution is periodic by construction.\\

Table~\ref{table:reaction} reports the prediction errors for the reaction–diffusion problem across PINN and PIKAN architectures under different optimizers and activation functions. We matched the number of trainable parameters between the MLP and Chebyshev-based PIKAN with degree 5, resulting in approximately $14.6\times10^3$ parameters for this problem. For degrees $3$ and $7$, the parameter counts differ due to the polynomial basis expansion.
Among PINN architectures, SS-Broyden with Tanh activation achieves the best overall performance, yielding the lowest relative $L^2$ error and consistently strong spectral metrics while maintaining short convergence time. In contrast, Adam exhibits significantly larger errors, particularly in higher-order spectral norms, reflecting the limitations of purely first-order gradient updates in resolving oscillatory residual structure. SOAP improves over Adam in the Tanh setting, consistent with its geometry-aware preconditioning that enhances gradient scaling and partially accounts for curvature effects. However, this advantage does not uniformly extend to SIREN activations. This behavior likely arises because SIREN’s sinusoidal activation already embeds strong oscillatory structure and provides smoother gradient propagation across frequency modes. In such cases, additional preconditioning may offer limited benefit and can even damp useful high-frequency gradient components. Within the PIKAN family, SS-Broyden achieves the strongest performance overall, with the best results observed for the Tanh-cPIKAN representation at higher polynomial degree. When parameter counts are matched, the Chebyshev-based PIKAN achieves comparable accuracy to the PINN with Tanh activation while requiring shorter time. Increasing the polynomial degree to $5$ or $7$ does not uniformly improve performance; although higher-degree polynomials increase expressivity, they also introduce greater curvature complexity and amplify higher-order spectral components within the loss landscape. Because the network depth is kept fixed across polynomial degrees, this additional curvature is not automatically balanced, leading to sensitivity in optimization dynamics. These observations suggest that performance is not determined solely by representational capacity, but rather by the interaction between polynomial degree, optimizer behavior, and the spectral structure induced by the Fourier feature embedding at the network input. In this context, the influence of the proposed Tanh-cKAN (and its physics-informed counterpart, Tanh-cPIKAN) is governed by the interplay between these elements.

\begin{table}[H]
\centering
\caption{\textbf{Steady state diffusion-reaction equation prediction errors.} Comparison of PINN and PIKAN performance for the diffusion-reaction problem.}
\label{table:reaction}
\setlength{\tabcolsep}{3pt}
\resizebox{\textwidth}{!}{%
\begin{tabular}{lccccccc}
\hline
Method &
{\makecell{Rel. $L^2$\\Error}} &
{\makecell{Barron Norm\\Rel. $L^2$ Error}} &
$\log(e_{\mathcal{F},p=0})$ &
$\log(e_{\mathcal{F},p=2})$ &
$\log(e_{\mathcal{F},p=4})$ & {\makecell{\#params\\K}}   & {\makecell{Convergence\\Time (hr)}} \\
\hline

\multicolumn{7}{l}{\textbf{PINN}} \\
\hline
Adam (Tanh)        & $1.41 \times 10^{-1}$   & $1.21 \times 10^{-4}$   & 1.69 & -2.39 & 1.99 & 14.6 & 0.18 \\
Adam (SIREN)       & $3.14 \times 10^{-3}$  & $2.66 \times 10^{-3}$   & -1.60 & 0.23 & 3.44 & 14.6 & 0.11 \\
SOAP (Tanh)        & $ 6.31 \times 10^{-2}$ & $3.17 \times 10^{-5}$ & 0.99 & -3.26 & -0.56 & 14.6 & 0.38 \\
SOAP (SIREN)       & $ 1.34 \times 10^{-2}$ & $ 3.93 \times 10^{-6}$ & -0.35 & -3.99 & -1.39 & 14.6 & 0.29\\
{SS-Broyden (Tanh)}  & \boldsymbol{$1.27 \times 10^{-6}$} & \boldsymbol{$7.08 \times 10^{-10}$} & \textbf{-8.39} & \textbf{-12.63} & \textbf{-10.11} & {14.6} & {0.06} \\
SS-Broyden (SIREN)  & $6.05 \times 10^{-6}$ & $3.48 \times 10^{-9}$ & -7.04 & -11.28 & -8.74 & 14.6 & \textbf{0.05} \\

\hline
\multicolumn{7}{l}{\textbf{PIKAN}} \\
\hline

Adam (C3)        & $1.33 \times 10^{-1}$ &  $9.73 \times 10^{-5}$ & 1.65 & -1.73 & 1.82 & 9.7 &0.07\\
Adam (C5)       & $1.81 \times 10^{-1}$ & $3.40 \times 10^{-5}$ & 1.91 & -1.82 & 1.43 & 14.6 & 0.11 \\
Adam (C7)       & $1.51 \times 10^{-1}$ & $4.21 \times 10^{-5}$ & 1.75 & -2.07 & 1.39 & 19.5 & 0.15\\
SOAP (C3)        & $ 1.00 \times 10^{-1}$ &  $ 2.93 \times 10^{-5}$ & 1.40  & -2.68 &  0.54 & 9.7 & 0.09 \\
SOAP (C5)       & $9.29 \times 10^{-2}$ & $8.11 \times 10^{-6}$ & 1.33 & -2.69 & 0.91 & 14.6 & 0.14  \\
SOAP (C7)       & $ 9.10\times 10^{-2}$ &  $2.24 \times 10^{-6}$ & 1.32 & -2.48 & 0.71  & 19.5 &0.18 \\
{SS-Broyden (C3)}   & $1.09 \times 10^{-3}$ & $6.15 \times 10^{-7}$ & -2.52 & -6.77 &  -4.26 & 9.7 & \textbf{0.04} \\
SS-Broyden (C5)  & $1.15 \times 10^{-4}$ &$6.49 \times 10^{-8}$ & -4.48 & -8.72 & -6.20 & 14.6 &0.06 \\
SS-Broyden (C7)  & $9.41 \times 10^{-4}$ & $5.59 \times 10^{-7}$ & -2.61 & -6.85 &-4.34 &  19.5 &0.10 \\
{SS-Broyden (Tanh- C7)}  & \boldsymbol{$1.94 \times 10^{-5}$} &  \boldsymbol{$1.05 \times 10^{-8}$} & \textbf{-6.03} & \textbf{-10.28} & \textbf{-7.71} &  {19.5} & {0.11} \\
\hline
\end{tabular}
}
\end{table}

\noindent Since the exact solution is periodic and dominated by a small number of Fourier modes, the Fourier feature layer spans the true spectral support of the solution, substantially mitigating spectral bias at the representation level \cite{tancik2020fourier}. Consequently, Tanh-cKAN does not increase expressivity; Chebyshev polynomials act on already oscillatory coordinates, and the additional Tanh merely rescales and saturates intermediate representations. Its role is therefore limited to (i) regularizing optimization geometry through improved Hessian conditioning and reduced curvature anisotropy, and (ii) controlling high-order mode amplification, including suppression of spurious Chebyshev components and residual-induced harmonics from the nonlinear reaction term. These mechanisms matter only when the optimizer--basis interaction becomes ill-conditioned. For first-order methods such as Adam, which do not exploit curvature information, improved conditioning provides no direct advantage. In contrast to data-driven settings without Fourier features, where spectral bias dominates and Tanh-cPIKAN can significantly aid Adam, the present PDE setup is geometry-limited rather than spectrum-limited, and no measurable improvement is observed under Adam. For SS-Broyden combined with high-degree Chebyshev expansions (C7), the polynomial basis induces strong curvature anisotropy that destabilizes quasi-Newton updates. Here, Tanh-cKAN compresses the Hessian spectrum and stabilizes inverse-Hessian approximations, yielding substantial accuracy gains. When curvature is already well-conditioned (C3--C5) or explicitly moderated by SOAP, the Tanh modification neither improves nor degrades performance. Accordingly, Tanh-cKAN results for those configurations are excluded from Table \ref{table:reaction} . Overall, Tanh-cKAN and Tanh-cPIKAN act as targeted mechanisms for mitigating curvature-induced optimization pathologies rather than as universal expressivity enhancements.\\

\begin{figure}[H]
    \centering
    \includegraphics[width=0.78\linewidth]{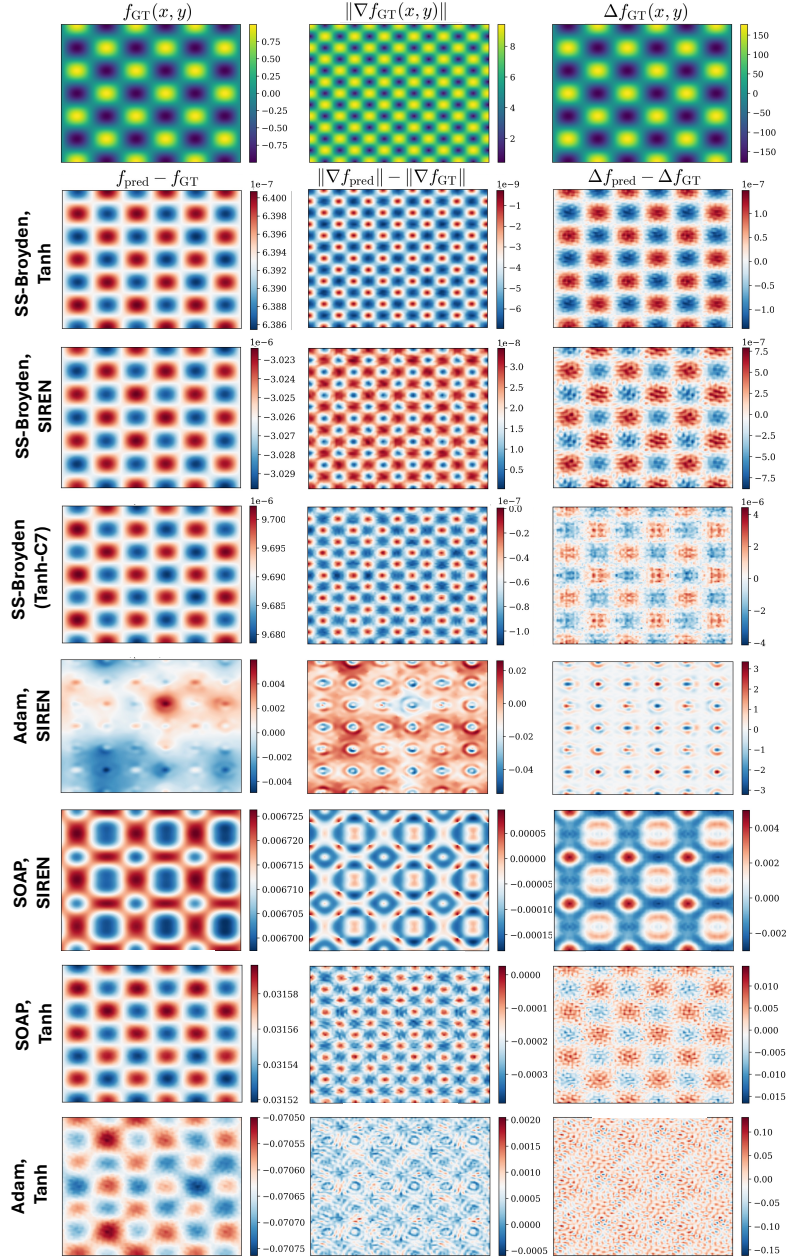}
    \caption{\textbf{Steady-state diffusion–reaction equation: Effect of optimizer and activation function.} Comparison of ground-truth field, gradient magnitude, and Laplacian (top row) with corresponding prediction errors (bottom row) for representative PINN and PIKAN configurations. The rows are organized from the most to the least accurate results (top to bottom). The error maps illustrate qualitative differences in spatial and derivative accuracy that are not fully captured by scalar norms alone. These visualizations complement the quantitative metrics reported in Table~\ref{table:reaction}.}

    \label{fig:Reaction}
\end{figure}

Figure~\ref{fig:Reaction} complements Table~\ref{table:reaction} by visualizing the field error, gradient error, and Laplacian error for each optimizer–activation configuration. While Table~\ref{table:reaction} quantifies global error norms, the figure reveals how these errors distribute spatially and across derivative orders. For Adam with Tanh activation, the field error exhibits smooth low-frequency distortions, and the Laplacian shows noticeable high-frequency artifacts, indicating incomplete recovery of higher modes. Adam with SIREN reduces the field error magnitude but introduces structured oscillations in the gradient and Laplacian, reflecting enhanced high-frequency excitation without stable derivative consistency. SOAP produces more spatially coherent patterns than Adam due to its geometry-aware preconditioning. With Tanh, this mainly reduces diffuseness without altering spectral character, whereas SOAP–SIREN yields organized periodic residuals aligned with dominant Fourier modes, though derivative-level oscillations persist. In contrast, SS-Broyden yields nearly uniform and significantly smaller errors across field, gradient, and Laplacian levels. The residual patterns appear homogeneous and close to numerical noise. It confirms that curvature-aware updates resolve both low- and high-frequency components simultaneously.

\subsection{Spectral bias in neural operators}

\noindent Here, we consider two different applications. First, we consider a sonic jet at low resolution and we aim to reconstruct it at high resolution.
Next, we consider earthquake dynamics and its effect on a six-story reinforced concrete frame building subjected to ground acceleration. 

\subsubsection{High-speed Schlieren imaging}

\noindent High-speed Schlieren imaging is a standard tool for visualizing compressible flows by converting line-of-sight refractive-index gradients (and hence density-gradient features) into intensity variations \cite{settles2017review,settles2022schlieren}. 
Here, we consider an impinging-jet experiment in which an under-expanded air jet (nozzle-exit Mach number \(M_e \approx 1\)) is directed onto a flat plate at ambient conditions using a conventional Z-type Schlieren setup. 
The resulting flow exhibits coherent shock structures in the near field together with shear-layer instabilities and a turbulent wall-jet after impingement, producing abundant fine-scale features in the Schlieren field. 
These multiscale structures carry significant energy into high spatial wavenumbers, leading to a relatively slow-decaying spatial energy spectrum and making the dataset a stringent testbed for studying spectral bias in learned surrogates.\\

The dataset consists of 1000 Schlieren snapshots at a spatial resolution of \([128,256]\) over the domain \([-0.8,\,5.6]\times[-1.45,\,1.75]\), with successive frames separated by \(\tau = 4.76\,\mu s\). 
We use the first 800 snapshots for training, the next 100 for validation, and the final 100 for testing. 
Let \(u_{\mathrm{HR}}(\bm{x},t)\) denote the high-resolution Schlieren field. 
In this work we focus on \emph{spatial} super-resolution only: the low-resolution observation \(u_{\mathrm{LR}}\) is obtained by spatially subsampling \(u_{\mathrm{HR}}\) by a factor of 8, and the surrogate is trained to learn the mapping
\(u_{\mathrm{LR}}\mapsto u_{\mathrm{HR}}\). 
This isolates the reconstruction challenge to recovering fine spatial detail - including high-wavenumber content associated with shock-turbulence interactions - from heavily downsampled Schlieren measurements \cite{wang2022deep,zhang2025operator}.\\

\begin{figure}[H]
    \centering
    \includegraphics[width=0.9\linewidth]{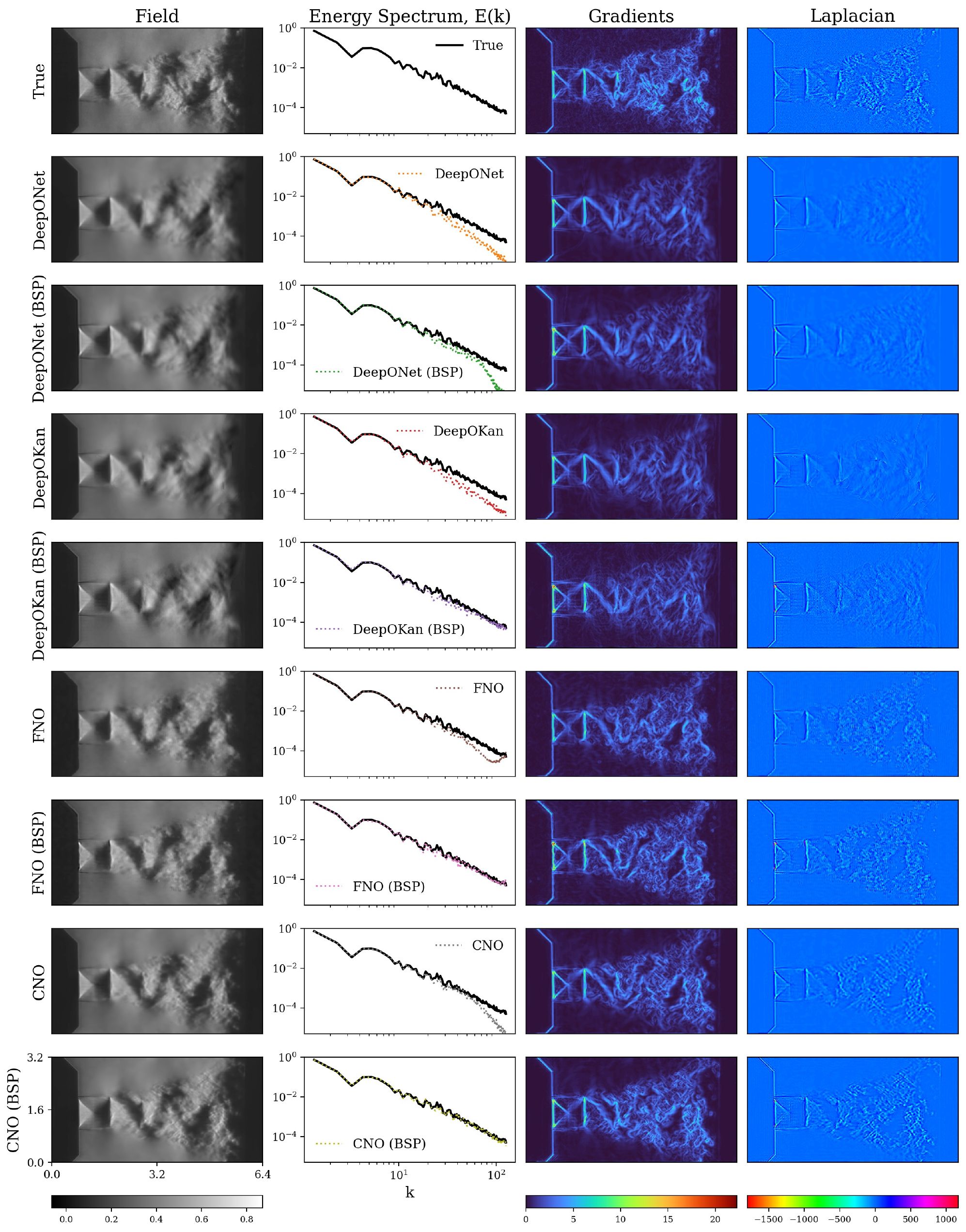}
    \caption{\textbf{Turbulent jet: Spectral bias in neural operators.} 
    Columns show (from left to right) schlieren images of an impinging jet, the corresponding energy spectrum E(k), spatial gradients, and the Laplacian. Rows compare the ground truth against predictions from different neural operator architectures (with and without BSP training). We observe that using Binned Spectral Power (BSP) loss helps mitigate spectral bias, improving agreement with the true energy spectrum across wavenumbers. All the results are based on SOAP optimizer.
    }
    \label{fig:bsp_arch}
\end{figure}

Results are summarized in \autoref{fig:bsp_arch} and \autoref{tab:impjet_no}. 
Across architectures, standard $L^2$ error minimization training recovers the dominant shock topology and large-scale flow organization, but exhibits clear spectral bias: high-wavenumber content is systematically attenuated, yielding spectra $E(k)$ that deviate rapidly relative to the ground truth. 
This loss of small-scale power is consistent with the qualitative diagnostics in \autoref{fig:bsp_arch}, where baseline predictions appear overly smooth and the corresponding gradient and Laplacian fields under-represent sharp density-gradient features that are prominent in the Schlieren images. 
In contrast, incorporating the minimization of binned spectral power (BSP) loss\cite{chakraborty2025binned} during training substantially improves fidelity at fine scales. 
We implemented a log-transformed variant of the BSP loss as described below.
\begin{equation}
\mathcal{L}
=
\sum_{i=1}^{N}\underbrace{ \left\| v_i - \hat{v}_i \right\|_2}_{\text{Field error}}
+
\underbrace{\left\| \mathcal{B}(v_i) - \mathcal{B}(\hat{v}_i) \right\|_2}_{\text{BSP error}},
\label{eq:bsp_loss}
\end{equation}
where \(v=u_{HR}\), \(\hat{v}=\mathcal{G}_{\theta}(u_{LR})\) and \(\mathcal{B}(\cdot)\) denotes \(\log\) of the binned energy spectrum representation\cite{chakraborty2025binned}.
To ensure that both the loss terms have a similar order of magnitude, both the field and BSP errors are min-max normalized with respect to min and max computed from the training dataset. 
The predicted spectra follows the ground truth much more closely into the high-$k$ regime, and the reconstructed fields retain the fine-scale turbulence structures, reflected by more accurate gradients and Laplacians. Quantitatively, BSP yields large reductions in spectral and Barron-norm errors while leaving the field normalized RMSE (nRMSE) largely unchanged (\autoref{tab:impjet_no}). 
The energy-spectrum nRMSE drops by $3.1\times$ for DeepOKAN (0.1289$\rightarrow$0.0416), $3.4\times$ for FNO (0.0865$\rightarrow$0.0256), and $5.6\times$ for CNO (0.1172$\rightarrow$0.0211), with commensurate reductions in Barron-norm error of roughly $2.2\text{-}4.1\times$ (DeepOKAN: 0.3659$\rightarrow$0.1662; FNO: 0.2641$\rightarrow$0.0828; CNO: 0.3229$\rightarrow$0.0792). 
These improvements are consistent with mitigating spectral bias rather than simply rescaling the output. BSP primarily targets the high-frequency tail and associated sharp features, which are poorly captured by the baseline loss. 
DeepONet is comparatively insensitive to BSP in this setting, showing no improvement in spectral error and only a modest reduction in Barron-norm error. 
Finally, since BSP modifies only the training objective, parameter counts and inference times are unchanged across each model pair, so the gains in spectral fidelity come at no additional inference cost.\\

Beyond the training objective, we find that the choice of optimizer can also influence high-frequency fidelity.
In particular, SOAP tends to preserve more high-frequency energy than Adam, leading to sharper gradient and Laplacian diagnostics.
A focused comparison is provided in \autoref{app:optimizer_spectral_bias} and \autoref{fig:no_opt}.

\begin{table}[t]
\centering
\small
\setlength{\tabcolsep}{2pt}
\renewcommand{\arraystretch}{0.85}
\begin{tabular}{@{}lccccc@{}}
\toprule
\textbf{Model} &
\textbf{\shortstack{Field\\Error}} &
\textbf{\shortstack{Energy-\\Spectrum\\Error}} &
\textbf{\shortstack{Barron-\\Norm\\Error}} &
\textbf{\shortstack{\# params\\(M)}} &
\textbf{\shortstack{Inference\\time\\(s)}} \\
\midrule
DeepONet        & 0.0529 & 0.1566 & 0.4171 & 3.8 & \textbf{0.00086} \\
DeepONet (BSP)  & 0.0535 & 0.1675 & 0.4083 & 3.8 & 0.00086 \\
DeepOKAN        & 0.0538 & 0.1289 & 0.3659 & 3.0 & 0.00991 \\
DeepOKAN (BSP)  & 0.0547 & 0.0416 & 0.1662 & 3.0 & 0.00991 \\
FNO             & 0.0517 & 0.0865 & 0.2641 & 3.4 & 0.00433 \\
FNO (BSP)       & 0.0556 & 0.0256 & 0.0828 & 3.4 & 0.00433 \\
CNO             & \textbf{0.0512} & 0.1172 & 0.3229 & 3.2 & 0.01238 \\
CNO (BSP)       & 0.0543 & \textbf{0.0211} & \textbf{0.0792} & 3.2 & 0.01238 \\
\bottomrule
\end{tabular}
\caption{Comparison of neural operator errors and costs for the impinging jet problem.}
\label{tab:impjet_no}
\end{table}

\subsubsection{Earthquake problem}
\noindent Here we consider the dynamic response of a six-story reinforced concrete frame building subjected to ground acceleration records from the PEER NGA-West2 database (\url{https://ngawest2.berkeley.edu/})\footnote{Ground acceleration records are taken from the Pacific Earthquake Engineering Research Center (PEER: \url{https://ngawest2.berkeley.edu/}, \url{https://peer.berkeley.edu/})}.
The governing equation of motion for the linear multi-degree-of-freedom system is
\begin{equation}
\mathbf{M}\ddot{\mathbf{x}} + \mathbf{C}\dot{\mathbf{x}} + \mathbf{K}\mathbf{x} = \mathbf{M}\boldsymbol{\iota}\ddot{u}_g(t),
\label{eq:earthquake_gov}
\end{equation}
where $\mathbf{M}$, $\mathbf{C}$, and $\mathbf{K} \in \mathbb{R}^{504 \times 504}$ are the mass, damping, and stiffness matrices from finite element discretizations, $\mathbf{x}(t)$ is the displacement vector, $\boldsymbol{\iota}$ is the influence vector distributing ground motion to degrees of freedom, and $\ddot{u}_g(t)$ is the ground acceleration. The system starts at rest with $\mathbf{x}(0) = \dot{\mathbf{x}}(0) = \mathbf{0}$. For a linear system, the response can be expressed via convolution with the Green's function,
\begin{equation}
\mathbf{x}(t) = \int_0^t \ddot{u}_g(\tau) \mathbf{h}(t-\tau) \, d\tau,
\label{eq:greens_convolution}
\end{equation}
where the neural operator $\mathcal{G}_\theta$ approximates the mapping $\ddot{u}_g(t) \mapsto x_1(t)$ from ground acceleration to roof displacement.\\

The dataset comprises 144 earthquake ground motion records, with 100 samples for training and 44 for testing.
Each record spans 80 seconds at 50 Hz sampling (4000 timesteps), with records pre-processed using a Butterworth filter (0.1--24.9 Hz) and resampled to uniform $\Delta t = 0.02$ s.
The structural response exhibits both low-frequency global modes and high-frequency content from impulsive ground motions, presenting a challenging multi-scale prediction problem. The temporal resolution of this problem creates a significantly more demanding spectral learning task than the impinging-jet experiment.
For a discrete signal of $N$ points sampled at rate $f_s$, the Nyquist frequency is $f_s/2$ and there are $N/2$ discrete frequency bins between zero and the Nyquist limit.
The earthquake time series ($N = 4000$, $f_s = 50$\,Hz) therefore contains 2000 resolvable frequency bins up to 25\,Hz, consistent with the Butterworth filter's 24.9\,Hz cutoff, compared to 64 modes for the impinging jet's 128-point spatial dimension.
This roughly $30\times$ increase in the number of spectral modes the network must resolve makes the spectral bias problem increasingly challenging.\\

The neural operator architectures used for this problem differ in how they process temporal information.
DeepONet and DeepOKAN employ a causal windowing approach where the response at each timestep depends only on past input history, implemented via zero-padding with a shifting window that encodes physical causality; the output state at the current timestep is not affected by input states at future timesteps.
For additional details on the causal formulation and dataset, we refer readers to Liu et al.~\cite{liu2024causality}.
In contrast, FNO \& CNO process the entire input sequence to predict the full output time-series in a single forward pass, and subsequently, the causality was not strictly enforced. 
All metrics reported are computed in real (physical) space on the de-normalized predictions.\\

\begin{figure}[H]
    \centering
    \includegraphics[width=\linewidth]{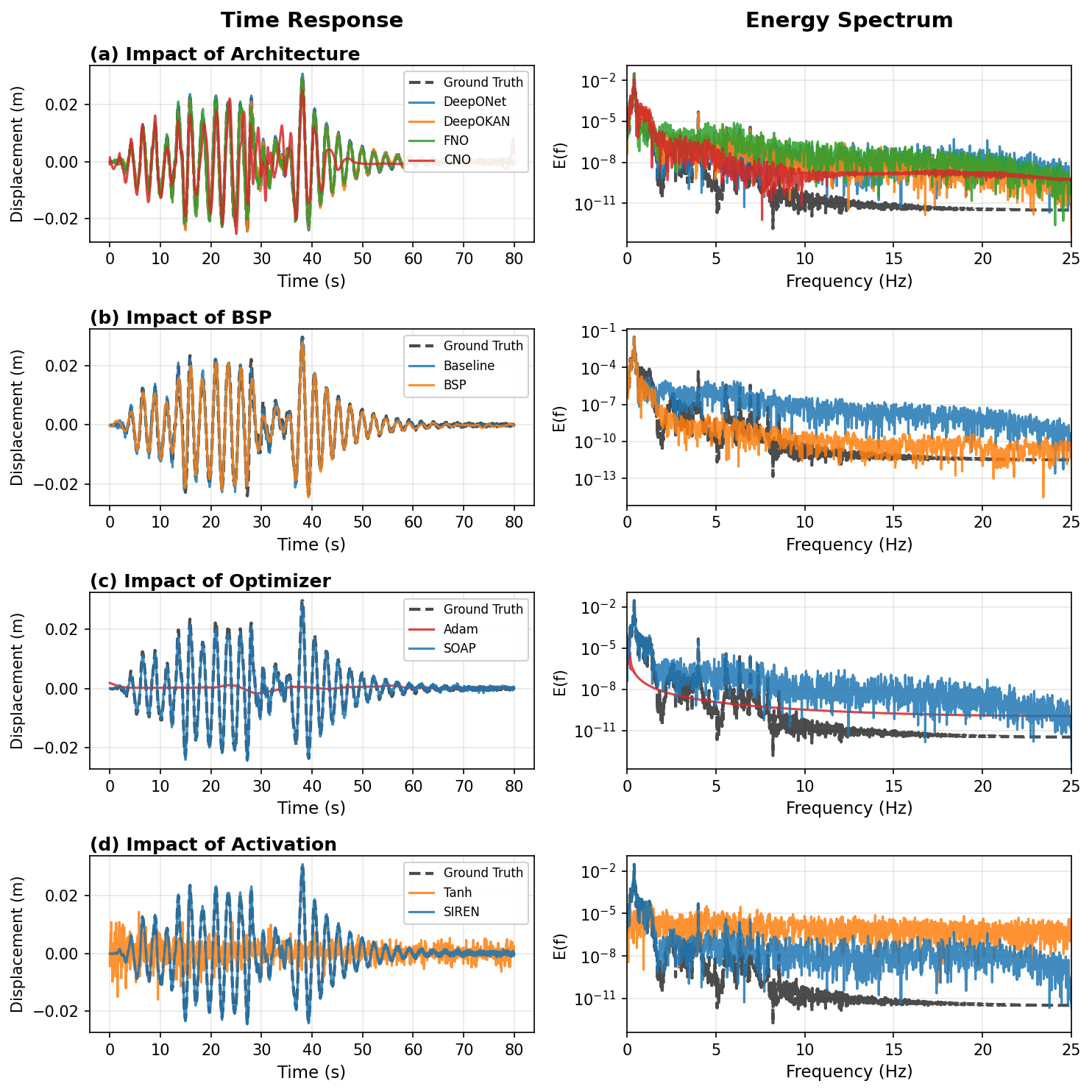}
    \caption{\textbf{Representative earthquake response predictions illustrating key design choices.} Each row shows the time-domain response (left) and energy spectrum (right) for the same test sample. (a)~DeepONet (SIREN) achieves the lowest field error (NRMSE $= 0.0008$; Table~\ref{tab:earthquake_arch}), followed by DeepOKAN ($0.0022$), FNO ($0.0038$), and CNO ($0.0133$). (b)~BSP improves FNO spectral fidelity, reducing log spectral error from $0.149$ to $0.082$ ($1.8\times$), visible in the high-frequency alignment. (c)~SOAP enables DeepOKAN convergence (NRMSE $= 0.0022$), whereas Adam fails across all seeds. (d)~SIREN activation enables DeepONet convergence without BSP (NRMSE $= 0.0008$), while Tanh fails to converge under the same conditions. Black dashed lines denote ground truth; extended comparisons are in Figures~\ref{fig:earthquake_bsp}--\ref{fig:earthquake_causal}.}
    \label{fig:earthquake_compact}
\end{figure}

We also investigate minimizing the BSP error (see \autoref{eq:bsp_loss}) as a strategy to mitigate spectral bias. 
Computing this term requires computing FFT over the entire predicted sequence, so the BSP loss can only be evaluated once all timesteps of the predicted outputs are assembled.
Training with BSP error minimization was straightforward for FNO \& CNO.
However, for DeepONet/DeepOKAN with causal training~\cite{liu2024causality}, predictions from all timesteps must first be collected before computing BSP. \label{sec:loss_alignment} Results are summarized in Table~\ref{tab:earthquake_arch} and Figure~\ref{fig:earthquake_compact}.
All architectures use SOAP optimizer and causal training for DeepONet/DeepOKAN, with error metrics computed in physical space.

\begin{table}[H]
\centering
\small
\setlength{\tabcolsep}{2pt}
\renewcommand{\arraystretch}{0.85}
\caption{Architecture and loss comparison for earthquake structural response prediction. SOAP optimizer, causal training for DeepONet/DeepOKAN.}
\label{tab:earthquake_arch}
\begin{tabular}{@{}lcccccc@{}}
\toprule
\textbf{Model} &
\multicolumn{2}{c}{\textbf{\shortstack{Field\\Error}}} &
\multicolumn{2}{c}{\textbf{\shortstack{Log Spectral\\Error}}} &
\multicolumn{2}{c}{\textbf{\shortstack{Barron Norm\\Error}}} \\
\cmidrule(lr){2-3} \cmidrule(lr){4-5} \cmidrule(lr){6-7}
 & Baseline & BSP & Baseline & BSP & Baseline & BSP \\
\midrule
DeepONet (SIREN)  & 0.0008 & $0.0009$ & $0.1257$ & $0.1366$ & $0.0019$ & $0.0027$ \\
DeepONet (Tanh)   & $-$ & \textbf{0.0006} & $-$ & $0.1103$ & $-$ & \textbf{0.0012} \\
DeepOKAN          & $0.0022$ & $0.0044$ & $0.1230$ & $0.1750$ & $0.0029$ & $0.0077$ \\
FNO               & $0.0038$ & $0.0041$ & $0.1485$ & $0.0819$ & $0.0045$ & $0.0030$ \\
CNO               & $0.0133$ & $0.0145$ & $0.1323$ & \textbf{0.0758} & $0.0040$ & $0.0043$ \\
\bottomrule
\multicolumn{7}{l}{\footnotesize $-$ Failed to converge.}
\end{tabular}
\end{table}

\noindent Consistent with the impinging-jet findings, BSP improves spectral fidelity when the field loss also operates on the full output sequence (FNO, CNO), as illustrated for FNO in Figure~\ref{fig:earthquake_compact}(b) (see Figure~\ref{fig:earthquake_bsp} for all architectures).
FNO achieves a $1.8\times$ reduction in log spectral error ($0.149 \rightarrow 0.082$) and CNO a $1.7\times$ reduction ($0.132 \rightarrow 0.076$), with corresponding improvements in Barron norm error.

\noindent The strongest evidence for BSP's effect comes from derivative metrics, where the ground truth is dominated by high-frequency content.
FNO acceleration error ($d^2u/dt^2$) drops $2.2\times$ ($0.064 \rightarrow 0.030$; Table~\ref{tab:earthquake_opt}).
Similarly, CNO acceleration improves more modestly ($0.032 \rightarrow 0.026$, $1.2\times$; Table~\ref{tab:earthquake_opt}).\\

Unlike the impinging-jet problem, we did not find the same improvement from BSP for DeepONet and DeepOKAN.
DeepONet shows slight degradation across all metrics with BSP: spectral error worsens ($0.126 \rightarrow 0.137$), field error increases ($0.0008 \rightarrow 0.0009$), and Barron norm error rises ($0.0019 \rightarrow 0.0027$).
DeepOKAN also exhibits degradation with BSP. As noted earlier, BSP requires the full sequence while the causal field loss operates per-timestep, so the two objectives see different views of the prediction.
Looking at the cosine similarity between the per loss gradients we see that these competing views produce opposing gradient signals: DeepONet (SIREN) exhibits strongly negative cosine similarity between L2 and BSP gradients ($-0.48$ by the end of training), while FNO maintains positive alignment ($+0.28$).
Crucially, when the same architectures are trained in non-causal mode (removing the causal/non-causal mismatch), gradient alignment shifts toward positive: DeepONet (SIREN) moves from $-0.48$ to $+0.13$, see Table~\ref{tab:gradient_cosine} and Figure~\ref{fig:cosine_sim_completion} in \autoref{sec:earthquake_gradient}. 
Nevertheless, we found causality to be necessary to make meaningful predictions for DeepONet/DeepOKAN. When causality is removed and these architectures are trained like FNO and CNO, they mostly fail to converge; this is discussed further in \autoref{sec:earthquake_causal} (Table~\ref{tab:earthquake_causal}). The notable exception of BSP's performance with a causal approach is DeepONet with Tanh activation: without BSP the model fails to converge ($0.024$ NRMSE, $0.468$ acceleration error), but BSP enables convergence to the best field accuracy ($0.0006$) with acceleration error dropping to $0.030$.\\

Beyond the training objective, optimizer selection and activation function also interact with BSP effectiveness on this dataset.
SOAP provides $2.0\times$ lower NRMSE for DeepONet SIREN and enables convergence for DeepOKAN (Adam fails across all seeds, Figure~\ref{fig:earthquake_compact}(c)), with strong performance on FNO and CNO as well (see \autoref{sec:earthquake_optimizer}).
Activation function and spectral regularization interact non-trivially: SIREN's periodic activations provide implicit spectral bias mitigation sufficient for convergence without BSP, while Tanh requires BSP to converge.
This suggests that SIREN and BSP address overlapping spectral deficiencies, whereas Tanh and BSP provide more complementary mechanisms (see \autoref{sec:earthquake_activation} and Figure~\ref{fig:earthquake_compact}(d)).
Detailed comparisons are provided in \autoref{app:optimizer_spectral_bias}.

\section{Conclusions and Outlook}
\label{summary}

\noindent We presented a systematic investigation of spectral bias in physics-informed neural networks, physics-informed Kolmogorov–Arnold networks, and neural operators. Through a combination of theoretical analysis and extensive numerical experiments across elliptic, hyperbolic, and dispersive PDEs, we demonstrated that spectral bias in scientific machine learning is not solely a representational limitation of neural architectures, but it depends on optimization dynamics, loss formulation, and their interaction with network expressivity.\\

From a theoretical side, linearized analyses based on neural tangent kernel and Gauss–Newton dynamics clarify why first-order optimization methods preferentially learn low-frequency modes, while higher-frequency components converge slowly. In contrast, quasi-second-order and second-order optimization strategies substantially reduce the frequency dependence of convergence rates by implicitly preconditioning the training dynamics. Although these arguments rely on idealized assumptions, our empirical results show that their qualitative implications persist well beyond the asymptotic regime.\\

Across a wide range of benchmark problems, we observed that optimization choice plays a dominant role in mitigating spectral bias in PINNs and PIKANs. First-order optimizers such as Adam often converge to spectrally incomplete solutions that satisfy low-order statistics and PDE residuals in an averaged sense, while failing to capture gradients, curvatures, and localized high-frequency features. Quasi-second-order methods such as SOAP, and in particular fully second-order methods such as  SS-Broyden, significantly improve both convergence behavior and spectral resolution, enabling accurate recovery of higher-order moments and Barron norms. These improvements are especially pronounced in stiff elliptic and dispersive problems, where spectral imbalance has the most severe physical consequences.\\

Architectural choices and activation functions were shown to play a complementary but secondary role. Oscillatory activations such as sine functions and SIREN-type parameterizations improve high-frequency representation under first-order optimization and can help escape early stagnation. However, their influence diminishes as optimization strength increases, indicating that representational enhancements alone are insufficient to fully overcome spectral bias without high-order optimization. Similar trends were observed in physics-informed KANs, where polynomial-based univariate functions exhibit classical spectral limitations for non-smooth solutions unless combined with strong optimization strategies.\\

For neural operators, our results demonstrate that spectral bias persists even in architectures that explicitly operate in Fourier space. Standard $L^2$ training objectives remain dominated by low-frequency content, leading to systematic underestimation of high-wavenumber energy. Spectral-aware loss formulations, such as binned spectral power losses, provide an effective and computationally inexpensive approach to restore spectral balance across operator architectures, including DeepONet, DeepOKAN, FNO, and CNO. These findings highlight that loss design is also critical for achieving spectrally resolved operator learning.\\

In summary, our study suggests that spectral bias in scientific machine learning should be viewed primarily as a dynamical phenomenon arising from ill-conditioned training objectives, rather than as an intrinsic failure of neural representations. While expressive architectures and carefully chosen activations can improve performance, robust mitigation of spectral bias requires curvature-aware optimization and, in operator settings, spectrally informed loss functions. From a practical standpoint, our results lead to concrete guidelines: strong second-order or quasi-second-order optimizers are essential for physics-informed learning in multiscale PDEs, while spectral losses play a key role in operator learning when high-frequency fidelity is required.\\

Several open questions remain. Extending these findings to large-scale three-dimensional problems, understanding the interaction between spectral bias and adaptive sampling strategies, and developing scalable second-order optimization methods tailored to physics-informed objectives are promising directions for future work. For example, currently, SOAP can use mini-batching but SS-Broyden cannot, so this is a concrete direction to be pursued given the accuracy superiority of this method.

\section*{CRediT authorship contribution statement}
\noindent \textbf{Siavash Khodakarami:} Writing - review \& editing, Writing - original draft, Visualization, Validation, Software, Methodology, Investigation, Formal analysis, Data Curation, Conceptualization.
\textbf{Vivek Oommen:} Writing - review \& editing, Writing - original draft, Visualization, Validation, Software, Methodology, Investigation, Formal analysis, Data Curation, Conceptualization.
\textbf{Nazanin Ahmadi Daryakenari:} Writing - review \& editing, Writing - original draft, Visualization, Validation, Software, Methodology, Investigation, Formal analysis, Data Curation, Conceptualization.
\textbf{Maxim Beekenkamp:} Writing - review \& editing, Writing - original draft, Visualization, Validation, Software, Methodology, Investigation, Formal analysis, Data Curation.
\textbf{George Em Karniadakis} Writing - review \& editing, Writing - original draft, Supervision, Funding acquisition, Validation, Methodology, Formal analysis, Conceptualization.

\section*{Declaration of competing interest}
\noindent The authors declare that they have no known competing financial interests or personal relationships that could have appeared to influence the work reported in this paper.

\section*{Acknowledgements}
\noindent We would like to acknowledge funding from the Office of Naval Research as part of MURI-METHODS project with grant number N00014242545 and the ONR
Vannevar Bush Faculty Fellowship (N00014-22-1-2795). The authors would like to acknowledge the computational resources and services at the Center for Computation and Visualization (CCV), Brown University. We acknowledge that the Schlieren impinging-jet experiments used in this work were conducted at the SMC Lab, Tsinghua University. 
The corresponding Schlieren dataset originates from Tsinghua University, and the associated intellectual property of this dataset remains with Tsinghua University.

\section*{Data Availability}
\noindent All codes and datasets except the Schlieren dataset will be made publicly available at \url{https://github.com/SiaK4/PIN_NO_Spectral_Bias.git} upon publication.\\
Please contact Prof. He Feng (hefeng@tsinghua.edu.cn) to access the impinging jet Schlieren dataset.

\bibliography{References}

@article{zhang2025operator,
  title={Operator Learning for Reconstructing Flow Fields from Sparse Measurements: an Energy Transformer Approach},
  author={Zhang, Qian and Krotov, Dmitry and Karniadakis, George Em},
  journal={arXiv preprint arXiv:2501.08339},
  year={2025}
}

@article{settles2017review,
  title={A review of recent developments in schlieren and shadowgraph techniques},
  author={Settles, Gary S and Hargather, Michael J},
  journal={Measurement Science and Technology},
  volume={28},
  number={4},
  pages={042001},
  year={2017},
  publisher={IOP Publishing}
}

@article{settles2022schlieren,
  title={Schlieren and BOS velocimetry of a round turbulent helium jet in air},
  author={Settles, Gary S and Liberzon, Alex},
  journal={Optics and Lasers in Engineering},
  volume={156},
  pages={107104},
  year={2022},
  publisher={Elsevier}
}

@article{wang2022deep,
  title={Deep-learning-based super-resolution reconstruction of high-speed imaging in fluids},
  author={Wang, Zhibo and Li, Xiangru and Liu, Luhan and Wu, Xuecheng and Hao, Pengfei and Zhang, Xiwen and He, Feng},
  journal={Physics of Fluids},
  volume={34},
  number={3},
  year={2022},
  publisher={AIP Publishing}
}

@article{JAGTAP2020109136,
title = {Adaptive activation functions accelerate convergence in deep and physics-informed neural networks},
journal = {Journal of Computational Physics},
volume = {404},
pages = {109136},
year = {2020},
issn = {0021-9991},
doi = {https://doi.org/10.1016/j.jcp.2019.109136},
url = {https://www.sciencedirect.com/science/article/pii/S0021999119308411},
author = {Ameya D. Jagtap and Kenji Kawaguchi and George Em Karniadakis}
}

@article{SHUKLA2024117290,
title = {A comprehensive and {FAIR} comparison between {MLP} and {KAN} representations for differential equations and operator networks},
journal = {Computer Methods in Applied Mechanics and Engineering},
volume = {431},
pages = {117290},
year = {2024},
issn = {0045-7825},
doi = {https://doi.org/10.1016/j.cma.2024.117290},
url = {https://www.sciencedirect.com/science/article/pii/S0045782524005462},
author = {Khemraj Shukla and Juan Diego Toscano and Zhicheng Wang and Zongren Zou and George Em Karniadakis}
}

@article{vyas2024soap,
  title={Soap: Improving and stabilizing {Shampoo} using {Adam}},
  author={Vyas, Nikhil and Morwani, Depen and Zhao, Rosie and Kwun, Mujin and Shapira, Itai and Brandfonbrener, David and Janson, Lucas and Kakade, Sham},
  journal={arXiv preprint arXiv:2409.11321},
  year={2024}
}

@article{URBAN2025113656,
title = {Unveiling the optimization process of physics informed neural networks: How accurate and competitive can {PINNs} be?},
journal = {Journal of Computational Physics},
volume = {523},
pages = {113656},
year = {2025},
issn = {0021-9991},
doi = {https://doi.org/10.1016/j.jcp.2024.113656},
url = {https://www.sciencedirect.com/science/article/pii/S0021999124009045},
author = {Jorge F. Urbán and Petros Stefanou and José A. Pons}
}

@article{zabusky1965interaction,
  title={Interaction of" solitons" in a collisionless plasma and the recurrence of initial states},
  author={Zabusky, Norman J and Kruskal, Martin D},
  journal={Physical Review Letters},
  volume={15},
  number={6},
  pages={240},
  year={1965},
  publisher={APS}
}

@article{sitzmann2020implicit,
  title={Implicit neural representations with periodic activation functions},
  author={Sitzmann, Vincent and Martel, Julien and Bergman, Alexander and Lindell, David and Wetzstein, Gordon},
  journal={Advances in Neural Information Processing Systems},
  volume={33},
  pages={7462--7473},
  year={2020}
}

@article{barron2002universal,
  title={Universal approximation bounds for superpositions of a sigmoidal function},
  author={Barron, Andrew R},
  journal={IEEE Transactions on Information Theory},
  volume={39},
  number={3},
  pages={930--945},
  year={2002},
  publisher={IEEE}
}

@article{wang2024expressiveness,
  title={On the expressiveness and spectral bias of KANs},
  author={Wang, Yixuan and Siegel, Jonathan W and Liu, Ziming and Hou, Thomas Y},
  journal={arXiv preprint arXiv:2410.01803},
  year={2024}
}

@article{jagtap2020locally,
  title={Locally adaptive activation functions with slope recovery for deep and physics-informed neural networks},
  author={Jagtap, Ameya D and Kawaguchi, Kenji and Em Karniadakis, George},
  journal={Proceedings of the Royal Society A},
  volume={476},
  number={2239},
  pages={20200334},
  year={2020},
  publisher={The Royal Society}
}

@article{chen1995universal,
  title={Universal approximation to nonlinear operators by neural networks with arbitrary activation functions and its application to dynamical systems},
  author={Chen, Tianping and Chen, Hong},
  journal={IEEE Transactions on Neural Networks},
  volume={6},
  number={4},
  pages={911--917},
  year={1995},
  publisher={IEEE}
}

@inproceedings{raonic2023convolutional,
  title={Convolutional neural operators},
  author={Raonic, Bogdan and Molinaro, Roberto and Rohner, Tobias and Mishra, Siddhartha and de Bezenac, Emmanuel},
  booktitle={ICLR 2023 workshop on Physics for Machine Learning},
  year={2023}
}

@article{liu2024causality,
author = {Liu, Lizuo and Nath, Kamaljyoti and Cai, Wei},
year = {2024},
month = {05},
pages = {1194-1228},
title = {A Causality-DeepONet for Causal Responses of Linear Dynamical Systems},
volume = {35},
journal = {Communications in Computational Physics},
doi = {10.4208/cicp.OA-2023-0078}
}

@article{RAISSI2019686,
title = {Physics-informed neural networks: A deep learning framework for solving forward and inverse problems involving nonlinear partial differential equations},
journal = {Journal of Computational Physics},
volume = {378},
pages = {686-707},
year = {2019},
issn = {0021-9991},
doi = {https://doi.org/10.1016/j.jcp.2018.10.045},
url = {https://www.sciencedirect.com/science/article/pii/S0021999118307125},
author = {M. Raissi and P. Perdikaris and G.E. Karniadakis}
}

@article{xu2025fp64,
  title={FP64 is All You Need: Rethinking Failure Modes in Physics-Informed Neural Networks},
  author={Xu, Chenhui and Liu, Dancheng and Nassereldine, Amir and Xiong, Jinjun},
  year={2025},
  note={arXiv:2505.10949}
}

@article{amari1998natural,
  title={Natural Gradient Works Efficiently in Learning},
  author={Amari, Shun-ichi},
  journal={Neural Computation},
  volume={10},
  number={2},
  pages={251--276},
  year={1998},
  doi={10.1162/089976698300017746}
}

@article{cai2021physics,
  title={Physics-informed neural networks for heat transfer problems},
  author={Cai, Shengze and Wang, Zhicheng and Wang, Sifan and Perdikaris, Paris and Karniadakis, George Em},
  journal={Journal of Heat Transfer},
  volume={143},
  number={6},
  pages={060801},
  year={2021},
  publisher={American Society of Mechanical Engineers}
}

@article{zhao2024comprehensive,
  title={A comprehensive review of advances in physics-informed neural networks and their applications in complex fluid dynamics},
  author={Zhao, Chi and Zhang, Feifei and Lou, Wenqiang and Wang, Xi and Yang, Jianyong},
  journal={Physics of Fluids},
  volume={36},
  number={10},
  year={2024},
  publisher={AIP Publishing}
}

@article{HE2020103610,
title = {Physics-informed neural networks for multiphysics data assimilation with application to subsurface transport},
journal = {Advances in Water Resources},
volume = {141},
pages = {103610},
year = {2020},
issn = {0309-1708},
doi = {https://doi.org/10.1016/j.advwatres.2020.103610},
url = {https://www.sciencedirect.com/science/article/pii/S0309170819311649},
author = {QiZhi He and David Barajas-Solano and Guzel Tartakovsky and Alexandre M. Tartakovsky}
}

@article{kingma2015adam,
  title={Adam: A Method for Stochastic Optimization},
  author={Kingma, Diederik P. and Ba, Jimmy},
  journal={International Conference on Learning Representations (ICLR)},
  year={2015},
  note={arXiv:1412.6980}
}

@inproceedings{gupta2018shampoo,
  title={Shampoo: Preconditioned Stochastic Tensor Optimization},
  author={Gupta, Vineet and Koren, Tomer and Singer, Yoram},
  booktitle={International Conference on Machine Learning (ICML)},
  year={2018},
  publisher={PMLR},
  note={arXiv:1802.09568}
}

@article{shukla2020physics,
  title={Physics-informed neural network for ultrasound nondestructive quantification of surface breaking cracks},
  author={Shukla, Khemraj and Di Leoni, Patricio Clark and Blackshire, James and Sparkman, Daniel and Karniadakis, George Em},
  journal={Journal of Nondestructive Evaluation},
  volume={39},
  number={3},
  pages={61},
  year={2020},
  publisher={Springer}
}

@article{lu2021learning,
  title={Learning nonlinear operators via DeepONet based on the universal approximation theorem of operators},
  author={Lu, Lu and Jin, Pengzhan and Pang, Guofei and Zhang, Zhongqiang and Karniadakis, George Em},
  journal={Nature Machine Intelligence},
  volume={3},
  number={3},
  pages={218--229},
  year={2021},
  publisher={Nature Publishing Group UK London}
}

@inproceedings{rahaman2019spectral,
  title={On the spectral bias of neural networks},
  author={Rahaman, Nasim and Baratin, Aristide and Arpit, Devansh and Draxler, Felix and Lin, Min and Hamprecht, Fred and Bengio, Yoshua and Courville, Aaron},
  booktitle={International Conference on Machine Learning},
  pages={5301--5310},
  year={2019},
  organization={PMLR}
}

@article{KHODAKARAMI2026108027,
title = {Mitigating spectral bias in neural operators via high-frequency scaling for physical systems},
journal = {Neural Networks},
volume = {193},
pages = {108027},
year = {2026},
issn = {0893-6080},
doi = {https://doi.org/10.1016/j.neunet.2025.108027},
url = {https://www.sciencedirect.com/science/article/pii/S0893608025009074},
author = {Siavash Khodakarami and Vivek Oommen and Aniruddha Bora and George Em Karniadakis}
}

@article{oommen2025integrating,
  title={Integrating neural operators with diffusion models improves spectral representation in turbulence modelling},
  author={Oommen, Vivek and Bora, Aniruddha and Zhang, Zhen and Karniadakis, George Em},
  journal={Proceedings of the Royal Society A},
  volume={481},
  number={2309},
  pages={20240819},
  year={2025},
  publisher={The Royal Society}
}

@article{tancik2020fourier,
  title={Fourier features let networks learn high frequency functions in low dimensional domains},
  author={Tancik, Matthew and Srinivasan, Pratul and Mildenhall, Ben and Fridovich-Keil, Sara and Raghavan, Nithin and Singhal, Utkarsh and Ramamoorthi, Ravi and Barron, Jonathan and Ng, Ren},
  journal={Advances in Neural Information Processing Systems},
  volume={33},
  pages={7537--7547},
  year={2020}
}

@article{WANG2025107179,
title = {On spectral bias reduction of multi-scale neural networks for regression problems},
journal = {Neural Networks},
volume = {185},
pages = {107179},
year = {2025},
issn = {0893-6080},
doi = {https://doi.org/10.1016/j.neunet.2025.107179},
url = {https://www.sciencedirect.com/science/article/pii/S0893608025000589},
author = {Bo Wang and Heng Yuan and Lizuo Liu and Wenzhong Zhang and Wei Cai}
}

@article{hong2022activation,
  title={On the activation function dependence of the spectral bias of neural networks},
  author={Hong, Qingguo and Siegel, Jonathan W and Tan, Qinyang and Xu, Jinchao},
  journal={arXiv preprint arXiv:2208.04924},
  year={2022}
}

@article{wang2025multi,
  title={Multi-scale DeepOnet (Mscale-DeepOnet) for Mitigating Spectral Bias in Learning High Frequency Operators of Oscillatory Functions},
  author={Wang, Bo and Liu, Lizuo and Cai, Wei},
  journal={arXiv preprint arXiv:2504.10932},
  year={2025}
}

@article{oommen2025learning,
  title={Learning turbulent flows with generative models: Super-resolution, forecasting, and sparse flow reconstruction},
  author={Oommen, Vivek and Khodakarami, Siavash and Bora, Aniruddha and Wang, Zhicheng and Karniadakis, George Em},
  journal={arXiv preprint arXiv:2509.08752},
  year={2025}
}

@ARTICLE{chai_10630853,
  author={Chai, Xintao and Cao, Wenjun and Li, Jianhui and Long, Hang and Sun, Xiaodong},
  journal={IEEE Transactions on Geoscience and Remote Sensing}, 
  title={Overcoming the Spectral Bias Problem of Physics-Informed Neural Networks in Solving the Frequency-Domain Acoustic Wave Equation}, 
  year={2024},
  volume={62},
  number={},
  pages={1-20},
  doi={10.1109/TGRS.2024.3440471}}

@article{sallam2023use,
  title={On the use of Fourier Features-Physics Informed Neural Networks ({FF-PINN}) for forward and inverse fluid mechanics problems},
  author={Sallam, Omar and F{\"u}rth, Mirjam},
  journal={Proceedings of the Institution of Mechanical Engineers, Part M: Journal of Engineering for the Maritime Environment},
  volume={237},
  number={4},
  pages={846--866},
  year={2023},
  publisher={SAGE Publications Sage UK: London, England}
}

@article{exp,
title = {{CMINNs}: Compartment model informed neural networks — Unlocking drug dynamics},
journal = {Computers in Biology and Medicine},
volume = {184},
pages = {109392},
year = {2025},
issn = {0010-4825},
doi = {https://doi.org/10.1016/j.compbiomed.2024.109392},
url = {https://www.sciencedirect.com/science/article/pii/S001048252401477X},
author = {Nazanin {Ahmadi Daryakenari} and Shupeng Wang and George Karniadakis},
}

@article{sine,
  title={{AI-Aristotle}: A physics-informed framework for systems biology gray-box identification},
  author={Ahmadi Daryakenari, Nazanin and De Florio, Mario and Shukla, Khemraj and Karniadakis, George Em},
  journal={PLOS Computational Biology},
  volume={20},
  number={3},
  pages={e1011916},
  year={2024},
  publisher={Public Library of Science San Francisco, CA USA}
}

@article{review1,
  title={Physics-informed machine learning in biomedical science and engineering},
  author={Ahmadi, Nazanin and Cao, Qianying and Humphrey, Jay D and Karniadakis, George Em},
  journal={arXiv preprint arXiv:2510.05433},
  year={2025}
}

@article{toscano2410pinns,
  title={From {PINNs} to {PIKANs}: Recent advances in physics-informed machine learning, 2024},
  author={Toscano, Juan Diego and Oommen, Vivek and Varghese, Alan John and Zou, Zongren and Ahmadi Daryakenari, Nazanin and Wu, Chenxi and Karniadakis, George Em},
  journal={Machine Learning for Computational Science and Engineering},
  volume={1},
  number={1},
  pages={15},
  year={2025},
  publisher={Springer}
}

@article{liu2025diminishing,
  title={Diminishing spectral bias in physics-informed neural networks using spatially-adaptive Fourier feature encoding},
  author={Liu, Yarong and Gu, Hong and Yu, Xiangjun and Qin, Pan},
  journal={Neural Networks},
  volume={182},
  pages={106886},
  year={2025},
  publisher={Elsevier}
}

@article{WU2023115671,
title = {A comprehensive study of non-adaptive and residual-based adaptive sampling for physics-informed neural networks},
journal = {Computer Methods in Applied Mechanics and Engineering},
volume = {403},
pages = {115671},
year = {2023},
issn = {0045-7825},
doi = {https://doi.org/10.1016/j.cma.2022.115671},
url = {https://www.sciencedirect.com/science/article/pii/S0045782522006260},
author = {Chenxi Wu and Min Zhu and Qinyang Tan and Yadhu Kartha and Lu Lu}
}

@article{xiong2025separated,
  title={Separated-variable spectral neural networks: a physics-informed learning approach for high-frequency pdes},
  author={Xiong, Xiong and Zhang, Zhuo and Hu, Rongchun and Gao, Chen and Deng, Zichen},
  journal={arXiv preprint arXiv:2508.00628},
  year={2025}
}

@article{jacot2018neural,
  title={Neural tangent kernel: Convergence and generalization in neural networks},
  author={Jacot, Arthur and Gabriel, Franck and Hongler, Cl{\'e}ment},
  journal={Advances in Neural Information Processing Systems},
  volume={31},
  year={2018}
}

@article{chakraborty2025binned,
  title={Binned spectral power loss for improved prediction of chaotic systems},
  author={Chakraborty, Dibyajyoti and Mohan, Arvind T and Maulik, Romit},
  journal={arXiv preprint arXiv:2502.00472},
  year={2025}
}

@article{kiyani2025optimizing,
  title={Optimizing the optimizer for physics-informed neural networks and Kolmogorov-Arnold networks},
  author={Kiyani, Elham and Shukla, Khemraj and Urb{\'a}n, Jorge F and Darbon, J{\'e}r{\^o}me and Karniadakis, George Em},
  journal={Computer Methods in Applied Mechanics and Engineering},
  volume={446},
  pages={118308},
  year={2025},
  publisher={Elsevier}
}

@article{wang2025gradient,
  title={Gradient alignment in physics-informed neural networks: A second-order optimization perspective},
  author={Wang, Sifan and Bhartari, Ananyae Kumar and Li, Bowen and Perdikaris, Paris},
  journal={arXiv preprint arXiv:2502.00604},
  year={2025}
}

@book{nocedal2006numerical,
  title={Numerical optimization},
  author={Nocedal, Jorge},
  publisher={Springer},
  year={2006}
}

@article{toscano2025variational,
  title={A Variational Framework for Residual-Based Adaptivity in Neural PDE Solvers and Operator Learning},
  author={Toscano, Juan Diego and Chen, Daniel T and Oommen, Vivek and Darbon, J{\'e}r{\^o}me and Karniadakis, George Em},
  journal={arXiv preprint arXiv:2509.14198},
  year={2025}
}

@article{AHMADIDARYAKENARI,
title = {Representation meets optimization: Training {PINNs} and {PIKANs} for gray-box discovery in systems pharmacology},
journal = {Computers in Biology and Medicine},
volume = {201},
pages = {111393},
year = {2026},
issn = {0010-4825},
doi = {https://doi.org/10.1016/j.compbiomed.2025.111393},
url = {https://www.sciencedirect.com/science/article/pii/S0010482525017470},
author = {Nazanin {Ahmadi Daryakenari} and Khemraj Shukla and George Em Karniadakis},
}

@book{seber2003linear,
  title={Linear Regression Analysis},
  author={Seber, George AF and Lee, Alan J},
  year={2003},
  publisher={John Wiley \& Sons}
}

@inproceedings{
li2021fourier,
title={Fourier Neural Operator for Parametric Partial Differential Equations},
author={Zongyi Li and Nikola Borislavov Kovachki and Kamyar Azizzadenesheli and Burigede liu and Kaushik Bhattacharya and Andrew Stuart and Anima Anandkumar},
booktitle={International Conference on Learning Representations},
year={2021},
url={https://openreview.net/forum?id=c8P9NQVtmnO}
}

\begin{appendices}

\renewcommand{\thefigure}{\Alph{section}.\arabic{figure}}
\renewcommand{\thetable}{\Alph{section}.\arabic{table}}
\renewcommand{\theequation}{\Alph{section}.\arabic{equation}}

\setcounter{figure}{0}
\setcounter{table}{0}
\setcounter{equation}{0}

\section{Statistical moments}
\label{App:Moments}

\noindent First moment represents the average or direct component of the distribution.
\begin{equation}
    \mu(t)
    =
    \frac{1}{N}
    \sum_{i=1}^{N}
    u_i(t)
\end{equation}

\noindent Second moment represents the variance and is directly related to the total spectral energy and therefore still dominated by low-frequency modes in most physical systems.
\begin{equation}
    \sigma^2(t)
=
\frac{1}{N}
\sum_{i=1}^{N}
\left(u_i(t) - \mu(t)\right)^2
\end{equation}

\noindent Third moment represents the skewness and asymmetry of the distribution and is related to higher modes.
\begin{equation}
    \gamma_1(t)
=
\frac{1}{N}
\sum_{i=1}^{N}
\left(
\frac{u_i(t) - \mu(t)}{\sigma(t)}
\right)^3
\end{equation}

\noindent Fourth moment (Kurtosis) measures the intermittency of the distribution, which correspond to high-frequency spectral content.

\begin{equation}
    \gamma_2(t)
=
\frac{1}{N}
\sum_{i=1}^{N}
\left(
\frac{u_i(t) - \mu(t)}{\sigma(t)}
\right)^4
\end{equation}

\section{Loss history for adaptive activations}

\begin{figure}[H]
    \centering
    \includegraphics[width=1.0\linewidth]{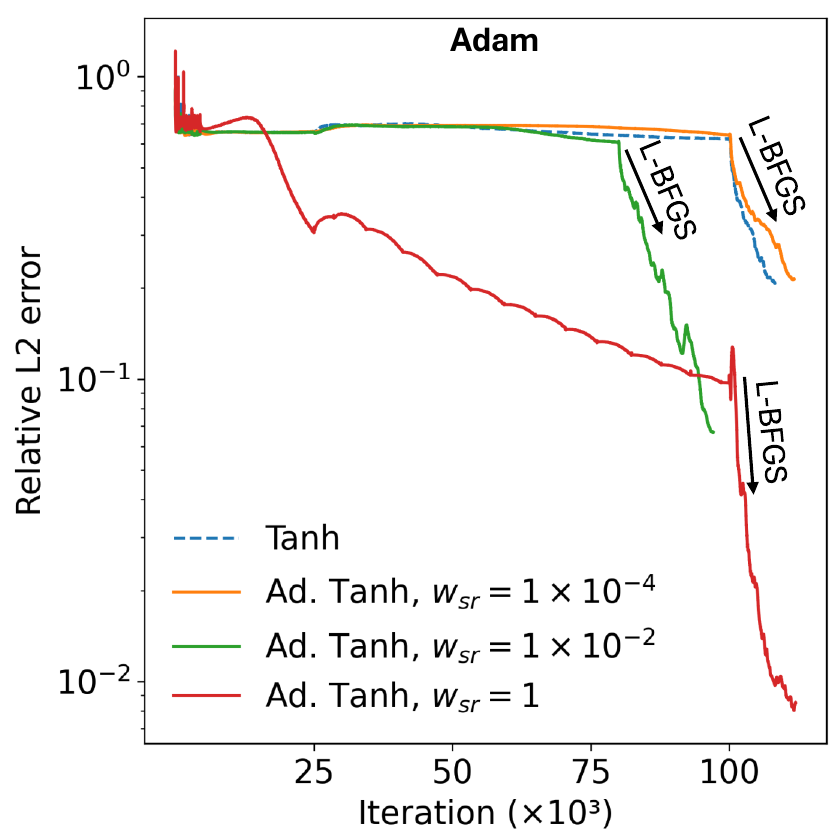}
    \caption{\textbf{PINN with adaptive Tanh activation function training history for KdV equation.} Relative $L^2$ error during training with adaptive activation functions including slope recovery loss term with different weights ($w_{sr}$), trained with Adam. Note that by increasing the weight for the slope recovery term, the errors starts dropping down earlier.}
    \label{fig:KdV_loss_history}
\end{figure}

\section{Wave equations results}

\begin{table}[H]
\centering
\caption{\textbf{Wave equation prediction errors.} 
Comparison of PINN performance with different activation functions and optimizers
for cases (i)-(iii). The results for case (iv) are shown in Table \ref{table:Wave_Eq} of the main text. Note that for case (i) with no multi-scale features or high frequencies, using SIREN with $w_0 = 5$ is better than using SIREN with $w_0 = 30$.}
\label{tab:wave_cases_123}
\setlength{\tabcolsep}{3pt}
\resizebox{\textwidth}{!}{%
\begin{tabular}{lccccc}
\hline
PINN & {\makecell{Rel. $L^2$\\ Error}} &
{\makecell{Barron Norm\\ Rel. $L^2$ Error}} &
\textbf{$\log(e_{\mathcal{F},p=0})$} &
\textbf{$\log(e_{\mathcal{F},p=2})$} &
\textbf{$\log(e_{\mathcal{F},p=4})$}  \\
\hline
\multicolumn{6}{c}{\textbf{Case (i)}} \\
\hline
Adam (Tanh)        & $5.08 \times 10^{-3}$ & $2.94 \times 10^{-3}$ & -5.19 & -1.83 & 2.27  \\
Adam (SIREN)       & $6.11 \times 10^{-3}$ & $4.45 \times 10^{-3}$ & -5.03 & -1.58 & 2.54  \\
SOAP (Tanh)        & $1.19 \times 10^{-3}$ & $2.86 \times 10^{-4}$ & -6.44 & -4.28 & 0.80  \\
SOAP (SIREN, ${w}_0=30$)       & $1.59 \times 10^{-3}$ & $2.19 \times 10^{-4}$ & -5.11 & -4.24 & 0.49  \\
SOAP (SIREN, ${w}_0=5$)       & $8.45 \times 10^{-4}$ & $1.81 \times 10^{-4}$ & -6.59 & -4.20 & 0.50  \\
{SS-Broyden (Tanh)}  & $4.33 \times 10^{-8}$ & $4.99 \times 10^{-8}$ & -14.83 & -9.69 & -3.92  \\
\textbf{SS-Broyden (SIREN)} & $\boldsymbol{2.32 \times 10^{-8}}$ & $\boldsymbol{2.89 \times 10^{-8}}$ & \textbf{-14.95} & \textbf{-10.06} & \textbf{-4.17}  \\
\hline
\multicolumn{6}{c}{\textbf{Case (ii)}} \\
\hline
Adam (Tanh)        & $8.71 \times 10^{-2}$ & $6.20 \times 10^{-2}$ & -2.75  & 0.85 & 5.48 \\
Adam (SIREN)       & $1.72 \times 10^{-2}$ & $1.07 \times 10^{-2}$ & -4.16 & -1.14 & 3.92 \\
SOAP (Tanh)        & $1.32 \times 10^{-3}$ & $1.08 \times 10^{-3}$ & -6.39 & -2.95 & 1.99 \\
SOAP (SIREN)       & $6.19 \times 10^{-4}$ & $3.95 \times 10^{-4}$ & -7.05 & -3.46 & 1.51 \\
{SS-Broyden (Tanh)}  & $1.16 \times 10^{-6}$ & $8.01 \times 10^{-7}$ & -12.50 & -6.99 & -1.39 \\
\textbf{SS-Broyden (SIREN)} & $\boldsymbol{2.81 \times 10^{-7}}$ & $\boldsymbol{2.68 \times 10^{-7}}$ & \textbf{-13.71}  & \textbf{-8.26} & \textbf{-2.50}  \\
\hline
\multicolumn{6}{c}{\textbf{Case (iii)}} \\
\hline
Adam (Tanh)        & 0.84 & 0.68 & -1.39 & 1.96 & 6.91 \\
Adam (SIREN)       & 0.18 & 0.11 & -2.73 & 0.19 & 5.38  \\
SOAP (Tanh)        & $1.38 \times 10^{-3}$ & $1.19 \times 10^{-3}$ & -6.96 & -3.3 & 1.76 \\
SOAP (SIREN)       & $1.23 \times 10^{-3}$ & $7.42 \times 10^{-4}$  & -7.08 & -3.50 & 1.45 \\
{SS-Broyden (Tanh)}  & $4.01 \times 10^{-6}$ & $2.82 \times 10^{-6}$ & -12.04  & -3.30  & 1.76  \\
\textbf{SS-Broyden (SIREN)} & $\boldsymbol{8.37 \times 10^{-7}}$ & $\boldsymbol{7.01 \times 10^{-7}}$ & \textbf{-13.39} & \textbf{-7.56} & \textbf{-1.66} \\
\hline
\end{tabular}
}
\end{table}

\section{Effect of the optimizer on spectral bias in neural operators}
\label{app:optimizer_spectral_bias}

\begin{figure}[t]
    \centering
    \includegraphics[width=0.98\linewidth]{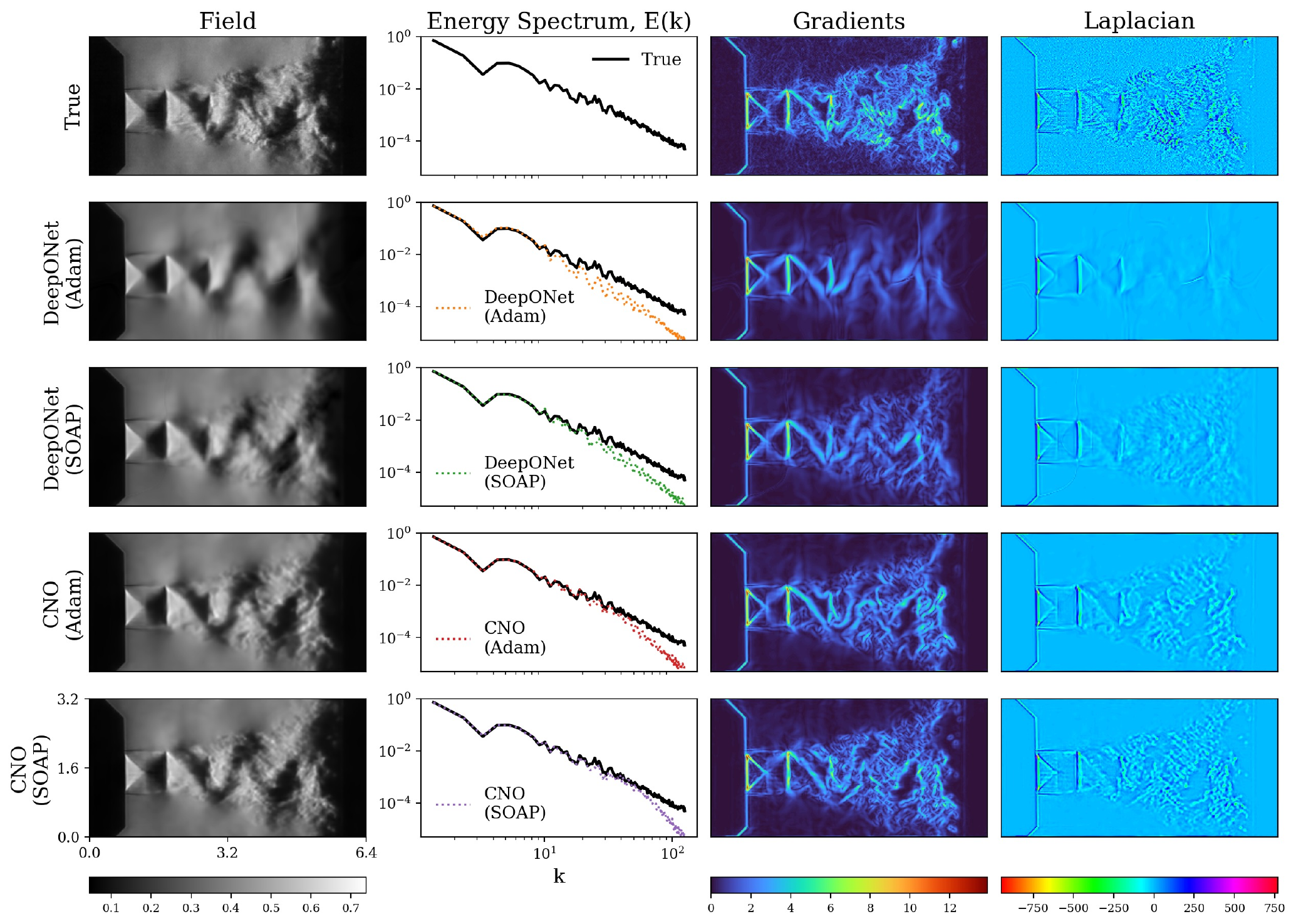}
    \caption{\textbf{Optimizer effects on spectral bias in neural operators.}
    Columns show (from left to right) Schlieren fields, the isotropic energy spectrum $E(k)$, spatial gradients, and the Laplacian.
    Rows compare the ground truth against DeepONet and CNO trained with Adam and SOAP.
    SOAP yields improved agreement with the reference spectrum at high wavenumbers and recovers sharper derivative-based features, indicating reduced attenuation of fine scales relative to Adam.}
    \label{fig:no_opt}
\end{figure}

\noindent In addition to the training objective, the optimizer can materially affect how fast and how well high spatial frequencies are learned.
To isolate optimizer effects from architectural and loss-design choices, we repeat the impinging-jet super-resolution experiment using the same data split and model configurations as in the main text, and we vary only the optimizer.
\autoref{fig:no_opt} visualizes representative reconstructions together with frequency-domain and derivative-based diagnostics.

\autoref{fig:no_opt} compares Adam \cite{kingma2015adam} and SOAP \cite{vyas2024soap} for DeepONet and CNO.
Across both architectures, Adam recovers the dominant large-scale shock topology but exhibits stronger attenuation of the high-wavenumber tail of the energy spectrum, consistent with visibly over-smoothed reconstructions.
SOAP yields closer agreement with the reference spectrum into the high-$k$ regime and produces sharper gradients and Laplacians, indicating improved recovery of fine-scale density-gradient features that are under-represented by Adam.
These observations suggest that optimizer-induced preconditioning can either amplify or suppress the effective learning rate of high-frequency modes, thereby interacting with spectral bias.

Conceptually, Adam is a first-order method that applies a diagonal, per-parameter adaptive scaling based on exponential moving averages of the gradient and squared gradient \cite{kingma2015adam}.
SOAP is a preconditioned method that builds on Shampoo-style curvature information \cite{gupta2018shampoo} and runs Adam-like updates in a rotated (preconditioned) coordinate system, which empirically improves stability and convergence in large-scale training \cite{vyas2024soap}.
We also explored second-order optimizers like SS-Broyden for neural operators, motivated by their strong performance in PINNs.
However, making such optimizers work for data-driven neural operators was a challenge we could not resolve in this work.
Two practical differences exist.
First, many PINN studies successfully use full-batch quasi-Newton training because the number of collocation points is typically modest, for example, \cite{RAISSI2019686} explicitly uses L-BFGS as a quasi-Newton, full-batch optimizer when the training set is small.
Second, precision can be critical in PDE-residual optimization. Recent evidence shows that FP64 can prevent precision-induced stalls of L-BFGS in PINNs \cite{xu2025fp64}.
In contrast, neural operators are typically much bigger, commonly trained in mini-batches on large datasets, and are often run in FP32 for throughput.
Under these constraints, stable estimation and inversion of Fisher or Gauss-Newton structure typically requires aggressive approximations (subsampling, low-rank structure, truncated solves), and the resulting updates can be sensitive to noise, damping, and numerical conditioning.
Developing reliable second-order  variants tailored to neural operators remains a promising direction for future work, especially if combined with improved precision control, better curvature estimators, and scalable linear-solve strategies \cite{amari1998natural}.\\

\begin{figure}[H]
    \centering
    \includegraphics[width=\linewidth]{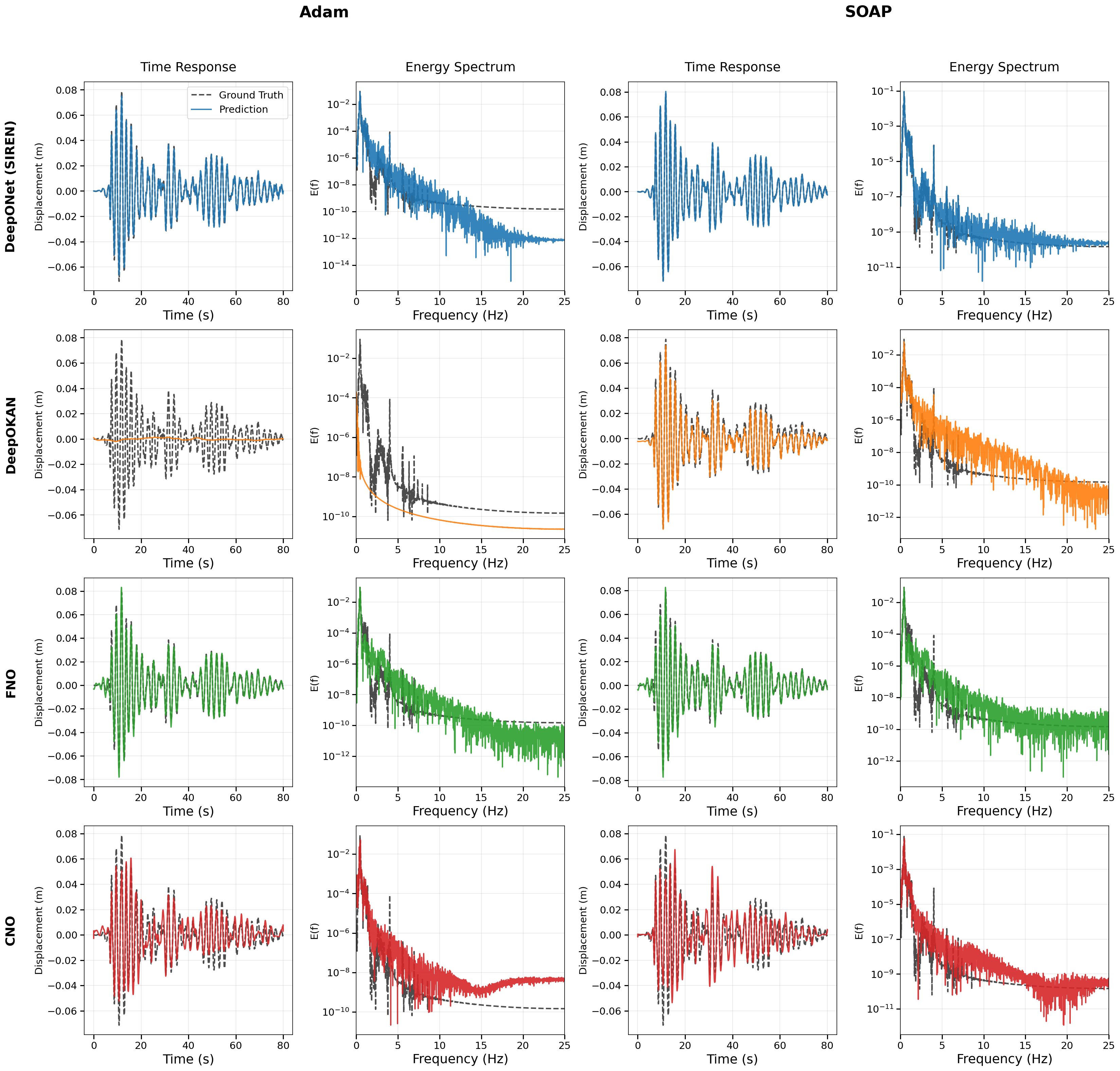}
    \caption{\textbf{SOAP vs Adam for earthquake response prediction.} Time response and energy spectrum for a representative test sample comparing Adam (left pair) and SOAP (right pair). Adam DeepOKAN fails to converge (near-flat prediction and flat spectral tail), while SOAP enables convergence for all architectures with improved spectral accuracy at high frequencies.}
    \label{fig:earthquake_optimizer}
\end{figure}

On the earthquake structural response problem SOAP has a stronger effect, and is critical for DeepONet and DeepOKAN architectures. \label{sec:earthquake_optimizer}
For DeepONet with SIREN, SOAP achieves $2.0\times$ lower NRMSE ($0.0008$ vs $0.0015$) in the baseline configuration, with the advantage persisting under BSP ($1.4\times$ lower NRMSE, $0.0009$ vs $0.0013$).
DeepOKAN with Adam \cite{kingma2015adam} fails to converge entirely, while SOAP achieves $0.0022$ NRMSE, highlighting SOAP's importance for architectures with challenging optimization landscapes.
These optimizer effects hold for both baseline (L2 only) and BSP configurations, as shown by the paired rows in Table~\ref{tab:earthquake_opt}.
FNO and CNO show minimal optimizer sensitivity in field accuracy, but SOAP provides spectral improvement for FNO BSP ($0.082$ vs $0.098$, $1.2\times$).
Moreover, SOAP's advantage extends beyond error magnitude to reliability: across repeated runs with different random seeds, Adam occasionally fails to converge even for DeepONet SIREN (which converges reliably with SOAP), while DeepOKAN with Adam fails consistently across all seeds.
Figure~\ref{fig:earthquake_optimizer} illustrates these effects for a representative test sample.

\section{Additional considerations for the earthquake problem}

\noindent Table~\ref{tab:earthquake_opt} compares SOAP and Adam across all architectures for both baseline and BSP loss configurations with causal training for DeepONet/DeepOKAN, including both SIREN and Tanh activations for DeepONet.

\begin{table}[H]
\centering
\small
\setlength{\tabcolsep}{2pt}
\renewcommand{\arraystretch}{0.85}
\caption{Optimizer, activation, and loss comparison for earthquake response prediction (causal training). $-$ Failed to converge.}
\label{tab:earthquake_opt}
\begin{tabular}{@{}lcccccccc@{}}
\toprule
\textbf{Model} &
\multicolumn{2}{c}{\textbf{\shortstack{Field\\Error}}} &
\multicolumn{2}{c}{\textbf{\shortstack{Log Spectral\\Error}}} &
\multicolumn{2}{c}{\textbf{\shortstack{Barron Norm\\Error}}} &
\multicolumn{2}{c}{\textbf{\shortstack{Accel.\\Error}}} \\
\cmidrule(lr){2-3} \cmidrule(lr){4-5} \cmidrule(lr){6-7} \cmidrule(lr){8-9}
 & Adam & SOAP & Adam & SOAP & Adam & SOAP & Adam & SOAP \\
\midrule
\shortstack[l]{DeepONet\\(SIREN)}       & $0.0015$ & $0.0008$ & $0.0974$ & $0.1257$ & $0.0025$ & $0.0019$ & $0.0381$ & $0.0506$ \\
\shortstack[l]{DeepONet\\(SIREN, BSP)}  & $0.0013$ & $0.0009$ & $0.1026$ & $0.1366$ & $0.0020$ & $0.0027$ & $0.0314$ & $0.0619$ \\
\midrule
\shortstack[l]{DeepONet\\(Tanh)}        & $-$ & $-$ & $-$ & $-$ & $-$ & $-$ & $-$ & $-$ \\
\shortstack[l]{DeepONet\\(Tanh, BSP)}   & $0.0013$ & $0.0006$ & $0.1090$ & $0.1103$ & $0.0021$ & $0.0012$ & $0.0373$ & $0.0303$ \\
\midrule
DeepOKAN               & $-$ & $0.0022$ & $-$ & $0.1230$ & $-$ & $0.0029$ & $-$ & $0.0584$ \\
\shortstack[l]{DeepOKAN\\(BSP)}         & $-$ & $0.0044$ & $-$ & $0.1750$ & $-$ & $0.0077$ & $-$ & $0.1595$ \\
\midrule
FNO                    & $0.0037$ & $0.0038$ & $0.1463$ & $0.1485$ & $0.0047$ & $0.0045$ & $0.0672$ & $0.0645$ \\
\shortstack[l]{FNO\\(BSP)}              & $0.0042$ & $0.0041$ & $0.0983$ & $0.0819$ & $0.0034$ & $0.0030$ & $0.0309$ & $0.0296$ \\
\midrule
CNO                    & $0.0126$ & $0.0133$ & $0.1183$ & $0.1323$ & $0.0039$ & $0.0040$ & $0.0281$ & $0.0325$ \\
\shortstack[l]{CNO\\(BSP)}              & $0.0197$ & $0.0145$ & $0.0774$ & $0.0758$ & $0.0059$ & $0.0043$ & $0.0272$ & $0.0263$ \\
\bottomrule
\end{tabular}
\end{table}

\subsection{Extended BSP comparison}
\label{app:earthquake_bsp}
\begin{figure}[H]
    \centering
    \includegraphics[width=\linewidth]{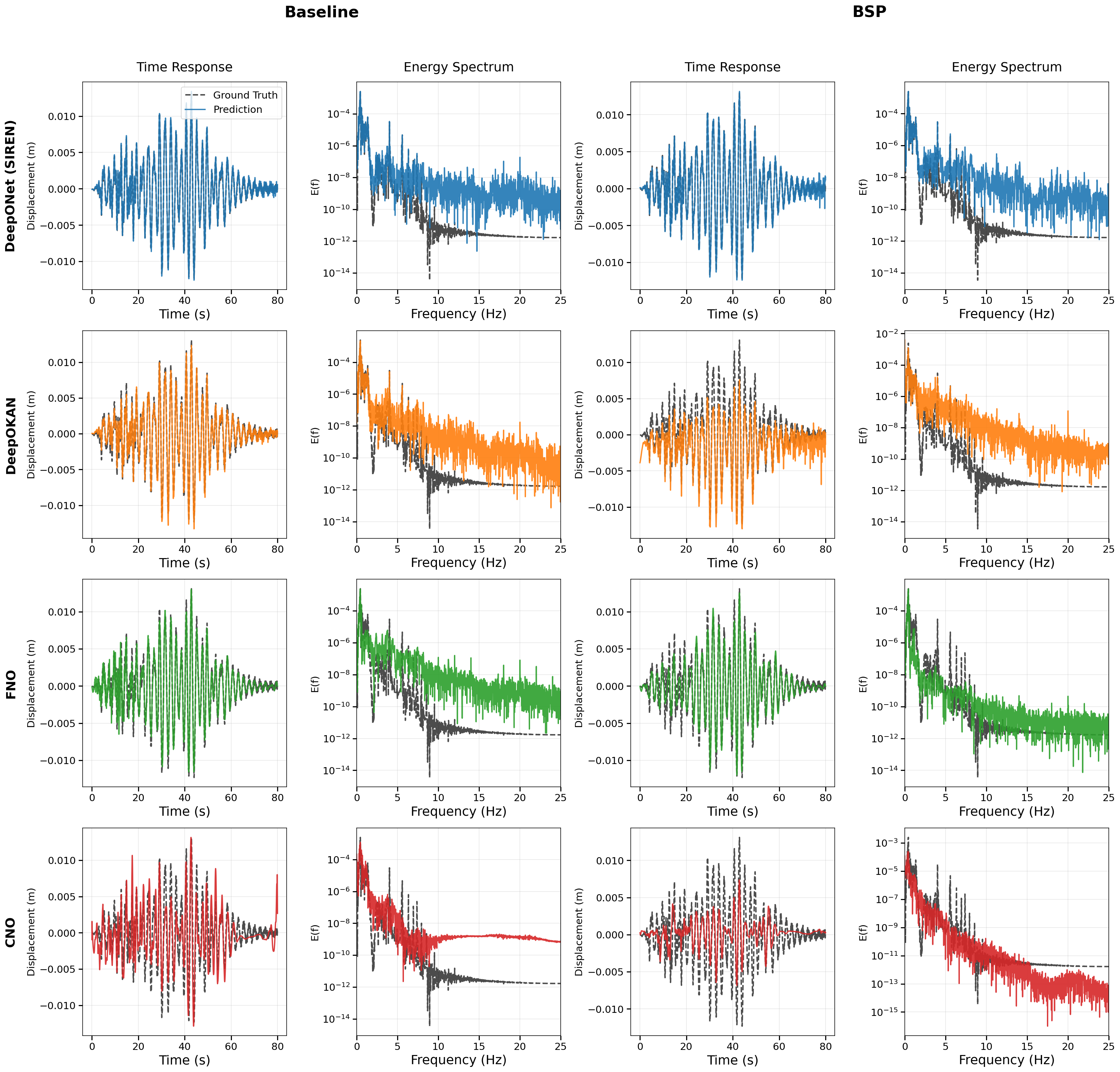}
    \caption{\textbf{Effect of BSP on spectral fidelity for earthquake response prediction.} Time response (columns 1, 3) and energy spectrum (columns 2, 4) for a representative test sample comparing baseline (left pair) and BSP (right pair) training across all four architectures. Note the difference in scale of energy values (y-axis) when comparing these results to the impinging-jet problem.}
    \label{fig:earthquake_bsp}
\end{figure}

\noindent Figure~\ref{fig:earthquake_bsp} extends the representative BSP example in Figure~\ref{fig:earthquake_compact}(b) to all four architectures, showing the full baseline vs BSP comparison.

\subsection{Activation effects (SIREN vs Tanh)}
\label{sec:earthquake_activation}
\begin{figure}[H]
    \centering
    \includegraphics[width=\linewidth]{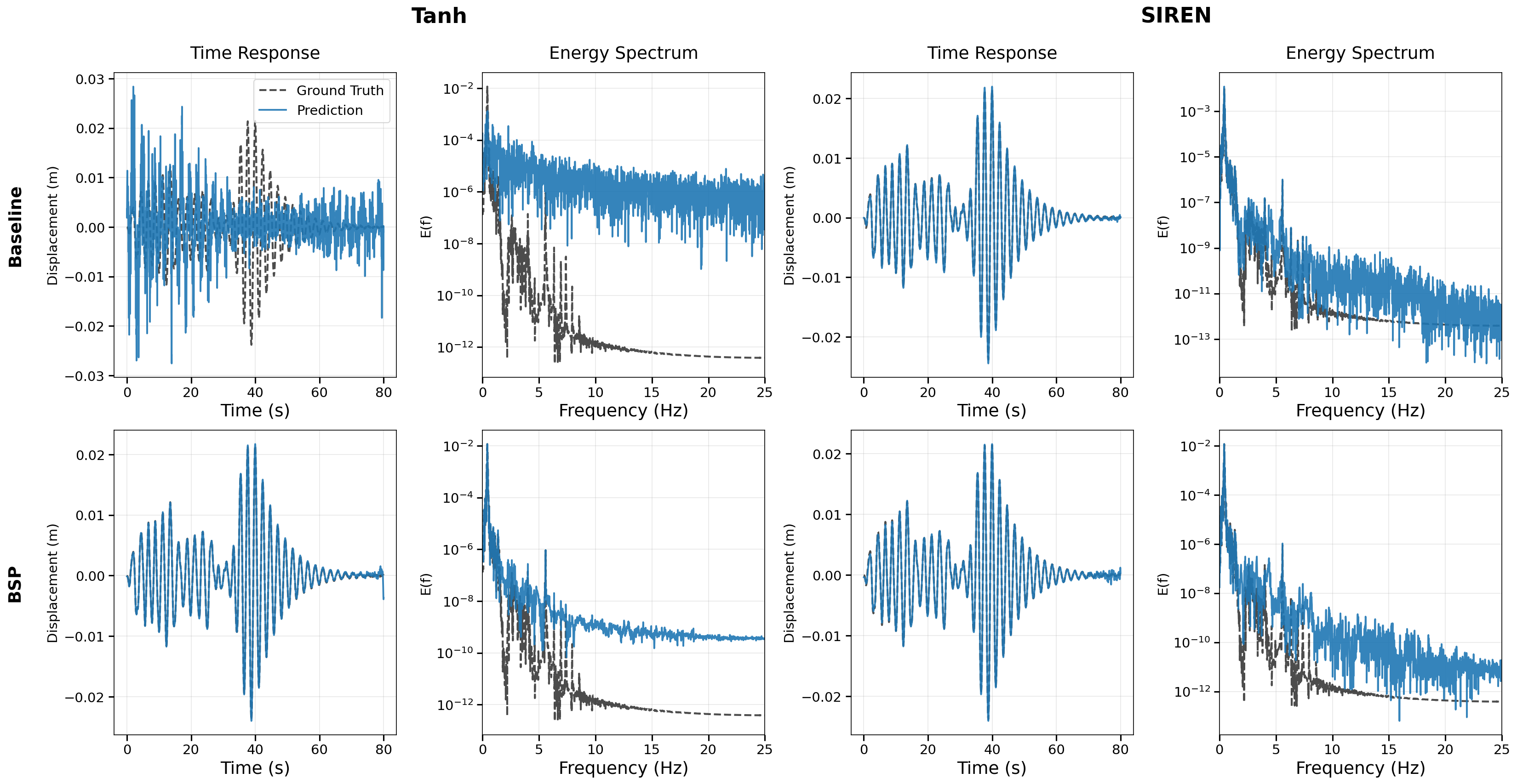}
    \caption{\textbf{Activation function effects on DeepONet.} Time response and energy spectrum comparing Tanh (left pair) and SIREN (right pair) activations for both baseline (top row) and Log-BSP (bottom row) training. Tanh baseline fails to capture the structural response, but BSP enables convergence to the best overall accuracy. SIREN converges reliably without BSP but shows diminishing returns when BSP is added.}
    \label{fig:earthquake_activation}
\end{figure}

\noindent The interaction between activation function and loss formulation reveals complementary mechanisms for spectral bias mitigation.
With baseline training, SIREN converges reliably ($0.0008$ NRMSE) while Tanh fails to converge ($0.024$), demonstrating SIREN's robustness.
However, with BSP, Tanh achieves the best absolute accuracy ($0.0006$ field, $0.110$ spectral) across all configurations, outperforming SIREN+BSP ($0.0009$ field, $0.137$ spectral).
Acceleration error confirms this pattern: Tanh+BSP achieves $0.030$ vs SIREN+BSP $0.062$ ($2.0\times$ better).
This suggests that SIREN's periodic activations provide implicit spectral bias mitigation that overlaps with BSP's explicit regularization; when both are active, the redundant spectral gradients may interfere rather than compound.
In contrast, Tanh relies entirely on BSP for high-frequency content, allowing complementary rather than redundant optimization gradients.
The cosine similarity between the per loss gradients  (Table~\ref{tab:gradient_cosine}) corroborates this interpretation: SIREN exhibits the strongest negative gradient alignment ($-0.48$ final cosine similarity), while Tanh shows weaker conflict ($-0.18$), permitting BSP's spectral signal to provide net benefit.
Figure~\ref{fig:earthquake_activation} illustrates these contrasting mechanisms for a representative test sample.

\subsection{Causal training}
\label{sec:earthquake_causal}
\begin{figure}[H]
    \centering
    \includegraphics[width=\linewidth]{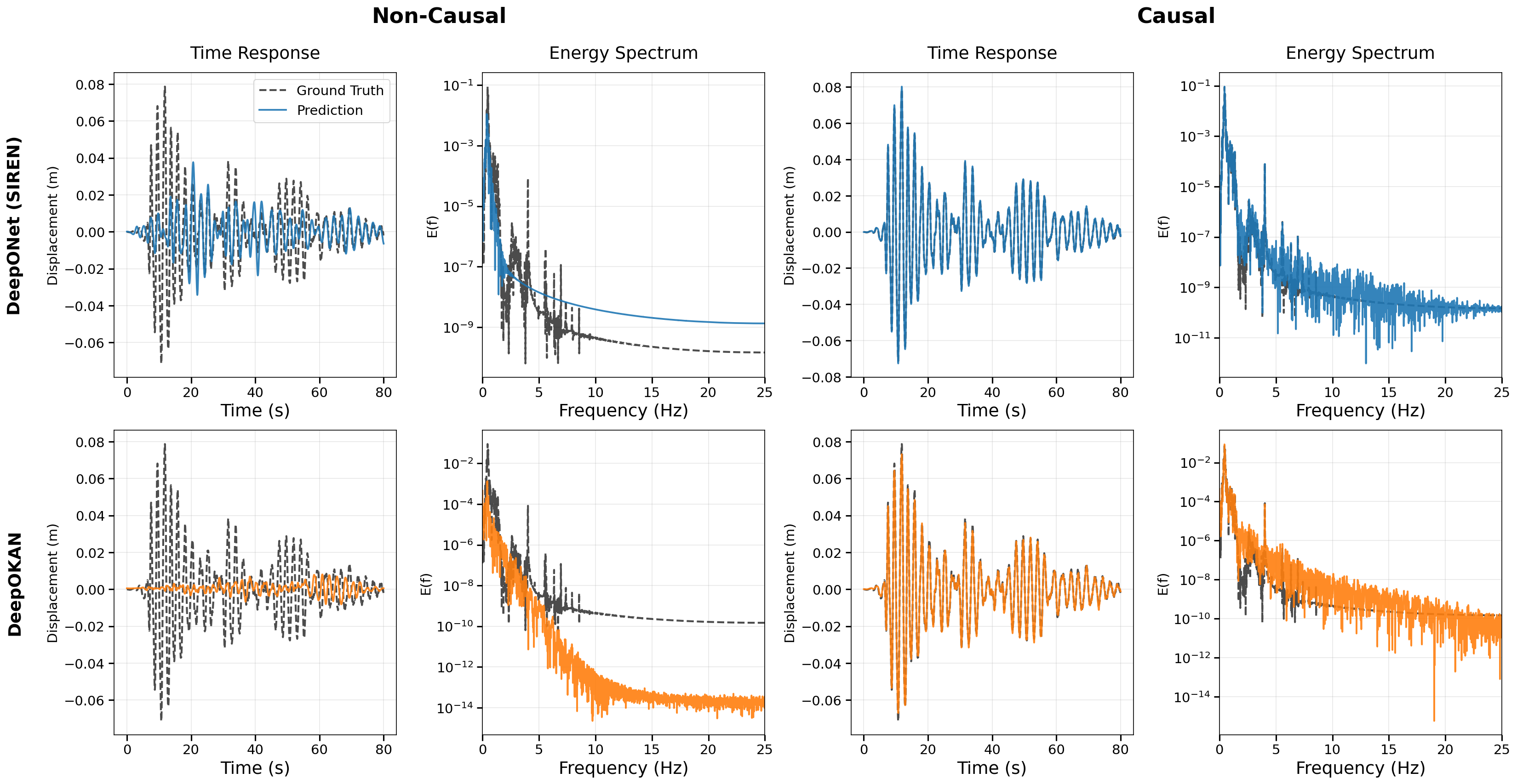}
    \caption{\textbf{Causal vs non-causal training for branch-trunk architectures.} Time response and energy spectrum comparing non-causal training (left pair) and causal training (right pair) for DeepONet and DeepOKAN. Non-causal DeepOKAN fails to converge entirely, while non-causal DeepONet achieves only marginal convergence with substantially degraded spectral fidelity compared to its causal counterpart.}
    \label{fig:earthquake_causal}
\end{figure}

\noindent Causal (per-timestep) training \cite{liu2024causality} proved essential for DeepONet and DeepOKAN on this small dataset.
Non-causal (sequence) variants mostly failed to converge, and those that did achieve marginal convergence produced field errors an order of magnitude worse than their causal counterparts, as shown in Table~\ref{tab:earthquake_causal}. Figure~\ref{fig:earthquake_causal} illustrates the failure mode of non-causal training for a representative test sample.

\begin{table}[H]
\centering
\small
\setlength{\tabcolsep}{2pt}
\renewcommand{\arraystretch}{0.85}
\caption{Causal (C) vs non-causal (NC) training for branch-trunk architectures (SOAP optimizer). $-$ Failed to converge (field error $\geq 0.023$).}
\label{tab:earthquake_causal}
\begin{tabular}{@{}lcccccccc@{}}
\toprule
\textbf{Model} &
\multicolumn{2}{c}{\textbf{\shortstack{Field\\Error}}} &
\multicolumn{2}{c}{\textbf{\shortstack{Log Spectral\\Error}}} &
\multicolumn{2}{c}{\textbf{\shortstack{Barron Norm\\Error}}} &
\multicolumn{2}{c}{\textbf{\shortstack{Accel.\\Error}}} \\
\cmidrule(lr){2-3} \cmidrule(lr){4-5} \cmidrule(lr){6-7} \cmidrule(lr){8-9}
 & C & NC & C & NC & C & NC & C & NC \\
\midrule
\shortstack[l]{DeepONet\\(SIREN)}       & $0.0008$ & $0.0211$ & $0.1257$ & $0.0961$ & $0.0019$ & $0.0054$ & $0.0506$ & \textbf{0.0277} \\
\shortstack[l]{DeepONet\\(SIREN, BSP)}  & $0.0009$ & $-$ & $0.1366$ & $-$ & $0.0027$ & $-$ & $0.0619$ & $-$ \\
\midrule
\shortstack[l]{DeepONet\\(Tanh)}        & $-$ & $-$ & $-$ & $-$ & $-$ & $-$ & $-$ & $-$ \\
\shortstack[l]{DeepONet\\(Tanh, BSP)}   & \textbf{0.0006} & $-$ & \textbf{0.1103} & $-$ & \textbf{0.0012} & $-$ & $0.0303$ & $-$ \\
\midrule
DeepOKAN               & $0.0022$ & $-$ & $0.1230$ & $-$ & $0.0029$ & $-$ & $0.0584$ & $-$ \\
\shortstack[l]{DeepOKAN\\(BSP)}         & $0.0044$ & $-$ & $0.1750$ & $-$ & $0.0077$ & $-$ & $0.1595$ & $-$ \\
\bottomrule
\end{tabular}
\end{table}

\subsection{Gradient alignment}
\label{sec:earthquake_gradient}
\noindent To investigate whether the structural mismatch between per-timestep L2 and sequence-level BSP produces competing gradient signals, we train each configuration while computing per-loss gradients and measuring their cosine similarity for each batch.
Positive values indicate aligned objectives; negative values indicate the two losses pull parameters in opposing directions.
Table~\ref{tab:gradient_cosine} summarises the results across all eight model configurations. To isolate the effect of the causal training structure, we also train DeepONet and DeepOKAN in non-causal mode (both L2 and BSP on the same full-sequence output, identical to FNO/CNO).

\begin{table}[H]
\centering
\small
\setlength{\tabcolsep}{4pt}
\renewcommand{\arraystretch}{0.85}
\caption{Cosine similarity between L2 field-loss and BSP spectral-loss gradients. Positive values indicate aligned optimization directions; negative values indicate competing gradients.}
\label{tab:gradient_cosine}
\begin{tabular}{@{}lcc@{}}
\toprule
\textbf{Model} &
\textbf{Training Mode} &
\textbf{Final Cos Sim} \\
\midrule
DeepONet (SIREN) & Causal & $-0.482$ \\
DeepONet (Tanh)  & Causal & $-0.181$ \\
DeepOKAN         & Causal & $+0.021$ \\
\midrule
DeepONet (SIREN) & Non-causal & $+0.126$ \\
DeepONet (Tanh)  & Non-causal & $+0.107$ \\
DeepOKAN         & Non-causal & $+0.042$ \\
\midrule
FNO              & Non-causal & $+0.282$ \\
CNO              & Non-causal & $-0.185$ \\
\bottomrule
\end{tabular}
\end{table}

\begin{figure}[H]
    \centering
    \includegraphics[width=0.9\linewidth]{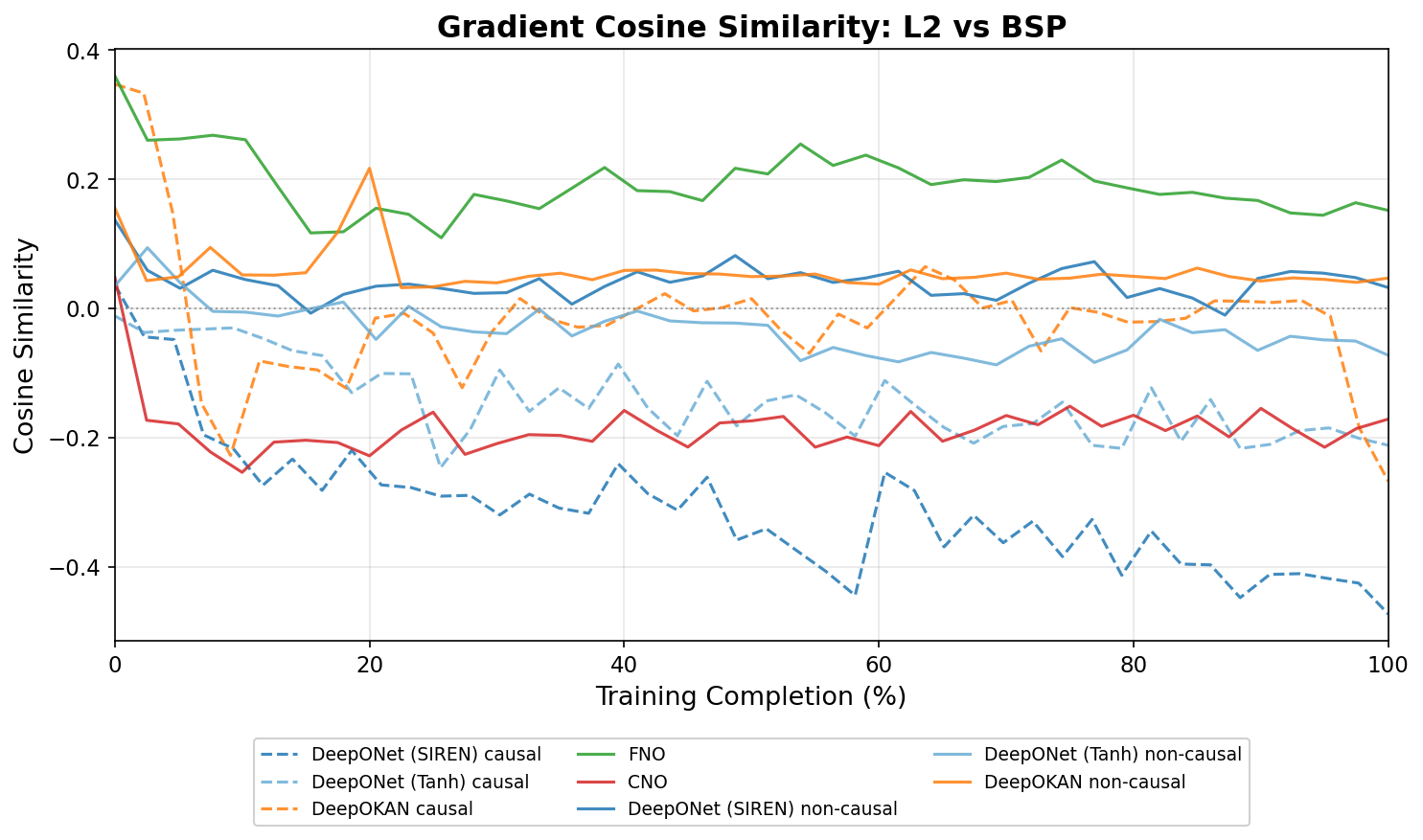}
    \caption{\textbf{Gradient cosine similarity over training.} Cosine similarity between L2 field-loss and BSP spectral-loss gradients as a function of training completion. Dashed lines indicate causal training; solid lines indicate non-causal training. DeepONet (SIREN) under causal training diverges toward strongly negative values, confirming increasing gradient conflict over the course of training.}
    \label{fig:cosine_sim_completion}
\end{figure}

\noindent Figure~\ref{fig:cosine_sim_completion} shows the evolution of gradient cosine similarity over the course of training, confirming that these endpoint values reflect sustained trends rather than transient fluctuations.

\noindent Among the non-causal models, FNO shows the strongest positive alignment ($+0.28$ by the end of training).
Both the field loss and the BSP loss see the same full-sequence forward pass output, so their gradient directions are compatible.
This explains BSP's clear spectral improvement for FNO reported in Table~\ref{tab:earthquake_arch}.\\

DeepONet with SIREN shows the strongest gradient conflict under causal training, with cosine similarity reaching $-0.48$ by the end of training (Figure~\ref{fig:cosine_sim_completion}).
The per-timestep L2 loss and the sequence-level BSP produce opposing gradient signals, directly explaining BSP's ineffectiveness for this configuration.
DeepONet with Tanh shows weaker conflict ($-0.18$ final), consistent with BSP still providing net benefit despite the structural mismatch.
The difference may reflect how SIREN's periodic activations create additional spectral gradient competition beyond the structural mismatch alone. DeepOKAN gradients are near-orthogonal under causal training ($+0.02$ final), consistent with BSP having minimal net directional effect on this architecture.\\

The non-causal results isolate the effect of the causal training structure. When the causal/non-causal mismatch is removed and both losses operate on the same full-sequence output, all three branch-trunk architectures shift toward positive alignment: DeepONet (SIREN) moves from $-0.48$ to $+0.13$, DeepONet (Tanh) from $-0.18$ to $+0.11$, and DeepOKAN from $+0.02$ to $+0.04$. This confirms that the negative gradient alignment is caused by the causal training structure rather than being an intrinsic property of the architectures.
Notably, SIREN's non-causal L2 gradients are slightly more aligned with BSP ($+0.13$) than Tanh's ($+0.11$), which is directionally consistent with the overlapping-mechanism interpretation from \autoref{sec:earthquake_activation}: SIREN's periodic activations naturally produce field-loss gradients that partially overlap with BSP's spectral objective, whereas Tanh's field-loss gradients are more orthogonal to it.\\

CNO shows negative alignment ($-0.19$) yet BSP still helps empirically, though its benefit is weaker than FNO's ($1.2\times$ spectral improvement vs FNO's $1.8\times$). This is potentially due to the model's inherent regularization (due to its bottleneck structure) making the loss landscape more navigable despite the gradient conflict. More investigation is needed to fully explain why BSP's improvement is contextual to the model it's used with.

\end{appendices}
\end{document}